\documentclass[twoside,11pt]{article}

\usepackage{blindtext}

\usepackage{microtype} 

\usepackage[abbrvbib, preprint]{jmlr2e}

\usepackage{mainsetting}

\usepackage{lastpage}
\jmlrheading{26}{2025}{1-\pageref{LastPage}}{12/24; Revised 8/25}{10/25}{24-2243}{Tenghui Li, Guoxu Zhou, Xuyang Zhao, and Qibin Zhao}

\ShortHeadings{Li, Zhou, Zhao, and Zhao}{Scaling Capability in Token Space}
\firstpageno{1}

\begin{document}

\title{Scaling Capability in Token Space: An Analysis of Large Vision Language Model}

\author{
  \name Tenghui Li \email tenghui.lee@foxmail.com \\
  \addr School of Automation, Guangdong University of Technology, Guangzhou 510006, China
  \AND
  \name Guoxu Zhou \thanks{corresponding author} \email gx.zhou@gdut.edu.cn \\
  \addr School of Automation, Guangdong University of Technology, Guangzhou 510006, China \\
  Key Laboratory of Intelligent Detection and the Internet of Things in Manufacturing, \\ 
  Ministry of Education, Guangzhou, China \\
  Guangdong Provincial Key Laboratory of Intelligent Systems and Optimization Integration(GDUT), Guangzhou 510006, China 
  \AND Xuyang Zhao \email xuyang.zhao@riken.jp \\
  \addr Medical Science Data-driven Mathematics Team, RIKEN Center for Interdisciplinary Theoretical and Mathematical Sciences, Yokohama 230-0045, Japan \\
  Medical Data Mathematical Reasoning Special Team, RIKEN Center for Integrative Medical Sciences, Yokohama 230-0045, Japan \\
  Department of Artificial Intelligence Medicine, Chiba University, Chiba 260-0856, Japan
  \AND Qibin Zhao \email qibin.zhao@riken.jp \\
  \addr Tensor Learning Team, RIKEN Center for Advanced Intelligence Project, Tokyo 103-0027, Japan\\
  School of Automation, Guangdong University of Technology, Guangzhou 510006, China
}

\editor{Kai-Wei Chang}

\maketitle

\begin{abstract}
  Large language models have demonstrated predictable scaling behaviors with respect to model parameters and training data.
  This study investigates whether a similar scaling relationship exist for vision-language models with respect to the number of vision tokens.
  A mathematical framework is developed to characterize a relationship between vision token number and the expected divergence of distance between vision-referencing sequences.
  The theoretical analysis reveals two distinct scaling regimes: sublinear scaling for less vision tokens and linear scaling for more vision tokens.
  This aligns with model performance relationships of the form \(S(n) \approx c / n^{\alpha(n)}\), where the scaling exponent relates to the correlation structure between vision token representations.
  Empirical validations across multiple vision-language benchmarks show that model performance matches the prediction from scaling relationship.
  The findings contribute to understanding vision token scaling in transformers through a theoretical framework that complements empirical observations.
\end{abstract}


\section{Introduction}
\label{sec:introduction}

Large language models (LLMs) have demonstrated remarkable capabilities across diverse natural language processing tasks \citep{GPT-3,achiam2023gpt4-technical-report,claude3, touvron2023llama-2,dubey2024llama-3, qwen}.
The successes of these transformer-based architectures have motivated their extension to multimodal settings \citep{fuyu-8b, alayrac2022flamingo,liu2024llava,Qwen-vl,chen2024internvl,lu2024deepseek-vl,hu2024mplug-docowl-1.5,li2023blip-2}.
This integration is typically achieved by employing vision encoders to convert images into discrete tokens that can be processed alongside text tokens within the unified transformer framework.

A fundamental challenge in multimodal transformers is the substantial number of vision tokens generated from images. For example, CLIP ViT-L/14 produces 256 tokens from a \((224 \times 224)\) image, while ViT-H/14 generates 576 tokens from a \((336 \times 336)\) image \citep{radford2021CLIP}. High-resolution models like InternLM-XComposer2-4KHD can produce up to 2377 tokens for a 4K image \citep{dong2024InternLM-xcomposer2-4khd}, leading to significant computational costs. Conversely, efficiency-focused approaches like BLIP-2's Q-Former compress visual information into only 32 fixed tokens \citep{li2023blip-2}, raising questions about potential information loss.
This yields a trade-off between computational efficiency and performance preservation, which motivates us to investigate the relationship between the number of vision tokens and model performance in vision language tasks.
A fundamental question arises: 
\begin{center}
\begin{minipage}[c]{0.80\linewidth}
  \textit{What is the relationship between the number of vision tokens and model performance in vision language tasks?}
\end{minipage}
\end{center}

To systematically investigate this fundamental question, there are some key aspects that need to be addressed.
First, a principled methodology is required to analyze the relationship under more general conditions. 
Second, mathematical operations can be formulated to characterize the relationship.
Third, it is also important to validate the findings in real world vision language benchmarks.
These considerations naturally lead to three interconnected research questions:
\begin{enumerate}
  \setlength{\itemsep}{0pt}
  \setlength{\parskip}{0pt}
  \setlength{\parsep}{0pt}
  \item Can a mathematical framework be formulated to characterize the relationship between vision tokens and model performance?
  \item What are the theoretical properties of this relationship, and do scaling behaviors similar to those observed in language models also exist for vision tokens in vision-language models?
  \item How can these findings be validated across diverse vision language benchmarks under real world conditions?
\end{enumerate}

These three questions are addressed through three complementary approaches that build understanding from theory setup to evaluation under real-world large vision language benchmarks.

\textbf{Mathematical framework.} A mathematical framework is developed to characterize how the number of vision tokens affects model performance.
The core approach is to measure the representational distance between a model's hidden states when processing two distinct input sequences that share identical prefix tokens but diverge in their vision-referencing components. 
The framework operates based on a fundamental insight: a model's ability to distinguish between similar inputs serves as a proxy for its discriminative capability and, consequently, its performance on vision-language tasks.
When the representational distance between two branching sequences is small, the model struggles to differentiate between them, leading to ambiguous predictions and reduced performance.
Conversely, when this distance is large, the model can reliably distinguish between the inputs, enabling confident and accurate responses.

\textbf{Theoretical analysis and scaling properties.}
The analysis reveals that model performance follows predictable mathematical relationships with the number of vision tokens. 
The expected divergence of distance between different vision-referencing sequences \(\mathcal{D}(n)\) scales as
\begin{equation}
  \mathbb{E}_{\mathcal{V}}\left[\mathcal{D}(n)\right] 
  = \bigO{\sqrt{n \left(1 - \psi_{\tequal}^{(AB)}(n)\right) + (n^2-n) \left( \psi_{\tcross}^{(AA)}(n) - \psi_{\tcross}^{(AB)}(n) \right)}},
\end{equation}
where \(n\) represents the number of differing vision tokens and the dependency measures \(\psi_{\tequal}^{(AB)}(n)\), \(\psi_{\tcross}^{(AA)}(n)\), and \(\psi_{\tcross}^{(AB)}(n)\) capture the correlation structure between hidden representations. 
The divergence bound exhibits two distinct scaling regimes: sublinear scaling \(\bigO{\sqrt{n}}\) for small \(n\) when the linear component dominates, and linear scaling \(\bigO{n}\) for large \(n\) when the quadratic component dominates.
This mathematical characterization aligns with scaling relationships analogizing to the scaling laws discovered in language models with respect to parameters and training data \citep{kaplan2020scalinglaw,hoffmann2022ChinchillaScaling}. The scaling has the form of
\begin{equation}
  S(n) \approx \frac{c}{n^{\alpha(n)}},
\end{equation}
where \(S(n)\) denotes model performance, \(c\) is a scaling constant, and \(\alpha(n)\) is the scaling exponent. 
Under specific conditions, the scaling exponent \(\alpha(n)\) can be expressed as \(\alpha(n) = \frac{\beta}{2} \log_{n} \left(
  \frac{1 + \left. n \right|_{\rho(n) = 1}}%
      {n + \left. n \right|_{\rho(n) = 1}}
\right) - \frac{\beta}{2}\),
where \(\left. n \right|_{\rho(n)=1}\) represents a critical balance point between \(\bigO{\sqrt{n}}\) and \(\bigO{n}\) scaling regimes, and \(\beta\) is a constant that relates model performance to the expected divergence. For a more comprehensive discussion of these scaling properties in token space, please refer to Subsection \ref{subsec:dtsa-divergence-scaling}.

\textbf{Model architecture and validation.} A specific vision language model architecture is designed to evaluate the theoretical predictions in practice.
Following the popular architecture design of vision language models \citep{liu2024llava,xu2024llava-uhd,dong2024InternLM-xcomposer2-4khd}, the model is built on vision encoder as vision token generator and large language model as the backbone. 
One main distinct requirement is that the architecture can be easily adapted to generate different numbers of vision tokens for controlled testing.
Comprehensive experiments across multiple vision-language benchmarks are conducted to validate the theoretical predictions and demonstrate the effectiveness of the architecture.
Results show that the model's performance follows the predicted scaling relationship, providing empirical evidence for the theoretical framework's accuracy.

\textit{Paper organization.}
The remainder of this paper is organized as follows.  
Relevant prior work is reviewed in Section~\ref{sec:related-works}.  
A theoretical framework for analyzing vision token scaling through sensitivity analysis is presented in Section~\ref{sec:div_token_analysis}.  
The proposed model architecture, designed to systematically investigate the scaling relationships, is described in Section~\ref{sec:proposed_method}.  
Comprehensive experimental validation across multiple benchmarks is provided in Section~\ref{sec:experiments}.  
Finally, conclusions, including implications and directions for future research, are drawn in Section~\ref{sec:conclusion}.

\section{Related Works}
\label{sec:related-works}

This section reviews relevant prior work across three key areas: vision language model architectures, scaling laws in neural models, and approaches to vision token optimization. The relationship between these areas and the present work is clarified to establish the distinctive contributions of this research.

\textbf{Vision large language models.}
Vision language models integrate visual and textual information to enable vision capabilities in language models.
Some methods incorporate specialized structures to facilitate cross-modal interactions.
For example, Flamingo \citep{alayrac2022flamingo} applies cross-attention mechanisms between vision and language tokens through gated cross-attention layers, while CogVLM \citep{wang2023cogvlm} incorporates vision expert modules comprising specialized attention and MLP components within each transformer layer.
A more prevalent approach leverages existing language models by concatenating vision tokens alongside text tokens, which has become the dominant paradigm in recent vision language models such as LLaVA \citep{liu2024llava}, MiniGPT-4 \citep{zhu2023minigpt-4} and other vision language models.

The number of vision tokens varies significantly across different methods and represents a critical design consideration.
InternLM-XComposer2-4KHD \citep{dong2024InternLM-xcomposer2-4khd} and LLaVA-UHD \citep{xu2024llava-uhd} utilize high-resolution vision processing and produce thousands of tokens per image, which is computationally expensive but enables detailed visual understanding.
Conversely, other methods aim to reduce the number of vision tokens while maintaining performance.
BLIP-2 \citep{li2023blip-2} compresses visual information into 32 fixed tokens through its Q-Former architecture, demonstrating substantial token reduction with maintained performance.
VisPrune \citep{zhang2024FasterVLM} and VisionZip \citep{yang2025visionzip} apply pruning and merging strategies to reduce token requirements, while VoCo-LLaMA \citep{ye2025voco-llama} introduces a learnable compression token to reduce vision KV cache during inference.

\textbf{Scaling properties.}
Scaling laws in large language models describe the performance of language models with respect to model size, training data, and computational resources.
Kaplan scaling \citep{kaplan2020scalinglaw} established that language model performance follows predictable power-law relationships with model parameters \(N\) and dataset size \(D\), where the cross-entropy loss \(L\) scales according to power-law relationships.
Subsequent work Chinchilla scaling \citep{hoffmann2022ChinchillaScaling} refined these relationships for optimal resource allocation, establishing improved scaling coefficients and training procedures.
Research by \citet{JMLR:ScalingData-ConstrainedLanguageModels} investigates scaling laws with limited amounts of unique data.

\textbf{Scaling with the number of vision tokens.}
Recent research has begun investigating vision token utilization from practical optimization perspectives, though with different objectives and methodologies than the present work.
Research \citet{li2024InferenceOptmalVLMsNeedFewerVisualTokens} investigate trade-offs between large models with fewer visual tokens and smaller models with more visual tokens under given computational budgets. Their work derives empirical scaling relationships through experiments across various model sizes and vision token configurations, focusing on inference-time optimization and resource allocation.
The present research differs fundamentally from this prior work in several key aspects.
First, the research scope addresses different fundamental questions: \citet{li2024InferenceOptmalVLMsNeedFewerVisualTokens} optimizes inference configurations by balancing model size against the number of vision tokens under computational budgets. This research investigates the intrinsic mathematical relationship between vision token quantity and model discriminative capacity, revealing distinct scaling regimes (\(\bigO{\sqrt{n}}\) and \(\bigO{n}\)) that emerge from correlation structures between vision representations independent of model size considerations.
Second, the methodological approach differs substantially: \citet{li2024InferenceOptmalVLMsNeedFewerVisualTokens} conducts empirical experiments across various model configurations to derive scaling relationships through curve fitting, whereas this research develops a mathematical framework based on representational distance analysis that measures how models distinguish between vision-referencing sequences with shared prefix tokens.
Third, the theoretical foundation diverges significantly: their work derives scaling relationships from observed experimental data patterns, while this research establishes mathematical descriptions based on the statistical properties of vision token representations and connects these properties to scaling behavior through divergence analysis of hidden state correlations.

While the existing literature demonstrates significant progress in vision token optimization and practical compression techniques, this work provides a systematic theoretical framework for understanding the fundamental relationship between the number of vision tokens and model performance.
Unlike empirical approaches, the mathematical analysis is possible to establishes a relatively general principles.

\section{Divergence Token Analysis}
\label{sec:div_token_analysis}

\subsection{Motivation}
\label{subsec:dtsa-motivation}

In this section, the relationship between the number of newly introduced vision tokens and model performance on text generation given text prompts and vision inputs is investigated.
However, directly measuring performance presents significant challenges, as it typically requires evaluation on specific datasets and may not generalize across different contexts.
To address this limitation, this problem is reformulated using a more tractable approach.

Instead of directly measuring performance, the model's sensitivity to input variations is analyzed.
This approach builds on a fundamental property of autoregressive models:
under deterministic generation settings, for example random sampling is disabled or greedy decoding is used, identical inputs produce identical outputs.
This deterministic behavior occurs because the model selects each token based on maximum likelihood estimation given the input prefix, eliminating stochastic variation.

Given this deterministic property, it is possible to examine how models respond to systematically varied inputs.
Specifically, pairs of inputs that differ only in their suffix tokens are considered. The central question becomes: \textit{how many suffix tokens are needed before the model produces noticeably different outputs?}

The intuition behind this approach is straightforward. When only one suffix token differs between two inputs, the differential information is limited, making it difficult for the model to distinguish between them.
After more and more suffix tokens are introduced, the model gains additional information to differentiate the inputs, leading to increasingly divergent outputs. Eventually, with sufficient differential information, the model should easily distinguish between the inputs and produce substantially different responses.

This sensitivity analysis serves as a proxy for model capability. Models with better representational capacity should demonstrate appropriate sensitivity, detecting meaningful differences while maintaining stability for minor variations. The relationship between vision token quantity and this sensitivity profile provides insights into how additional vision tokens contribute to model performance.

This approach offers several advantages: it requires no dataset-specific evaluation, provides a more general framework applicable across different contexts, and directly measures the model's discriminative capacity.
By analyzing how the introduction of vision tokens affects input sensitivity, it can help infer their impact on model performance without the complications of directly evaluation.

\subsection{Input Pattern Formalization}
\label{subsec:dtsa-input_pattern_formalization}

To simplify the sensitivity analysis systematically, a uniform representation is required.
Let \token{txt} denote text tokens and \token{vis} represent vision tokens. Input sequences in vision language model typically exhibit three fundamental patterns:
\begin{enumerate}
  \item \(\tokenm{txt} \ldots \tokenm{txt} \underline{\tokenm{vis} \ldots \tokenm{vis}}\): Vision tokens follow text tokens (e.g., ``What is in the image? \(\tokenm{vis} \ldots\)'')
  \item \(\tokenm{txt} \ldots \tokenm{txt} \underline{\tokenm{vis} \ldots \tokenm{vis}} \tokenm{txt} \ldots \tokenm{txt}\): Vision tokens are surrounded by text (e.g., ``There is an image \(\tokenm{vis} \ldots \tokenm{vis}\) Please describe it.'')
  \item \(\underline{\tokenm{vis} \ldots \tokenm{vis}} \tokenm{txt} \ldots \tokenm{txt}\): Vision tokens precede text (e.g., ``\(\tokenm{vis} \ldots \tokenm{vis}\) What is in the image?'')
\end{enumerate}

These patterns can be unified through logical decomposition. Pattern 1 partitions into prefix text tokens and suffix vision tokens, pattern 2 decomposes into prefix text tokens and a compound component of vision tokens with trailing text, and pattern 3 constitutes a special case with zero-length prefix text. This yields a uniform representation:
\begin{equation}
  \left[\tokenm{txt} \ldots \tokenm{txt}\right] \left[\tokenm{vis} \ldots \right].
\end{equation}

This representation can be further refined by systematically separate the prefix text tokens based on their relationship to visual content.
The first category encompasses purely instructional content without visual grounding. Examples include phrases such as ``Describe this image.'' or ``What is this?'' which serve as structural prompts that remain meaningful independent of visual content.
The second category contains content that explicitly anchors linguistic elements to visual features.
For instance, ``What is in the box?'', where ``What is in the'' does not refer to any specific visual element, while ``box?'' could be directly referenced to the visual content.
Through this separation, the uniform representation can be further updated to:
\begin{equation}
  \label{equ:logical_representation}
  \underbrace{\left[\tokenm{txt} \ldots \tokenm{txt}\right]}_{\text{non-referential}}
  \overbrace{
    \left[\tokenm{txt} \ldots \tokenm{txt}\right]
    \left[\tokenm{vis} \ldots \right]
  }^{\text{vision-referencing}}.
\end{equation}

Building upon the uniform input pattern formalization, a formal mathematical framework is introduced to analyze token sensitivity.
Without loss of generality, it can be assumed that non-referential prefix text tokens remain constant across input variations, while only the vision-referencing components differ between test cases.
Let \(\{\tvar{x}_1, \tvar{x}_2, \ldots, \tvar{x}_{k}\}\) denote the sequence of non-referential prefix text tokens that remain identical across input variations.
Suppose there are two distinct vision-referencing components, \(\{\tvar{a}_{k+1}, \tvar{a}_{k+2}, \ldots, \tvar{a}_{k+n}\}\) and \(\{\tvar{b}_{k+1}, \tvar{b}_{k+2}, \ldots, \tvar{b}_{k+n}\}\).
The input sequences can then be expressed as:
\begin{equation}
  \label{equ:input_sequences_diverge}
  \begin{matrix}
    \tvar{x}_{1} & \tvar{x}_{2} & \ldots & \tvar{x}_{k}
  \end{matrix}
  ~~ \left\{ ~~
  \begin{matrix}
    \tvar{a}_{k+1} & \tvar{a}_{k+2} & \ldots & \tvar{a}_{k+n} \\
    \tvar{b}_{k+1} & \tvar{b}_{k+2} & \ldots & \tvar{b}_{k+n} \\
  \end{matrix}
  \right.
\end{equation}

The divergence effect amplifies progressively as the number of differing tokens increases.
When only one input token at position \(k\) differs (\(\{\ldots, \tvar{x}_{k}, \tvar{a}_{k+1}\}\) and \(\{\ldots, \tvar{x}_{k}, \tvar{b}_{k+1}\}\), \(\tvar{a}_{k+1} \neq \tvar{b}_{k+1}\)), the contextual discrepancy remains minimal.
However, as more tokens differ (\(\tvar{a}_{k+1} \neq \tvar{b}_{k+1}, \tvar{a}_{k+2} \neq \tvar{b}_{k+2}, \ldots\)), the cumulative contextual differences compound at each generation step, leading to increasingly divergent output trajectories.

This phenomenon creates distinct \textit{branches} in the output space. Each unique suffix sequence following the shared prefix \(\{\tvar{x}_1, \tvar{x}_2, \ldots, \tvar{x}_{k}\}\) defines a separate trajectory in the model's generative process. The trajectory corresponding to suffix \(\{\tvar{a}_{k+1}, \tvar{a}_{k+2}, \ldots, \tvar{a}_{k+n}\}\) produces output branch \(\{ \tvar{a}_{k+n+1}, \tvar{a}_{k+n+2}, \ldots \}\), while suffix \(\{ \tvar{b}_{k+1}, \tvar{b}_{k+2}, \ldots, \tvar{b}_{k+n} \}\) generates output branch \(\{ \tvar{b}_{k+n+1}, \tvar{b}_{k+n+2}, \ldots \}\).
The autoregressive conditioning ensures that tokens generated at position \( > k + n\) depend recursively on the diverging contexts, amplifying the initial differences.

Based on this framework, two sub research questions are formulated to guide the investigation under the theoretical perspective:
\begin{itemize}
  \item \textbf{Metric definition:} How to quantify the distance between two input branches of \(\{\ldots, \tvar{x}_{k}, \tvar{a}_{k+1}, \tvar{a}_{k+2}, \ldots, \tvar{a}_{k+n}\}\) and \(\{\ldots, \tvar{x}_{k}, \tvar{b}_{k+1}, \tvar{b}_{k+2}, \ldots, \tvar{b}_{k+n}\}\)? This question addresses the fundamental tool on which how to measure the divergence effect and assess the impact of input diversity.
  \item \textbf{Divergence scaling:} What is the functional relationship between the number of differing input tokens \(n\) and the distance between two input branches? Understanding this relationship provides insights into how the divergence effect scales with input diversity, which is crucial for forecasting the threshold value on significant divergence.
\end{itemize}

\subsection{Metric Definition}
\label{subsec:dtsa-metric_definition}

In the following, the metric definition introduced in subsection \ref{subsec:dtsa-input_pattern_formalization} is further formalized to quantify the divergence between different computational branches in transformer-based language models.

Consider two input sequences with the pattern established in Equation \eqref{equ:input_sequences_diverge}.
With the uniform input pattern formalization, it can be assumed that non-referential prefix text tokens remain constant across input variations, while only the vision-referencing components differ between test cases.
Consider two input sequences with distinct vision-referencing components:
\begin{itemize}
  \item Input sequence A: \(\{\tvar{x}_{1}, \ldots, \tvar{x}_{k}, \tvar{a}_{k+1}, \ldots, \tvar{a}_{k+n}\}\),
  \item Input sequence B: \(\{\tvar{x}_{1}, \ldots, \tvar{x}_{k}, \tvar{b}_{k+1}, \ldots, \tvar{b}_{k+n}\}\),
\end{itemize}
where, \(\{\tvar{x}_{1}, \ldots, \tvar{x}_{k}\}\) constitutes the shared sequence of non-referential prefix text tokens that remain identical across input variations, while \(\{\tvar{a}_{k+1}, \ldots, \tvar{a}_{k+n}\}\) and \(\{\tvar{b}_{k+1}, \ldots, \tvar{b}_{k+n}\}\) represent the distinct vision-referencing components for sequences A and B, respectively.

In transformer-based large language models, the generation process is fundamentally linked to the hidden representations of the input sequence, specifically, the intermediate outputs of the model prior to the final linear projection layer.
When analyzing divergent branches, these hidden representations can provide a more appropriate and informative metric than the raw input sequences themselves.
Therefore, the hidden representations at each position in the sequence are considered as the primary object to be analyzed.

Since the prefix tokens \(\{\tvar{x}_{1}, \ldots, \tvar{x}_{k}\}\) are identical in both sequences, their corresponding hidden representations are also identical. Consequently, the distance between these sequences should be zero up to position \(k\), with divergence beginning at position \(k+1\) where the suffixes start to differ.
Denote \(\tvar{h}_i^{(A)}\) and \(\tvar{h}_i^{(B)}\) as the hidden representations at position \(k+i\) for sequences A and B, respectively, and they are computed through the transformer architecture as:
\begin{equation}
  \begin{aligned}
    \tvar{h}_i^{(A)} & = \LLM(\tvar{x}_1, \ldots, \tvar{x}_{k}, \tvar{a}_{k+1}, \ldots, \tvar{a}_{k+i})[-1] \\
    \tvar{h}_i^{(B)} & = \LLM(\tvar{x}_1, \ldots, \tvar{x}_{k}, \tvar{b}_{k+1}, \ldots, \tvar{b}_{k+i})[-1]
  \end{aligned}
\end{equation}
where \(\LLM\) denotes the transformer model without the final linear projection layer, and \([-1]\) indicates the final hidden state. Each hidden representation \(\tvar{h}_i^{(A)}\) and \(\tvar{h}_i^{(B)}\) is computed by processing all preceding tokens from \(0\) to \(k+i\).

The divergence metric between the two computational branches is formalized as a function of their respective hidden representation sequences:
\begin{equation}
  \mathcal{D}(n) := \mathcal{D}\left\{
    \tvar{h}_{1}^{(A)}, \ldots, \tvar{h}_{n}^{(A)};
    \tvar{h}_{1}^{(B)}, \ldots, \tvar{h}_{n}^{(B)}
  \right\}
\end{equation}
where \(n\) denotes the number of differing tokens in the vision-referencing components, and \(\mathcal{D}(n)\) quantifies the cumulative divergence between the two branches after processing \(n\) distinct tokens.

To develop an intuition of this distance metric, a geometric perspective is adopted, where hidden representations are considered as points in a high-dimensional space.
The sequence progression can be conceptualized as a trajectory through this space, analogous to a reflectively random walk where each step represents the incorporation of a new token.
Both branches begin at the same initial point (the shared prefix representation). As the sequences diverge through different vision-referencing components, each branch follows a distinct trajectory through the hidden space. The token at position \(k+i\) moves the branch from its current position to a new location, creating a path that reflects the cumulative influence of all processed tokens.

The cumulative displacement between the two branches after \(n\) steps is captured by the sum of individual step differences:
\begin{equation}
  \left( \sum_{i=1}^{n} \tvar{h}_{i}^{(A)} \right) -
  \left( \sum_{i=1}^{n} \tvar{h}_{i}^{(B)} \right)
  = \sum_{i=1}^{n} \left( \tvar{h}_{i}^{(A)} - \tvar{h}_{i}^{(B)} \right)
\end{equation}

This motivates our computation of the distance between two branches, and the distance \(\mathcal{D}(n)\) between the two branches after \(n\) steps can be expressed as:
\begin{equation}
  \label{equ:distance}
  \mathcal{D}(n)
  := \left\| \sum_{i=1}^{n} \left( \tvar{h}_i^{(A)} - \tvar{h}_i^{(B)} \right) \right\|_F,
\end{equation}
where \(\| \cdot \|_F\) represent the Frobenius norm.
This cumulative distance metric captures the divergence between computational branches, providing a principled foundation for analyzing how different token sequences lead to distinct model behaviors.

\subsection{Divergence Scaling}
\label{subsec:dtsa-divergence-scaling}

Subsection \ref{subsec:dtsa-metric_definition} established a distance metric. The subsequent investigation will explore how representational divergence scales with the number of differing vision-referencing tokens.
This analysis is essential for two primary reasons: First, it reveals when statistically significant divergence occurs during trajectory separation.
Second, it quantifies how input diversity propagates through the model's computational pathways, ultimately affecting output behaviors.

The foundation of our approach rests on the setting that distinct vision-referencing sequences generate unique computational trajectories in hidden space.
Each suffix following the shared prefix creates a separate representational pathway.
To derive generalizable insights beyond specific token sequences, an analysis over the expectation of various pairs of branches is considered to capture the model's general divergence behavior.

Let \(\mathcal{V}\) denote the set of all vision-referencing sequence pairs that differ in at least one position, where input sequences are denoted \(\{\ldots, \tvar{a}_{k+1}, \ldots, \tvar{a}_{k+n}\}\) and \(\{\ldots, \tvar{b}_{k+1}, \ldots, \tvar{b}_{k+n}\}\). The expected divergence after \(n\) differing tokens is computed as:
\begin{equation}
  \label{equ:expected_distance}
  \mathbb{E}_{\mathcal{V}}\left[ \mathcal{D}(n) \right]
  = \mathbb{E}_{\mathcal{V}}\left[ \left\| \sum_{i=1}^{n} \left( \tvar{h}_i^{(A)} - \tvar{h}_i^{(B)} \right) \right\|_F \right].
\end{equation}
Based on the settings, if \(n \le 0\), the sequence pairs are identical, and the divergence should be zero, \(\mathbb{E}_{\mathcal{V}}\left[ \mathcal{D}(n) \right] = 0\).
Hence, the expectation is only meaningful for integer \(n \ge 1\).
The expectation in Equation \eqref{equ:expected_distance} can be further bounded as:
\begin{equation}
  \label{equ:exp_sum_sqrt}
  \mathbb{E}_{\mathcal{V}} \left[ \left\| \sum_{i=1}^{n} \left( \tvar{h}_i^{(A)} - \tvar{h}_i^{(B)} \right) \right\|_F \right]
  \le \left(
  \mathbb{E}_{\mathcal{V}} \left[ \left\| \sum_{i=1}^{n} \left( \tvar{h}_i^{(A)} - \tvar{h}_i^{(B)} \right) \right\|_F^2 \right]
  \right)^{1/2}.
\end{equation}
The expectation of the squared Frobenius norm can be decomposed into diagonal \((i = j)\) and cross \((i \ne j)\) terms, revealing the contribution of both individual variances and cross-correlations. The expansion of the expectation in Inequality \eqref{equ:exp_sum_sqrt} can be expressed as:
\begin{equation}
  \label{equ:exp_sum_diag_corss_split}
  \begin{aligned}
    \mathbb{E}_{\mathcal{V}} \left[ \left\| \sum_{i=1}^{n} \left( \tvar{h}_i^{(A)} - \tvar{h}_i^{(B)} \right) \right\|^2_F \right]
     & = \sum_{i=1}^{n} \mathbb{E}_{\mathcal{V}} \left[ \left\| \tvar{h}_i^{(A)} - \tvar{h}_i^{(B)} \right\|_F^2 \right]                                                                      \\
     & \quad + \sum_{i,j=1, i\ne j}^{n} \mathbb{E}_{\mathcal{V}} \left[ \left( \tvar{h}_i^{(A)} - \tvar{h}_i^{(B)} \right)^{\top} \left( \tvar{h}_j^{(A)} - \tvar{h}_j^{(B)} \right) \right]. \\
  \end{aligned}
\end{equation}
The analysis of expected divergence should carefully consider the dependencies inherent in transformer-based sequence models. These dependencies manifest in two distinct but interconnected ways.

\textbf{Intra-sequence dependencies:} For a sequence model, the hidden representation at position \(k+i\) is computed based on all preceding tokens from position \(0\) to \(i\). This autoregressive nature creates strong dependencies within each sequence, where each hidden representation relates to all previous hidden representations. Consequently, \(\tvar{h}_{i}^{(A)}\) should have some dependency on \(\tvar{h}_{j}^{(A)}\) for \(1 \le j < i\). This temporal dependency structure means that the hidden representations in a sequence cannot be treated as independent samples, and their statistical relationships must be explicitly modeled in our analysis.

\textbf{Inter-branch dependencies:} When processing branching sequences that share the same initial sequence, the hidden representations in two branches are likely to exhibit a degree of commonality, especially when the length of the branching sequence is relatively short. This cross-branch dependency arises from three primary sources: First, both branches share the same initial context \(\{\tvar{x}_{1}, \ldots, \tvar{x}_{k}\}\), ensuring that the hidden representations generated in the branches are influenced by the same underlying information. Second, both branches are processed by the same sequence model, which applies its knowledge and understanding uniformly across both computational pathways. Third, when the branching sequences \(\{\tvar{a}_{k+1}, \ldots, \tvar{a}_{k+i}\}\) and \(\{\tvar{b}_{k+1}, \ldots, \tvar{b}_{k+i}\}\) are short, the new information introduced is relatively limited, which further reduces the divergence between the hidden representations. Consequently, \(\tvar{h}_{i}^{(A)}\) and \(\tvar{h}_{i}^{(B)}\) exhibit correlation that decreases as \(i\) increases, but remains significant for small values of \(i\).

Considering these dependencies, the right-hand side (RHS) of Equation \eqref{equ:exp_sum_diag_corss_split} becomes,
\begin{equation}
  \label{equ:E_diff_hijAB_expand_cos_M}
  \begin{aligned}
    \text{\eqref{equ:exp_sum_diag_corss_split} RHS}
     & \leq 2 M^2 \sum_{i=1}^{n} \left(
    1 - \mathbb{E}_{\mathcal{V}} \left[
      \cos\left( \Theta_{ii}^{(AB)} \right)
      \right]
    \right)                                     \\
     & ~ + M^2 \sum_{i,j=1, i \ne j}^{n} \left(
    \mathbb{E}_{\mathcal{V}}
    \left[ \cos\left( \Theta_{ij}^{(AA)} \right) \right]
    + \mathbb{E}_{\mathcal{V}}
    \left[ \cos\left( \Theta_{ij}^{(BB)} \right) \right]
    - 2 \mathbb{E}_{\mathcal{V}}
    \left[ \cos\left( \Theta_{ij}^{(AB)} \right) \right]
    \right),                                    \\
  \end{aligned}
\end{equation}
where the scalar \(M\) is defined as the largest value of the norm on arbitrary hidden states:
\begin{equation}
  M = \max\left(
  \mathbb{E}_{\mathcal{V}} \left[
  \left\| \tvar{h}_i^{(A)} \right\|_F
  \right],
  \mathbb{E}_{\mathcal{V}} \left[
  \left\| \tvar{h}_i^{(B)} \right\|_F
  \right]
  \right), ~~ \text{for all } i \in \{1, \ldots, n\},
\end{equation}
and \(\Theta\) denotes angles between hidden representations.
Specifically,
\(\Theta_{ii}^{(AB)}\) is the angle between \(\tvar{h}_{i}^{(A)}\) and \(\tvar{h}_{i}^{(B)}\);
while \(\Theta_{ij}^{(AA)}\) and \(\Theta_{ij}^{(BB)}\) correspond to the angles between
\(\tvar{h}_{i}^{(A)}, \tvar{h}_{j}^{(A)}\) and \(\tvar{h}_{i}^{(B)}, \tvar{h}_{j}^{(B)}\) respectively;
\(\Theta_{ij}^{(AB)}\) reflects the angle between \(\tvar{h}_{i}^{(A)}\) and \(\tvar{h}_{j}^{(B)}\).
For detailed derivative, please refer to Appendix \ref{appendix:derivative_divergence_scaling}, and for the distribution of cosines, please refer to Appendix \ref{appendix:empirical-distribution-cosine-similarities}.

To quantify the dependencies, four dependency measures that capture the essential correlation patterns in transformer architectures are defined.
These measures capture the complex interactions between hidden representations into bounded parameters that facilitate theoretical analysis while preserving the fundamental characteristics of sequence processing.
The dependency measures are constructed based on cosine similarities between hidden representations, which naturally capture the directional relationships between vectors in high-dimensional space.

\begin{definition}[Dependency measures]
  \label{def:dependency_measures}
  The dependency measures for sequence length \(n\) are defined as:
  \begin{equation}
    \label{equ:def_psi_ij}
    \begin{aligned}
      \psi_{\tequal}^{(AB)}(n) & = \max\left\{
      0, \sup_{i=1, \ldots, n} \mathbb{E}_{\mathcal{V}} \left[ \cos\left( \Theta_{ii}^{(AB)} \right) \right]
      \right\},                                \\
      \psi_{\tcross}^{(AB)}(n) & = \max\left\{
      0, \sup_{i,j=1, \ldots, n; i \ne j} \mathbb{E}_{\mathcal{V}} \left[ \cos\left( \Theta_{ij}^{(AB)} \right) \right]
      \right\},                                \\
      \psi_{\tcross}^{(AA)}(n) & = \max\left\{
      0, \sup_{i,j=1, \ldots, n; i \ne j} \mathbb{E}_{\mathcal{V}} \left[ \cos\left( \Theta_{ij}^{(AA)} \right) \right]
      \right\},                                \\
      \psi_{\tcross}^{(BB)}(n) & = \max\left\{
      0, \sup_{i,j=1, \ldots, n; i \ne j} \mathbb{E}_{\mathcal{V}} \left[ \cos\left( \Theta_{ij}^{(BB)} \right) \right]
      \right\},                                \\
    \end{aligned}
  \end{equation}
  where:
  \begin{itemize}
    \item \(\psi_{\tequal}^{(AB)}(n)\) measures the same-position correlation between two branches, capturing how similar the representations remain at corresponding positions despite different branches (equal position, cross branches);
    \item \(\psi_{\tcross}^{(AB)}(n)\) quantifies the cross-branch temporal correlation, measuring dependencies between representations at different positions across the two branches (cross positions, cross branches);
    \item \(\psi_{\tcross}^{(AA)}(n), \psi_{\tcross}^{(BB)}(n)\) capture intra-sequence temporal dependencies within each branch, reflecting how similar the representations are at different positions within the same branch (cross positions, equal branch).
  \end{itemize}
\end{definition}

The operator \(\max\{0, \cdot\}\) takes only positive correlations to the dependency bounds, and the operator \(\sup\) is used to capture the maximum correlation value over all possible hidden representations, capturing worst-case correlation scenarios for the theoretical bounds.
Without loss of generality, it can be assumed that \(\psi_{\tcross}^{(AA)}(n) = \psi_{\tcross}^{(BB)}(n)\) due to the symmetric nature of sequence branches, or can be considered as a replacement of parameters \(\psi_{\tcross}^{(AA)}(n) \Leftarrow \frac{1}{2}\left( \psi_{\tcross}^{(AA)}(n) + \psi_{\tcross}^{(BB)}(n) \right)\).

With the dependency measures defined, and considering the symmetry of \(\psi_{\tcross}^{(AA)}(n)\) and \(\psi_{\tcross}^{(BB)}(n)\), right-hand side of Inequality \eqref{equ:E_diff_hijAB_expand_cos_M} becomes,
\begin{equation}
  \label{equ:E_diff_hijAB_bound}
  \begin{aligned}
    \text{\eqref{equ:E_diff_hijAB_expand_cos_M} RHS}
     & \leq 2 M^2 \sum_{i=1}^{n} \left( 1 - \psi_{\tequal}^{(AB)}(n) \right)
    + M^2 \sum_{i,j=1, i \ne j}^{n} \left(
    \psi_{\tcross}^{(AA)}(n)
    + \psi_{\tcross}^{(BB)}(n)
    - 2 \psi_{\tcross}^{(AB)}(n)
    \right)                                                                  \\
     & \leq 2 M^2 \left(
    n \left(1 - \psi_{\tequal}^{(AB)}(n)\right)
    + (n^2-n) \left( \psi_{\tcross}^{(AA)}(n) - \psi_{\tcross}^{(AB)}(n) \right)
    \right).
  \end{aligned}
\end{equation}
For complete version of the derivative, please refer to Appendix \ref{appendix:derivative_divergence_scaling}.
To simply this notation, a shorthand function \(\Upsilon(n)\) is introduced:
\begin{equation}
  \label{equ:def_psi_n_big}
  \Upsilon(n) := n \left(1 - \psi_{\tequal}^{(AB)}(n)\right)
  + (n^2-n) \left( \psi_{\tcross}^{(AA)}(n) - \psi_{\tcross}^{(AB)}(n) \right).
\end{equation}
Recalling Equation \eqref{equ:expected_distance} and Inequality \eqref{equ:exp_sum_sqrt}, the expected divergence after \(n\) differing tokens is bounded by:
\begin{equation}
  \label{equ:expected_divergence_bound_sqrt}
  \mathbb{E}_{\mathcal{V}} \left[ \mathcal{D}(n) \right]
  = \bigO{\sqrt{
      n \left(1 - \psi_{\tequal}^{(AB)}(n)\right)
      + (n^2-n) \left( \psi_{\tcross}^{(AA)}(n) - \psi_{\tcross}^{(AB)}(n) \right)
    }}
  := \bigO{\sqrt{\Upsilon(n)}},
\end{equation}
where \(\bigO{\cdot}\) represent the big \(\mathcal{O}\) notation.

\subsection{Divergence Scaling Properties}
\label{subsec:dtsa-properties-divergence-scaling}

The asymptotic behavior of sequence divergence is critically determined by the relative magnitudes of the dependency measures \(\psi_{\tequal}^{(AB)}(n)\), \(\psi_{\tcross}^{(AA)}(n)\), and \(\psi_{\tcross}^{(AB)}(n)\).
To facilitate an understanding of these scaling properties, the analysis is first conducted under the simplified assumption that the dependency measures remain invariant with respect to sequence length. For analytical tractability, the discrete sequence length \(n\) is approximated as a continuous variable in the subsequent derivation.

Consider the simplified scenario where all dependency measures are constants independent of \(n\): \(\psi_{\tequal}^{(AB)}(n) \rightarrow \psi_{\tequal}^{(AB)}\), \(\psi_{\tcross}^{(AA)}(n) \rightarrow \psi_{\tcross}^{(AA)}\), and \(\psi_{\tcross}^{(AB)}(n) \rightarrow \psi_{\tcross}^{(AB)}\).
In the following, such symbol with \((n)\) stands for its function version, and that without \((n)\) stands for its constant version.
From Inequality \eqref{equ:expected_divergence_bound_sqrt}, the argument of the square root can be decomposed into two components:
\begin{equation}
  \label{equ:expected_divergence_bound_linear_and_quadratic}
  \begin{aligned}
    \text{Linear:} \quad    & n \left( 1 - \psi_{\tequal}^{(AB)} - \psi_{\tcross}^{(AA)} + \psi_{\tcross}^{(AB)} \right), \\
    \text{Quadratic:} \quad & n^2 \left( \psi_{\tcross}^{(AA)} - \psi_{\tcross}^{(AB)} \right).
  \end{aligned}
\end{equation}

\textbf{Asymptotic behaviors shift as \(n\) change.}
The asymptotic scaling behavior is determined by which component dominates, leading to fundamentally different growth patterns for sequence divergence.
If \(\psi_{\tcross}^{(AA)} = \psi_{\tcross}^{(AB)}\), the quadratic component vanishes entirely, yielding only the linear component and leading to a sublinear scaling \(\bigO{\sqrt{n}}\).
Otherwise, if \(\psi_{\tcross}^{(AA)} - \psi_{\tcross}^{(AB)} > 0\), a ratio \(\rho(n)\) of the linear coefficient to the quadratic coefficient is defined to better characterize witch component dominates,
\begin{equation}
  \label{equ:ratio_quadratic_linear}
  \rho(n)
  = \frac{
    1 - \psi_{\tequal}^{(AB)} - \psi_{\tcross}^{(AA)} + \psi_{\tcross}^{(AB)}
  }{
    n \left( \psi_{\tcross}^{(AA)} - \psi_{\tcross}^{(AB)} \right)
  }.
\end{equation}
The unusual case \(\psi_{\tcross}^{(AA)} - \psi_{\tcross}^{(AB)} < 0\) is excluded here, since the quadratic component would result in a negative value that violates the positive constrains of the bound (please refer to Appendix \ref{appendix:dependency_constraints} for detailed justification).
A critical balance point can be reached at \(\rho(n) = 1\) when the linear and quadratic components contribute equally, yielding,
\begin{equation}
  \label{equ:expected_divergence_bound_sqrt_balance}
  \left. n \right|_{\rho(n) = 1} = \frac{
    1 - \psi_{\tequal}^{(AB)}
  }{
    \psi_{\tcross}^{(AA)} - \psi_{\tcross}^{(AB)}
  } - 1.
\end{equation}
By setting \(\rho(n) > 1\), and \(\rho(n) < 1\), the divergence bound exhibits different scaling behaviors,
\begin{equation}
  \label{equ:quadratic_term_dominance}
  \bigO{\sqrt{n}} \Leftarrow n < \left. n \right|_{\rho(n) = 1},
  \qquad
  n > \left. n \right|_{\rho(n) = 1} \Rightarrow \bigO{n}.
\end{equation}
When \(n\) is small, the linear component dominates and the divergence bound approaches to a sublinear scaling \(\bigO{\sqrt{n}}\), indicating that sequence representations diverge slowly as the number of differing tokens increases. Conversely, when \(n\) is large, the quadratic component dominates and the bound degrades to a linear scaling \(\bigO{n}\), suggesting rapid divergence growth.
Under the linear scaling regime, a desire sequence divergence can be reached at a shorter sequence length compared to the quadratic scaling regime.

The transition between these scaling regimes can be understood through the lens of information availability and model confidence.
When \(n\) is small, the distinct sequence branches are short, providing limited differential information between the sequences. Under these conditions, the model exhibits reduced confidence in distinguishing between the inputs, resulting in the sublinear scaling behavior \(\bigO{\sqrt{n}}\) where divergence grows slowly.
As \(n\) increases, the distinct sequence branches become longer, offering substantially more differential information. With access to this enriched information, the model gains increased capability to differentiate between the inputs, leading to the linear scaling regime \(\bigO{n}\) where divergence exhibits rapid growth.

\textbf{Influence of \(\psi_{\tcross}^{(AA)} - \psi_{\tcross}^{(AB)}\) on divergence scaling.} The quadratic component in Equation \eqref{equ:expected_divergence_bound_linear_and_quadratic} is directly determined by the difference \(\psi_{\tcross}^{(AA)} - \psi_{\tcross}^{(AB)}\).
When this difference is large, the critical balance point \(\left. n \right|_{\rho(n) = 1}\) from Equation \eqref{equ:expected_divergence_bound_sqrt_balance} becomes smaller, resulting in a shorter sequence length required to achieve the same divergence threshold.
Conversely, when this difference is small, \(\left. n \right|_{\rho(n) = 1}\) increases, necessitating a longer sequence length to reach the same divergence threshold.

In practical terms, \(\psi_{\tcross}^{(AA)}\) quantifies the temporal correlation within a single sequence branch, while \(\psi_{\tcross}^{(AB)}\) measures the temporal correlation between two distinct sequence branches.
The difference \(\psi_{\tcross}^{(AA)} - \psi_{\tcross}^{(AB)}\) can be interpreted as the relative strength of intra-branch dependencies compared to cross-branch dependencies.
When this difference is large, the hidden representations of the two sequence branches exhibit substantial dissimilarity, enabling the model to distinguish between them with greater ease. Consequently, fewer tokens are required to reach the desired divergence threshold.

Conversely, when this difference is small, the hidden representations of the two sequence branches remain highly similar, making it more challenging for the model to differentiate between them.
Under such conditions, a longer sequence length is necessary to achieve the desired divergence threshold.

\begin{figure}[htb]
  \centering
  \begin{subfigure}{0.56\textwidth}
    \centering
    \includegraphics{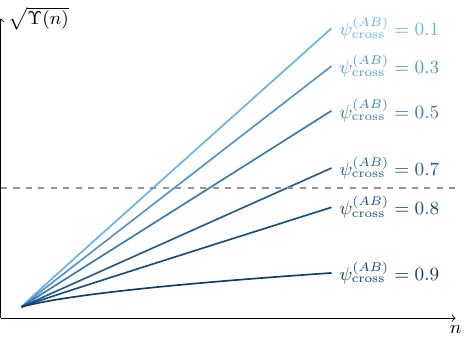}
    \caption{}
    \label{fig:divergence_bound_wrt_psi_cross_ab}
  \end{subfigure}
  \hfill
  \begin{subfigure}{0.43\textwidth}
    \centering
    \includegraphics{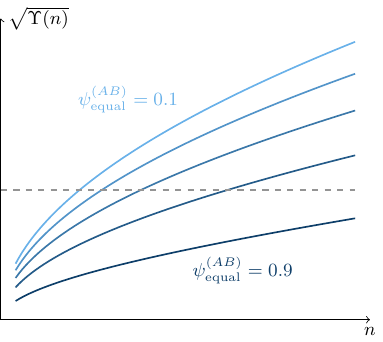}
    \caption{}
    \label{fig:divergence_bound_wrt_psi_equal_ab}
  \end{subfigure}
  \caption{
    (\subref{fig:divergence_bound_wrt_psi_cross_ab}) \(\Upsilon(n)\) with respect to \(\psi_{\tcross}^{(AB)}\) with fixed \(\psi_{\tequal}^{(AB)}=0.7\) and \(\psi_{\tcross}^{(AA)} = 0.9\). Lower values of \(\psi_{\tcross}^{(AB)}\) lead to faster divergence growth, enabling threshold crossing at shorter sequence lengths.
    (\subref{fig:divergence_bound_wrt_psi_equal_ab}) \(\Upsilon(n)\) with respect to \(\psi_{\tequal}^{(AB)}\) with fixed \(\psi_{\tcross}^{(AB)}=0.899\) and \(\psi_{\tcross}^{(AA)}=0.900\). Higher values of \(\psi_{\tequal}^{(AB)}\) lead to smaller values of \(\Upsilon(n)\).
  }
\end{figure}

To provide an illustration of this relationship, \(\psi_{\tcross}^{(AA)}\) and \(\psi_{\tequal}^{(AB)}\) are held constant while \(\psi_{\tcross}^{(AB)}\) is varied systematically, as demonstrated in Figure \ref{fig:divergence_bound_wrt_psi_cross_ab}.
The results align with the theoretical analysis: when \(\psi_{\tcross}^{(AB)}\) is small, the difference \(\psi_{\tcross}^{(AA)} - \psi_{\tcross}^{(AB)}\) becomes large, and the figure confirms that a smaller value of \(n\) is sufficient to reach the desired threshold.

\textbf{Influence of \(\psi_{\tequal}^{(AB)}\) on divergence scaling.}
From Equation \eqref{equ:expected_divergence_bound_linear_and_quadratic}, the linear component is governed by the term \(1 - \psi_{\tequal}^{(AB)}\). When this difference is large, the critical balance point \(\left. n \right|_{\rho(n)=1}\) from Equation \eqref{equ:expected_divergence_bound_sqrt_balance} increases accordingly.

The influence of \(\psi_{\tequal}^{(AB)}\) on the scaling behavior exhibits two distinct characteristics.
First, the parameter has limited impact on the balance point since \(1 - \psi_{\tequal}^{(AB)}\) appears in the numerator of \(\left. n \right|_{\rho(n)=1}\), making its influence weaker than that of the quadratic coefficient \(\psi_{\tcross}^{(AA)} - \psi_{\tcross}^{(AB)}\) in the denominator. The balance point varies linearly with \(\psi_{\tequal}^{(AB)}\), and since \(\psi_{\tequal}^{(AB)}\) is constrained to the interval \([0, 1]\), its impact on \(\left. n \right|_{\rho(n)=1}\) remains bounded.
Second, the parameter exerts direct influence on the linear component by primarily affecting the linear term in \(\Upsilon(n)\). When \(\psi_{\tequal}^{(AB)}\) is smaller, the coefficient \(1 - \psi_{\tequal}^{(AB)}\) becomes larger, resulting in a larger linear component and consequently faster approach to the desired divergence threshold.

The practical impact of \(\psi_{\tequal}^{(AB)}\) depends critically on which scaling regime dominates. When the quadratic term is dominant (large \(n\)), the influence of \(\psi_{\tequal}^{(AB)}\) becomes negligible.
However, when the linear term dominates (small \(n\)), \(\psi_{\tequal}^{(AB)}\) exerts substantial influence on the divergence scaling behavior.

In practical applications, \(\psi_{\tequal}^{(AB)}\) quantifies the similarity between hidden representations of two distinct sequence branches at corresponding positions in the input sequences.
When this value is large, the representations at corresponding positions across branches remain highly similar, making it more challenging for the model to differentiate between the sequences.
Consequently, during the initial processing phase (small \(n\)), when \(\psi_{\tequal}^{(AB)}\) is small, the model can readily distinguish between the two branches from the outset.
This early discrimination capability enables the model to reach the desired divergence threshold more rapidly.

Figure \ref{fig:divergence_bound_wrt_psi_equal_ab} illustrates this relationship. As \(\psi_{\tequal}^{(AB)}\) decreases, the divergence threshold is reached more quickly for a given sequence length.

\textbf{Connection to scaling behavior.}
Let \(S(n)\) denote the model's performance on a given task after processing \(n\) differing tokens. The performance is hypothesized to be proportional to the expected divergence:
\begin{equation}
  \label{equ:accuracy-divergence}
  S(n)
  \propto \left( \mathbb{E}_{\mathcal{V}}\left[\mathcal{D}(n)\right] \right)^{\beta}
  \propto \left(\sqrt{ \Upsilon(n) }\right)^{\beta},
\end{equation}
where \(\beta > 0\) is a constant that characterizes the relationship between performance and divergence. For instance, when \(\beta = 1\), the performance \(S(n)\) scales linearly with \(\mathbb{E}_{\mathcal{V}}\left[\mathcal{D}(n)\right]\).

\begin{figure}[tbp]
  \centering
  \includegraphics{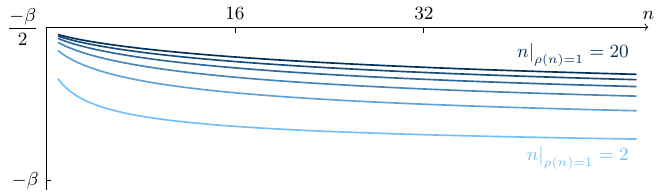}
  \caption{\(\alpha(n)\) with \(n>1\) and \(\psi_{\tcross}^{(AA)} - \psi_{\tcross}^{(AB)} > 0\).}
  \label{fig:alpha_n_psi_cross_aa_ab_positive}
\end{figure}

Following the scaling formulation introduced in \citet{kaplan2020scalinglaw,hoffmann2022ChinchillaScaling}, the relationship between performance and sequence length is considered to be
\begin{equation}
  \label{equ:scaling_accuracy-sequence-length}
  S(n) \approx \dfrac{c}{n^{\alpha(n)}},
\end{equation}
where \(c\) is the scaling constant, and \(\alpha(n)\) is the scaling exponent.
Consequently, it is natural to establish the scaling relationship
\begin{equation}
  \gamma \Upsilon(n)^{\beta/2} = \frac{c}{n^{\alpha(n)}},
\end{equation}
which connects the theoretical divergence bound to empirical scaling behavior.
The value \(\gamma\) is the scaling factor for matching \(S(n)\) and \(\Upsilon(n)^{\beta/2}\).
With the constant setting, three main results are summarized as below (Please refer to Appendix \ref{appendix:scaling_exponent} for the detailed discussion).
The normalization constant can be set as
\begin{equation}
  c = \gamma\left(1 - \psi_{\tequal}^{(AB)}\right)^{\frac{\beta}{2}}.
\end{equation}
When \(n>1\) and \(\psi_{\tcross}^{(AA)} - \psi_{\tcross}^{(AB)} = 0\), the scaling exponent is given by,
\begin{equation}
  \label{equ:addition_scaling_psi_corss_aa_ab_zero}
  \alpha(n) = - \dfrac{\beta}{2}.
\end{equation}
When \(n>1\) and \(\psi_{\tcross}^{(AA)} - \psi_{\tcross}^{(AB)} > 0\), the scaling exponent becomes,
\begin{equation}
  \label{equ:addition_scaling_psi_corss_aa_ab_positive}
  \alpha(n) = \dfrac{\beta}{2} \log_{n} \left(
    \dfrac{
      1 + \left. n \right|_{\rho(n) = 1}
      }{
      n + \left. n \right|_{\rho(n) = 1}
      }
    \right)
    - \frac{\beta}{2}.
\end{equation}
This \(\alpha(n)\) is a monotonically decreasing function with respect to \(n\) and will approach to \(-\beta\) as \(n\) increases.
In order to provide a more intuitive understanding, the curves of \(\alpha(n)\) under different values of \(\left. n \right|_{\rho(n) = 1}\) are shown in Figure \ref{fig:alpha_n_psi_cross_aa_ab_positive}, which shows the slope of the curves becomes stable as \(n\) increases.
Therefore, it is reasonable to simplify the expressions as a constant value \(\alpha(n) \rightarrow \alpha\) in the following estimations and evaluations. 

Smaller values of \(\alpha(n)\) (corresponding to larger values of \(|\alpha(n)|\)) lead to faster growth in performance \(S(n)\).
Under the condition \(n > 1\) and \(\psi_{\tcross}^{(AA)} - \psi_{\tcross}^{(AB)} > 0\), for a fixed sequence length \(n\), a smaller balance point \(\left. n \right|_{\rho(n) = 1}\) results in a correspondingly smaller value of \(\alpha(n)\).
This relationship also aligns with the observations from the upper bound \(\bigO{\sqrt{\Upsilon(n)}}\). A smaller balance point indicates that fewer tokens \(n\) are required to transition into the linear scaling regime \(\bigO{n}\), such that fewer tokens \(n\) can lead to a larger bound.

These expressions reveal how the scaling exponent \(\alpha(n)\) depends on the dependency measures, providing a direct link between the theoretical analysis of sequence divergence and the empirical observations of model performance scaling. 

\subsection{Empirical Estimation}
\label{subsec:dtsa-empirical-estimation}

The theoretical analysis of divergence between branching sequences has been established in the preceding subsections.
This subsection provides empirical validation of the theoretical framework through the estimation of the expected divergence \(\mathbb{E}_{\mathcal{V}}\left[\mathcal{D}(n)\right]\) and the dependency measures \(\psi_{\tequal}^{(AB)}(n)\), \(\psi_{\tcross}^{(AA)}(n)\), and \(\psi_{\tcross}^{(AB)}(n)\) on pretrained large language models using a dataset of paired sequences.
Through this empirical analysis, the validity of the theoretical predictions is demonstrated.

\textbf{Dataset construction.} To facilitate empirical estimation, a dataset of paired input sequences that align with our theoretical setting is required. Each pair consists of sequences that share an identical prefix but diverge in their vision-referencing components. Specifically, each sequence comprises three components: a system prompt, a query, and an answer. Within each pair, the system prompt and query remain identical (forming the shared prefix), while the answers differ, creating the branching sequences that correspond to branches A and B in the theoretical analysis.

The hidden representations are computed specifically based on the `answer' component of the paired sequences. This approach enables focused analysis of how representational divergence propagates through the model's computational pathways as the branching sequences evolve.
The dataset for Direct Preference Optimization (DPO) \citep{rafailov2024direct} matches the requirements perfectly.
For this empirical analysis, dataset `ocra-dpo-pairs' \footnote{\url{https://huggingface.co/datasets/Intel/orca_dpo_pairs}} is utilized. This dataset contains the necessary paired sequences with the required structure for our estimation procedure.
The `chosen' answer is the computed as branching sequence A, and the `rejected' answer is the computed as branching sequence B.

\textbf{Computational procedure.} The empirical estimation follows a systematic four-step process:
\begin{enumerate}
  \item \textit{Tokenization.} The positive and negative sequences are converted to input token sequences \(\{\tvar{x}_1, \ldots, \tvar{x}_{k}, \tvar{a}_{k+1}, \ldots, \tvar{a}_{k+n}\}\) and \(\{\tvar{x}_1, \ldots, \tvar{x}_{k}, \tvar{b}_{k+1}, \ldots, \tvar{b}_{k+n}\}\), where \(k\) represents the length of the shared prefix.

  \item \textit{Hidden representation extraction.} The input tokens are processed through the LLM (excluding the final output layer) to obtain the hidden representations:
        \begin{equation*}
          \begin{aligned}
            \tvar{h}_i^{(A)}
             & = \text{LLM}(\tvar{x}_1, \ldots, \tvar{x}_k, \tvar{a}_{k+1}, \ldots, \tvar{a}_{k+i})[-1], \\
            \tvar{h}_i^{(B)}
             & = \text{LLM}(\tvar{x}_1, \ldots, \tvar{x}_k, \tvar{b}_{k+1}, \ldots, \tvar{b}_{k+i})[-1].
          \end{aligned}
        \end{equation*}

  \item \textit{Statistical estimation.} The empirical estimates are computed across multiple data samples to ensure statistical reliability and robustness of the results.
\end{enumerate}

\textit{Empirical estimation of \(\mathbb{E}_{\mathcal{V}}\left[\mathcal{D}(n)\right]\).}
Following the definition in Equation~\eqref{equ:expected_distance}, the expected divergence is computed as the expected Frobenius norm of the cumulative hidden representation differences. The estimation is performed by systematically enumerating each sequence length \(n\) and computing the corresponding average Frobenius norm across the vocabulary distribution. For instance, the expected divergences for \(n=1\) and \(n=2\) are estimated as:
\begin{equation*}
  \mathbb{E}_{\mathcal{V}}\left[\mathcal{D}(1)\right] \Leftarrow \mathbb{E}_{\mathcal{V}}\left[\left\| \tvar{h}_{1}^{(A)} - \tvar{h}_{1}^{(B)} \right\|_F \right], 
  \mathbb{E}_{\mathcal{V}}\left[\mathcal{D}(2)\right] \Leftarrow \mathbb{E}_{\mathcal{V}}\left[
    \left\| \tvar{h}_{1}^{(A)} - \tvar{h}_{1}^{(B)} \right\|_F
    + \left\| \tvar{h}_{2}^{(A)} - \tvar{h}_{2}^{(B)} \right\|_F
  \right].
\end{equation*}

For the expectation \(\mathbb{E}_{\mathcal{V}}\), it is considered as an empirical average,
\begin{equation*}
  \mathbb{E}_{\mathcal{V}}\left[\mathcal{D}(n)\right] \Leftarrow \dfrac{1}{N} \sum_{i=1}^{N} \mathcal{D}_i(n),
\end{equation*}
where \(N\) is the number of samples and \(\mathcal{D}_i(n)\) is the \(i\)-th sample of the expected divergence for sequence length \(n\).

\textit{Empirical estimation of dependency measures.} The four dependency measures introduced in Definition \ref{def:dependency_measures} require empirical estimation from the data. The same-position and cross-branch temporal correlations are computed as:
\begin{equation}
  \psi_{\tequal}^{(AB)}(n) 
  \Leftarrow \mathbb{E}_{\mathcal{V}; i=1, \ldots, n} \cos\left(\Theta_{ii}^{(AB)}\right),
  \quad
  \psi_{\tcross}^{(AB)}(n)
  \Leftarrow \mathbb{E}_{\mathcal{V}; i,j=1, \ldots, n; i \ne j} \cos\left(\Theta_{ij}^{(AB)}\right). 
\end{equation}

Due to the inherent structure of the DPO dataset, where sequences are organized into distinct preference pairs, all `chosen' responses are assigned to branch A, and all `rejected' responses are assigned to branch B. This assignment introduces asymmetry between the sequence branches, violating the symmetry assumption \(\psi_{\tcross}^{(AA)}(n) = \psi_{\tcross}^{(BB)}(n)\) from the theoretical framework.

To address this asymmetry and obtain unbiased estimates, the intra-branch temporal correlations are computed separately for each branch:
\begin{equation}
  \psi_{\tcross}^{(++)}(n) \Leftarrow \mathbb{E}_{\mathcal{V}; i,j=1, \ldots, n; i \ne j} \cos\left(\Theta_{ij}^{(AA)}\right),
  \quad
  \psi_{\tcross}^{(--)}(n) \Leftarrow \mathbb{E}_{\mathcal{V}; i,j=1, \ldots, n; i \ne j} \cos\left(\Theta_{ij}^{(BB)}\right).
\end{equation}
The final symmetric estimate is obtained by averaging the two branch-specific correlations:
\begin{equation}
  \psi_{\tcross}^{(AA)}(n) = \psi_{\tcross}^{(BB)}(n) \Leftarrow \dfrac{1}{2} \left( \psi_{\tcross}^{(++)}(n) + \psi_{\tcross}^{(--)}(n) \right).
\end{equation}
This averaging procedure ensures that the theoretical symmetry assumption is preserved in the empirical analysis while accounting for the structural characteristics of the preference-based dataset.

\textit{Validation of the theoretical bound.}
Equation~\eqref{equ:expected_divergence_bound_sqrt} establishes an upper bound for the expected divergence in terms of the dependency measures. To validate the theoretical bound against empirical observations, a scaling factor \(\lambda\) is introduced such that the expected divergence can be approximated as:
\begin{equation}
  \mathbb{E}_{\mathcal{V}}\left[\mathcal{D}(n)\right] \approx \lambda \sqrt{\Upsilon(n)}.
\end{equation}
Given that all necessary dependency measures have been empirically estimated, the function \(\Upsilon(n)\) can be computed directly using the estimated parameters:
\begin{equation}
  \Upsilon(n) \Leftarrow 
  n \left(1 - \psi_{\tequal}^{(AB)}(n)\right) + (n^2 - n) \left(\psi_{\tcross}^{(AA)}(n) - \psi_{\tcross}^{(AB)}(n)\right).
\end{equation}
The optimal scaling factor \(\lambda\) is determined by minimizing the squared error between the empirical divergence and the theoretical prediction:
\begin{equation}
  \label{equ:empirical-estimation-lambda}
  \lambda = \argmin_{\lambda} \sum_{n} \left(\mathbb{E}_{\mathcal{V}}\left[ \mathcal{D}(n) \right] - \lambda \sqrt{\Upsilon(n)} \right)^2.
\end{equation}

\textbf{Empirical results.} The empirical estimations are taken under various of LLMs, including Vicuna-v1.5-7b \citep{vicuna2023}, Phi-3-mini \citep{abdin2024phi-3}, Llama2-7b \citep{touvron2023llama-2}, Llama-3-8B \citep{dubey2024llama-3}, and Qwen series \citep{qwen}.

\begin{figure}[htbp]
  \centering
  \includegraphics[width=0.99\linewidth]{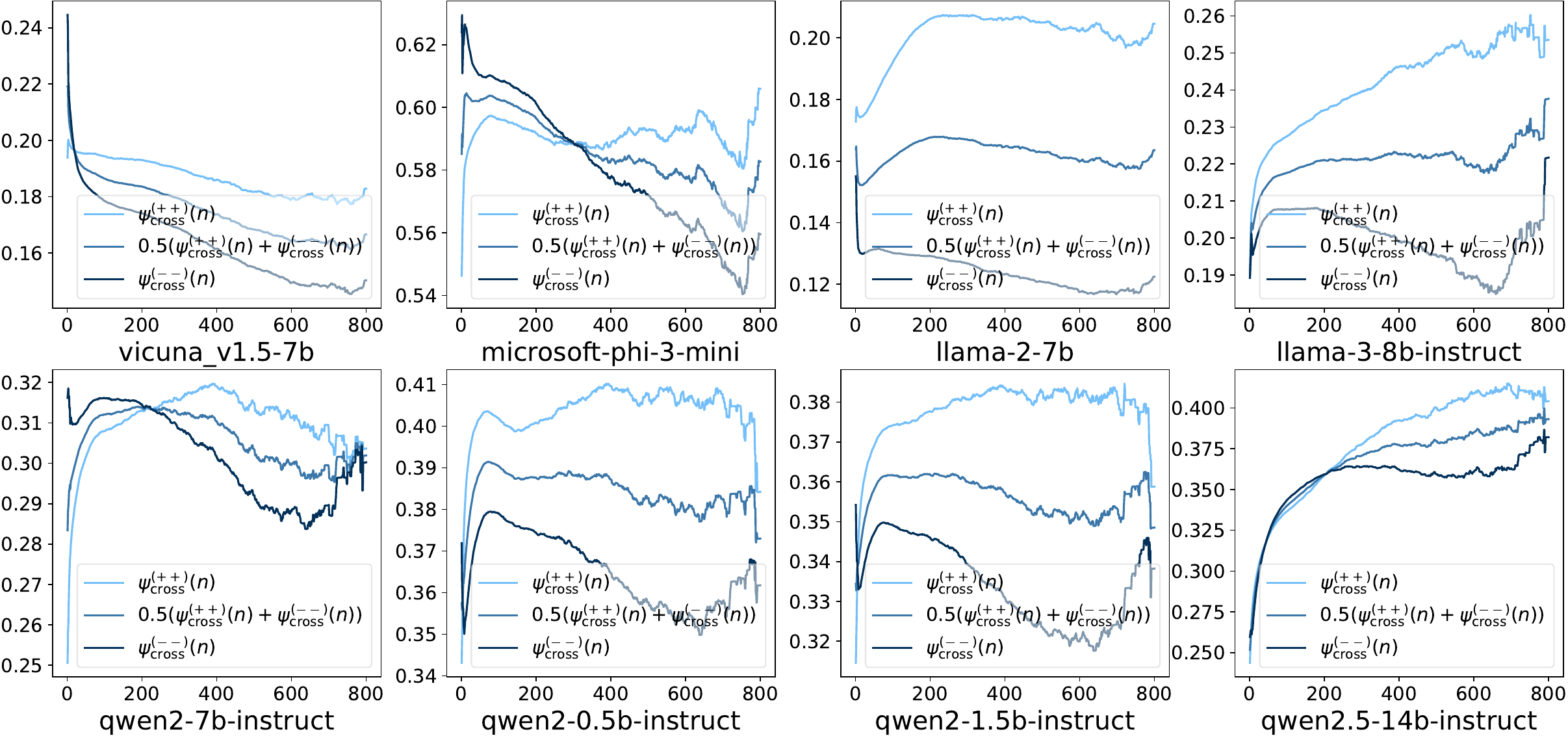}
  \caption{Estimated dependency measures: \(\psi_{\tcross}^{(AA)}(n) = \frac{1}{2}\left(\psi_{\tcross}^{(++)}(n) + \psi_{\tcross}^{(--)}(n)\right)\), where \(\psi_{\tcross}^{(++)}(n)\) corresponds to branches with `chosen' responses and \(\psi_{\tcross}^{(--)}(n)\) corresponds to branches with `rejected' responses.}
  \label{fig:compare_psi_aa_bb_aabb}
\end{figure}

Figure \ref{fig:compare_psi_aa_bb_aabb} presents the estimated dependency measures decomposed by response type.
The measures \(\psi_{\tcross}^{(++)}(n)\) and \(\psi_{\tcross}^{(--)}(n)\) exhibit similar trends with respect to sequence length \(n\), while their specific values differ across response categories.
Different base models demonstrate distinct patterns for \(\psi_{\tcross}^{(++)}(n)\) and \(\psi_{\tcross}^{(--)}(n)\), indicating model-specific dependency characteristics.

\begin{figure}[htbp]
  \centering
  \includegraphics[width=0.99\linewidth]{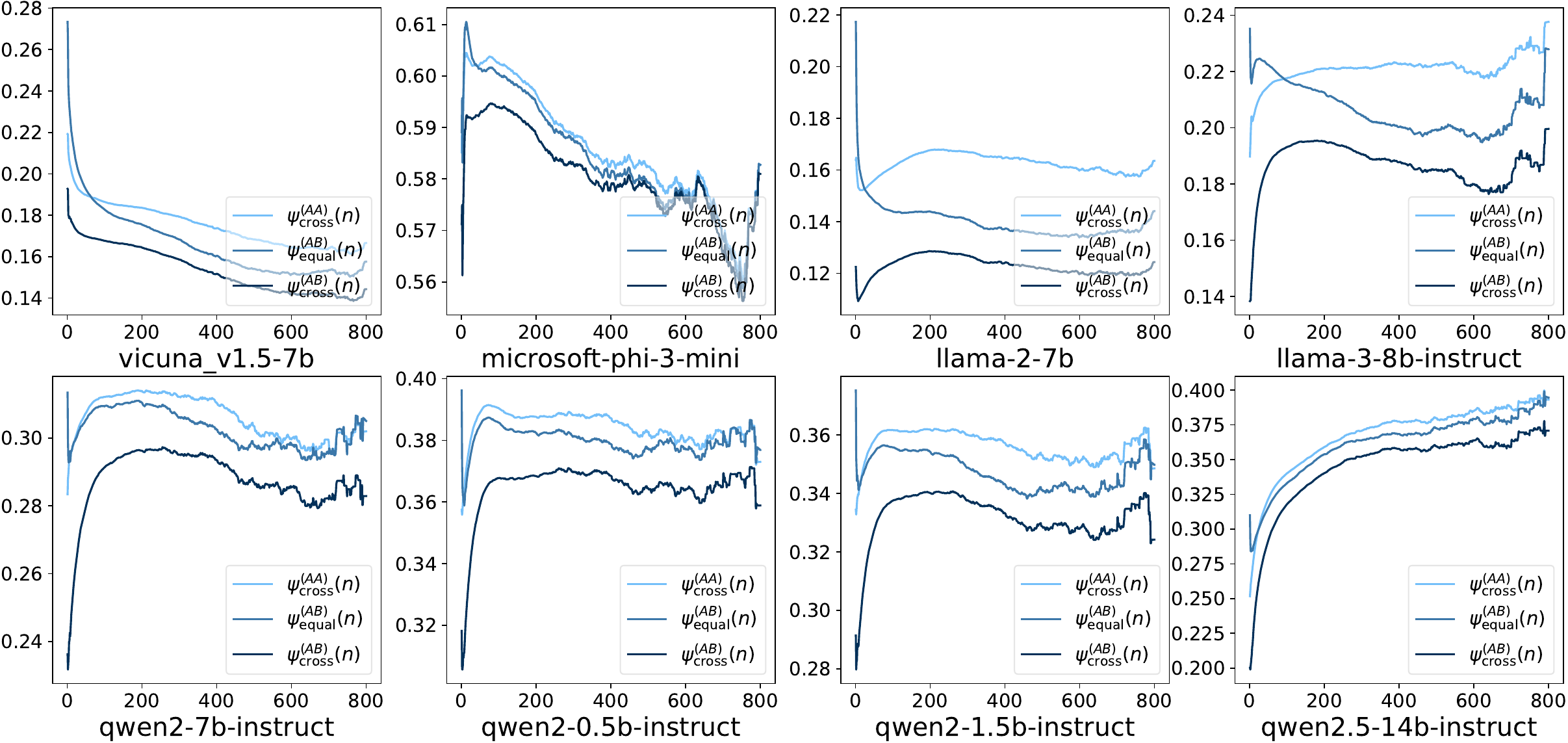}
  \caption{Estimated three dependency measures: \(\psi_{\tcross}^{(AA)}(n)\) (intra-sequence temporal correlation), \(\psi_{\tequal}^{(AB)}(n)\) (same-position cross-branch correlation), and \(\psi_{\tcross}^{(AB)}(n)\) (cross-branch temporal correlation).}
  \label{fig:compare_psi_cross_ab_aabb_equal_ab}
\end{figure}

Figure \ref{fig:compare_psi_cross_ab_aabb_equal_ab} illustrates the estimated values of the three fundamental dependency measures: \(\psi_{\tcross}^{(AA)}(n)\), \(\psi_{\tequal}^{(AB)}(n)\), and \(\psi_{\tcross}^{(AB)}(n)\).
The evolution patterns observed across different base models exhibit similar characteristics, beginning with an initial transient phase and rapidly converging towards a stable regime.
The variation range from minimal to maximal values remains constrained within approximately 0.2 across all base models.
Consequently, these dependency measures can be reasonably approximated as constants for analytical purposes, provided that the approximation error introduced by this simplification is deemed acceptable for the intended analysis.

\begin{figure}
  \centering
  \includegraphics[width=0.99\linewidth]{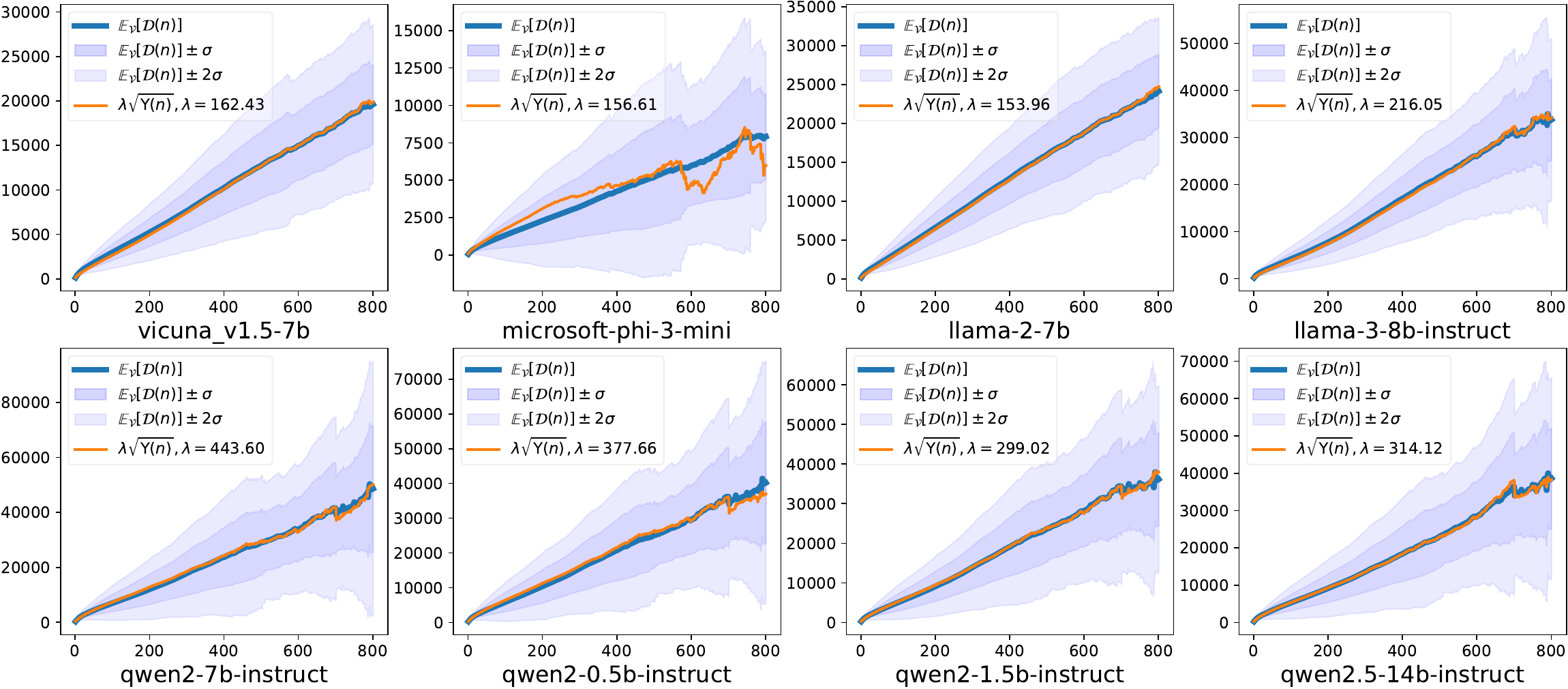}
  \caption{Comparison between empirical observations and theoretical predictions: estimated mean and standard deviation of \(\mathbb{E}_{\mathcal{V}}\left[\mathcal{D}(n)\right]\) alongside computed curves of \(\lambda \sqrt{\Upsilon(n)}\). The close alignment validates the theoretical framework.}
  \label{fig:various_E_mean_std_psi_N}
\end{figure}

With all necessary dependency measures estimated, the function \(\Upsilon(n)\) can be computed directly using Equation \eqref{equ:def_psi_n_big}. Combined with the estimation of \(\mathbb{E}_{\mathcal{V}}\left[\mathcal{D}(n)\right]\) and the scaling parameter \(\lambda\) from Equation \eqref{equ:empirical-estimation-lambda}, the theoretical predictions can be validated against empirical observations, as shown in Figure \ref{fig:various_E_mean_std_psi_N}.
The results demonstrate that the computed curves of \(\lambda \sqrt{\Upsilon(n)}\) closely align with the empirical observations of \(\mathbb{E}_{\mathcal{V}}\left[\mathcal{D}(n)\right]\), providing empirical support for the theoretical framework.

\section{Model Design}
\label{sec:proposed_method}

\subsection{Motivation}
\label{subsec:pmethod-motivation}

Section \ref{sec:div_token_analysis} provides a theoretical insight into vision token scaling behavior.
This section presents a vision language model architecture designed to systematically investigate the relationship between number of vision tokens and model performance, providing a controlled experimental setting for empirical validation.

The theoretical analysis establishes that performance scales as \(S(n) \approx c / n^{\alpha(n)}\), where \(n\) represents the number of differing vision-referencing tokens.
To empirically validate this scaling relationship, three architectural requirements emerge:

\textbf{High-resolution image support.} To provide sufficient visual information for discriminative analysis and enable systematic variation of the number of vision tokens, the model should be able to process high-resolution images that generate rich initial vision sequences. This ensures adequate visual detail while allowing flexible token count selection for scaling experiments.

\textbf{Controllable number of vision tokens.} Systematic investigation of the scaling relationship requires precise control over the number of vision tokens \(n\) fed into the language model.
The model should produce a fixed number of fused vision tokens, enabling controlled experimentation across different token counts.

\textbf{Question-conditioned vision processing.} The uniform input pattern formalization in Equation \eqref{equ:logical_representation} distinguishes between non-referential text tokens and vision-referencing components. This separation suggests that text tokens can guide vision token selection and processing.
By conditioning vision processing on question semantics, it is possible that the model can optimize the vision-referencing component, potentially improving the scaling efficiency.

The resulting architecture implements these principles through a fusion module that processes vision tokens conditioned on text questions, producing controllable numbers of fused tokens for systematic scaling analysis.

\subsection{Model Architecture}
\label{subsec:main_architecture-proposed_method}

\begin{figure}[tb]
  \centering
  \includegraphics[width=0.98\textwidth]{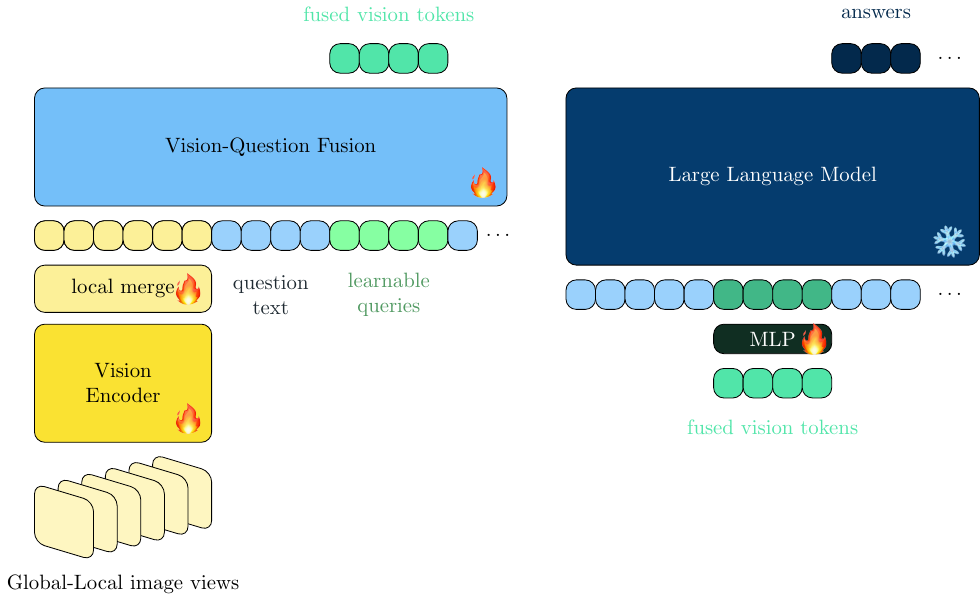}
  \caption{The architecture of the proposed model. Vision tokens are fused with question text and learnable queries before being processed by the LLM. Modules marked with `fire' are updated during fine-tuning, whereas modules marked with `snowflake' remain frozen.}
  \label{fig:model_pipeline}
\end{figure}

The proposed model architecture comprises three primary components: a vision encoder, a fusion module, and a large language model backbone, as illustrated in Figure \ref{fig:model_pipeline}. Each component addresses specific requirements derived from the theoretical analysis while enabling systematic investigation of vision token scaling behavior.

\textbf{Vision encoder.} The vision encoder processes input images to generate vision tokens that encapsulate visual information.
To generate sufficient vision tokens for scaling experiments, the high resolution image support requirement is addressed by employing a dual-view preprocessing strategy, inspired by \citet{dong2024InternLM-xcomposer2-4khd}.
Input images are transformed into one global and several local views: the global view provides holistic context through padding and zooming to match the target resolution, while local views capture fine-grained details by dividing the processed image into smaller patches.
This approach enables the vision encoder pretrained on smaller resolutions to process high resolution images effectively.
For example, the pretrained CLIP ViT-H/14 encoder \citep{radford2021CLIP} can be used to process high resolution content beyond its native \((336, 336)\) resolution limit.

To efficiently manage the large number of vision tokens generated during the global-local view preprocessing, a local merges strategy is employed, wherein multiple vision tokens, such as four neighboring tokens, are aggregated into a single token, thereby reducing the token count by a factor of four and streamlining the subsequent processing stages.
This preprocessing methodology enables the vision encoder to process high resolution images without any adaptations required to its underlying architecture.

\textbf{Fusion module.} The fusion module serves two distinct requirements: controlling the number of vision tokens and integrating vision tokens with text tokens from questions.
For token count control, the module incorporates learnable queries inspired by Q-Former \citep{li2023blip-2} that interact with vision tokens to selectively attend to relevant visual information, distilling a fixed number of output tokens for systematic scaling analysis.
For integration with text questions, the module employs a lightweight decoder-only self-attention transformer architecture following the structure of Llama2 \citep{touvron2023llama-2}, providing a full interaction of vision tokens, question text tokens and the learnable queries.

\textbf{Language model backbone.}
The large language model backbone pretrained exclusively on text data is extended to process multimodal input by concatenating the text tokens with the fused vision tokens from the fusion module.
To ensure proper alignment between the fused vision tokens and the text tokens, a projection layer is applied to match their dimensional compatibility and aligned two modalities.

\textbf{Fine-tuning.}
The fusion module, vision encoder, projection layers, and the large language model backbone are joined to build the main architecture.
During fine-tuning, the fusion module, vision encoder, and some other projection layers are updated to adapt to the integrated multimodal input, enabling the model to generate coherent, context-aware responses.
The large language model backbone remains frozen.
This approach allows for isolating the effects of vision token scaling from changes in language modeling capacity, enabling precise analysis of the scaling relationship established in our theoretical framework.

\subsection{Formats of Input Sequences}
\label{subsec:formats_of_token_sequences}

Two distinct token sequence formats are introduced in this subsection. The first format represents the input to the fusion module, while the second format pertains to the input to the large language model.

\begin{figure}[htb]
  \centering
  \includegraphics{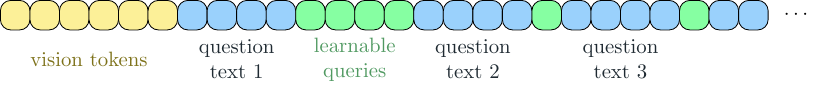}
  \caption{Example of the input token sequence to the fusion module.}
  \label{fig:input_token_sequence_fusion_model}
\end{figure}

\textbf{Input format of fusion module.}
Three distinct tokens are concatenated to form the input to the fusion module.
Vision tokens, generated by the vision encoder, encapsulate visual features extracted from the input image.
Text tokens are derived from the question text and mapped through the embedding layer.
Learnable queries are learnable codebook designed to facilitate selective attention within the fusion module.
Figure \ref{fig:input_token_sequence_fusion_model} illustrates an example of this format.
This format enables efficient processing of multiple question-answer pairs within a single forward pass.

To optimize computational efficiency, the fusion module employs an asymmetric allocation strategy.
The first question is paired with \(n_l\) learnable queries, while subsequent questions are each paired with \(n_s\) learnable queries, where \(n_s < n_l\).
This design addresses two fundamental considerations.
First, questions within a multi-question context often share common visual information that can be encoded once and reused.
The initial \(n_l\) learnable queries capture this shared contextual information and general visual features that benefit all subsequent questions.
The remaining questions, a shorter \(n_s\) query length, only encoding question-specific information.
This design efficiently balances the inclusion of shared and question-specific information while reducing redundancy.
Secondly, this approach significantly reduces the total number of learnable queries, as well as the corresponding fused vision tokens.
For \(K\) questions, using \(n_l\) learnable queries for all questions would result in \(K \times n_l\) learnable queries.
In contrast, by employing \(n_l\) learnable queries after the first question and \(n_s\) learnable queries after the remaining questions, the total number of tokens is reduced to \(n_l + K \times n_s\), where \(n_s < n_l\).
This reduction can enhance computational efficiency and scalability.

\textbf{Input format of language model.}
Each large language model requires its specific input format to ensure effective processing.
To incorporate the fused vision tokens into the LLM input, these tokens are positioned at the beginning of the input sequence for each question, while all other text structures remain unchanged.
This arrangement ensures that the visual information is readily accessible to the model, providing a consistent context for interpreting the question-specific tokens that follow.
An example of this input format is shown in Figure \ref{fig:input_token_sequence_fusion_model}. The lengths of the fused vision tokens follow the same pattern as the learnable queries, with the first question using \(n_l\) tokens and the remaining questions using \(n_s\) tokens.

\begin{figure}[htb]
  \centering
  \includegraphics{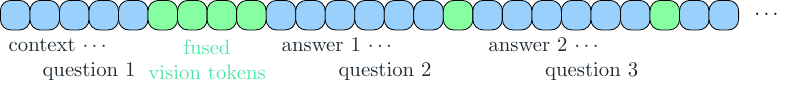}
  \caption{Example of the input token sequence to the fusion module. If needed, the fused vision token would be projected to match the shape of the LLM.}
  \label{fig:input_token_sequence_llm}
\end{figure}

\subsection{Loss Function}
\label{subsec:loss_functions}

The fusion module, being randomly initialized, requires carefully designed training objectives to effectively integrate visual and textual information.
A two-stage training strategy is employed, with contrastive loss for initial alignment and generation loss for answer production.
The training pipeline involves two principal stages. Initially, preliminary training is conducted on the vision encoder, the fused model with the contrastive loss, and subsequently, fine-tuning is performed with the large language model backbone on generation loss.

\textbf{Contrastive loss.}
The contrastive loss ensures semantic alignment between fused vision representations and text representations, following the CLIP framework.
This objective enables the fusion module to produce vision representations that are semantically compatible with text representations from the CLIP text encoder.

The loss operates on \token{eos} tokens, which serve as boundary markers for question-answer groups.
In multi-question scenarios, each question group (comprising questions and learnable queries) terminates with an \token{eos} token, as does each independently encoded answer.
The contrastive loss specifically targets these \token{eos} tokens to enforce alignment between question group representations and their corresponding answers.
For non-\token{eos} tokens within the learnable queries, we apply an orthogonality constraint through identity loss.
This encourages these tokens to encode distinct, complementary information, maximizing their representational diversity and capacity.

Let \(\Omega_{\tokenm{eos}}\) denote the index set of \token{eos} tokens.
For \(i, j \in \Omega_{\tokenm{eos}}\), let \(\tvar{v}_i\) represent the fused vision token corresponding to the \token{eos} token for the \(i\)-th question-answer pair, and let \(\tvar{s}_j\) denote the answer text token associated with the \token{eos} token for the \(j\)-th question-answer pair.
In addition, let \(\overline{\Omega}_{\tokenm{eos}}\) represent the index set of non-\token{eos} tokens within the fused vision tokens. For these tokens, define \(\tvar{Y}\) as a matrix where each row corresponds to a non-\token{eos} token, normalized to have Frobenius norm of one. Formally, each row is defined as \(\tvar{Y}_i = \tvar{y}_i / \|\tvar{y}_i\|_F\), where \(i \in \overline{\Omega}_{\tokenm{eos}}\).

The contrastive loss, \(\mathcal{L}_{\text{cont}}\), is defined as:
\begin{equation}
  \label{equ:loss_contrastive}
  \mathcal{L}_{\text{cont}}
  = \dfrac{-1}{|\Omega_{\tokenm{eos}}|} \sum_{k \in \Omega_{\tokenm{eos}}} \log \left(\dfrac{
    \exp\left( \tvar{v}_k^{\top} \tvar{s}_k \right)
  }{
    \sum_{i,j \in \Omega_{\tokenm{eos}}}
    \exp\left( \tvar{v}_{i}^{\top} \tvar{s}_{j} \right)
  } \right)
  + \dfrac{1}{|\overline{\Omega}_{\tokenm{eos}}|} \left\| \tvar{Y}^{\top} \tvar{Y} - I \right\|_F^2,
\end{equation}
where \(I\) is the identity matrix corresponding to the dimension of \(\tvar{Y}^{\top} \tvar{Y}\).

Once the parameter of the fusion module have been adequately warmed up through preliminary training, the generation loss is introduced to empower the model's capacity to generate accurate answers.

\textbf{Generation loss.}
The generation loss ensures the language model produces contextually appropriate textual responses using standard cross-entropy loss, a fundamental objective in language modeling.

The loss targets answer tokens \token{ans} and boundary \token{eos} tokens that terminate each answer.
Let \(\Omega' = \Omega_{\text{ans}} \cup \Omega_{\text{eos}}\) represent the combined index set of these target tokens.
For each \(i \in \Omega'\), let \(\mathbf{x}_i\) denote the output logits from the language model for the \(i\)-th token, where \(\mathbf{x}_{i, y_i}\) corresponds to the logit for the ground truth token \(y_i\), and \(\mathbf{x}_{i, c}\) represents the logit for candidate token \(c\).
The cross-entropy generation loss is defined as:
\begin{equation}
  \label{equ:loss_generative_ce}
  \mathcal{L}_{\text{ce}}
  = - \frac{1}{|\Omega'|} \sum_{i \in \Omega'} \log \frac{\exp(\mathbf{x}_{i,y_i})}{\sum_c \exp(\mathbf{x}_{i,c})}.
\end{equation}

\section{Experiment}
\label{sec:experiments}

\subsection{Setup}
\label{subsec:exp-experimental_setup}

This section systematically evaluates the proposed model across multiple vision language benchmarks to investigate the relationship between the number of vision tokens and performance, providing empirical support for the scaling behavior established in Section~\ref{sec:div_token_analysis}.

\textbf{Training procedure.} The training methodology follows a two-stage approach designed to isolate the effects of vision token scaling.
Firstly, a base model is trained with configuration \(n_l = 256\) and \(n_s = 8\) on the complete dataset to establish a strong foundation. Secondly, this base model is further fine-tuned using 10\% subsets of the training data with alternative configurations of \(n_l\) and \(n_s\) to systematically investigate scaling behavior while maintaining computational efficiency.

\textbf{Configuration space.} To systematically investigate the scaling relationship, the following configurations are considered: \(768(8)\), \(512(8)\), \(384(8)\), \(256(8)\), \(128(8)\), \(64(8)\), \(32(8)\), \(16(8)\), \(8(8)\), and \(1(1)\), where the first number represents \(n_l\) (number of learnable queries) and the number in parentheses denotes \(n_s\) (number of vision tokens for later questions). This configuration space enables comprehensive analysis of vision token scaling across multiple orders of magnitude.

\textbf{Evaluation framework.} The VLMEvalKit \citep{duan2024vlmevalkit} is employed to ensure the standardization and reproducibility of evaluations across multiple vision language benchmarks, including MME \citep{fu2023mme}, HallusionBench \citep{HallusionBench}, POPE \citep{POPE}, and other benchmarks.

For detailed description of the datasets employed in the experiments, please refer to Appendix~\ref{appendix:datasets}.
Complete implementation details\footnote{The code is available at \url{https://github.com/tenghuilee/ScalingCapFusedVisionLM.git}}, including hyperparameters, training configurations, and computational resources, are provided in Appendix~\ref{appendix:implementation_details}.

\subsection{Analysis of Scaling}
\label{subesc:exp-scaling_analysis}

This section investigates the scaling properties of model performance with respect to the number of vision tokens \(n_l\).
Following the approach proposed in \citep{kaplan2020scalinglaw} and Equation \eqref{equ:scaling_accuracy-sequence-length}, and consider the simplified constant setting, \(\alpha(n) \rightarrow \alpha\), the relationship between the performance metric \(S(n_l)\) and the number of vision tokens \(n_l\) can be expressed as:
\begin{equation}
  \label{equ:scaling-law-token-space}
  S(n_l) \approx \dfrac{c}{n_{l}^{\alpha}}.
\end{equation}
The parameter \(c\) reflects baseline model performance, while \(\alpha\) determines the rate at which performance changes as a function of token count. Smaller \(\alpha\) values indicate slower performance degradation with token reduction, while larger values signify more pronounced impact.
The fitted curves provide guidance for optimizing vision language models by balancing computational resources with desired performance levels.
Detailed visualizations are provided in Appendix \ref{appendix:further-analysis-Nl}.

\begin{figure}[htb]
  \center
  \includegraphics[width=0.999\linewidth]{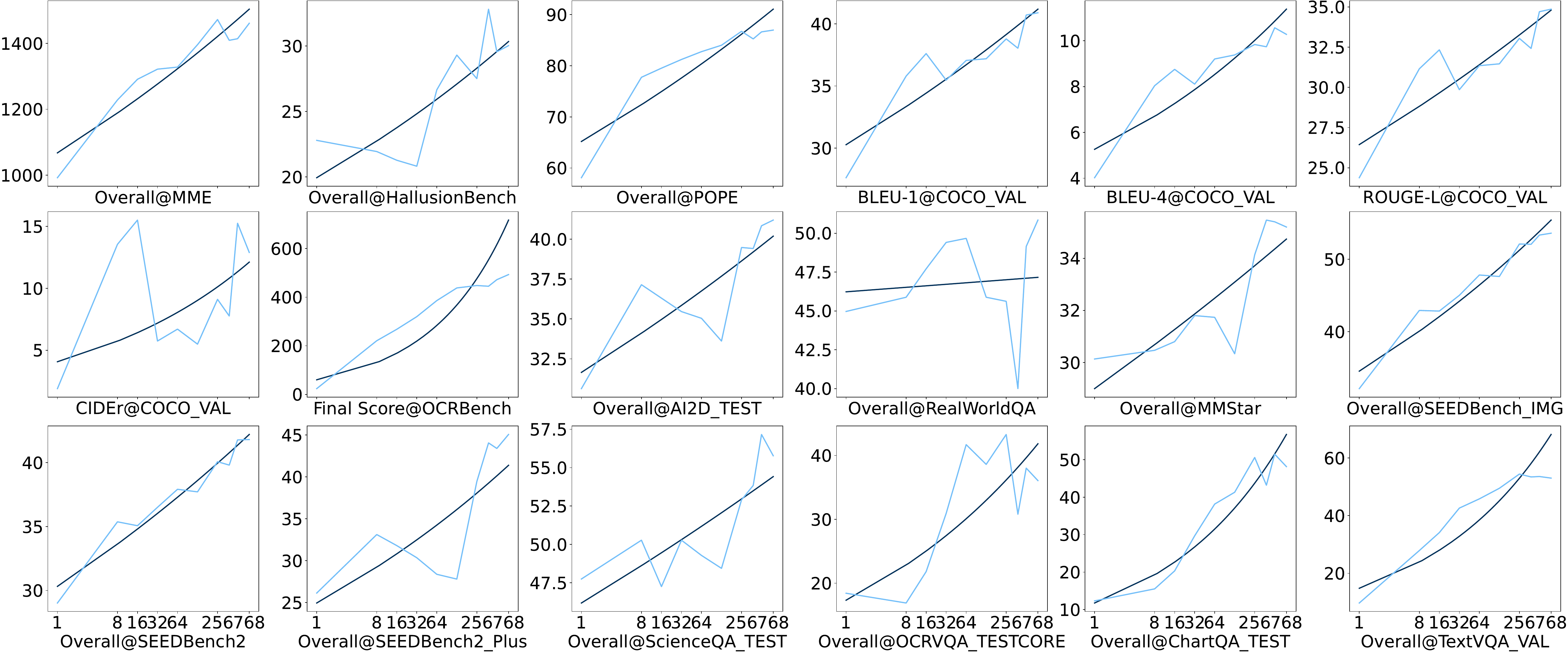}
  \caption{Scaling behavior of performances with respect to \(n_l\). The models are taking vision tokens, user's questions and learnable queries as input (``Vision Question Queries''). The x-axis in each subplot is log scalded, \(\log_2(n_l)\). The deep blue lines stand for the fitted curves, and the light blue lines represent the original data points.}
  \label{fig:scaling-fit-vqq}
\end{figure}

\begin{figure}[htb]
  \center
  \includegraphics[width=0.999\linewidth]{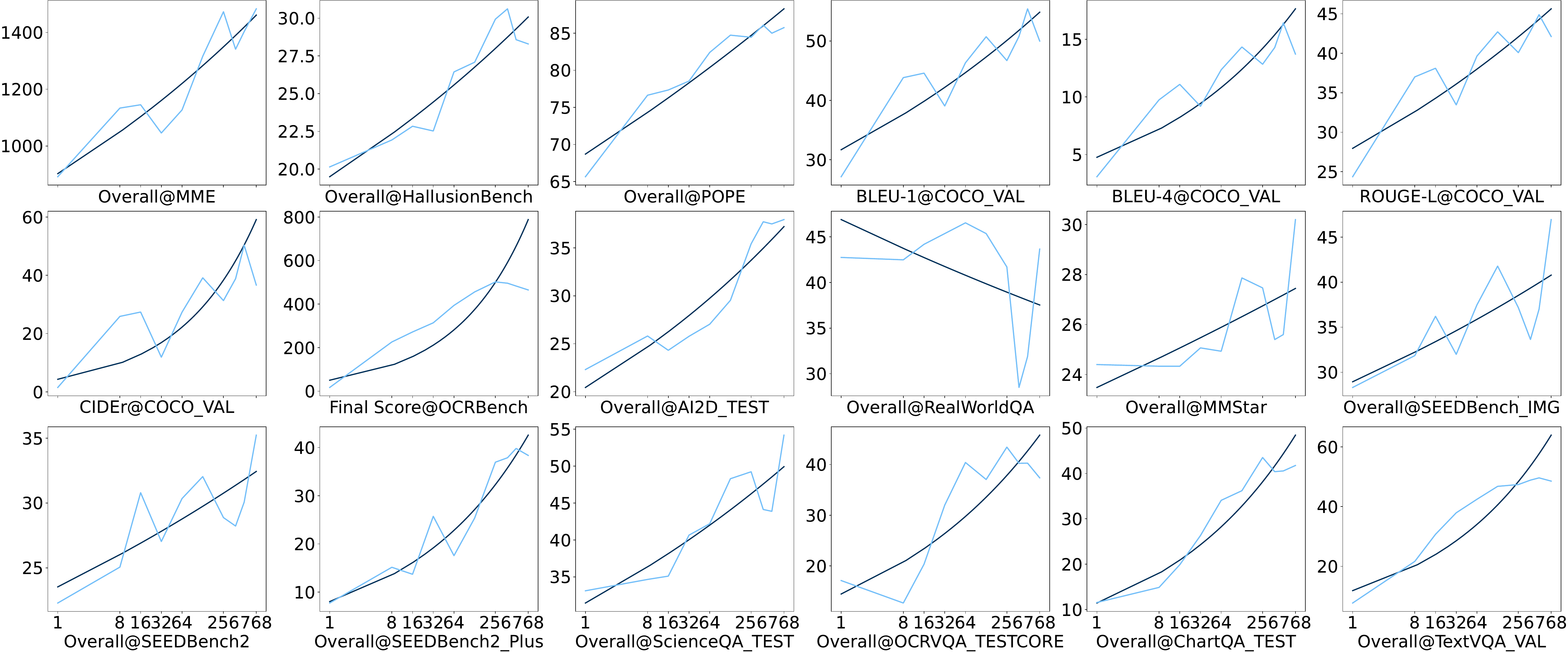}
  \caption{Scaling behavior of performances with respect to \(n_l\). The further fine-tuned models are taking vision tokens and learnable queries as input (``Vision Queries (ft)''). The x-axis in each subplot is log scalded, \(\log_2(n_l)\). The deep blue lines stand for the fitted curves, and the light blue lines represent the original data points.}
  \label{fig:scaling-fit-vq-ft}
\end{figure}

To improve numerical stability, logarithms are applied, and the curves are fitted via the following mean square optimization problem:
\begin{equation}
  \label{equ:optim-scaling-law-compute-vision-token}
  \argmin_{\alpha, z} \dfrac{1}{\left|\Omega_{n_l}\right|} \sum_{n \in \Omega_{n_l}}
  \left( \log(S(n_l)) - \left( \log(c) - \alpha \log(n_l) \right) \right)^2,
\end{equation}
where \(\left|\Omega_{n_l}\right|\) denotes the number of sampled points.
Using the default BFGS optimizer from SciPy, performance data is fitted to generate the curves illustrated in Figures \ref{fig:scaling-fit-vqq} and \ref{fig:scaling-fit-vq-ft}, with corresponding fitted parameters \(c\) and \(\alpha\) summarized in Table \ref{tab:scaling-analysis}.

\begin{table}[htb]
  \small
  \centering
  \caption{
    The fitted scaling parameters, \(c, \alpha\), for each benchmark as illustrated in Figure \ref{fig:scaling-fit-vqq} and Figure \ref{fig:scaling-fit-vq-ft}.
    The columns starting with ``Vision Question Queries'' are for models taking vision tokens, user's questions and learnable queries as input.
    The columns starting with ``Vision Queries (ft)'' are for further fine-tuned models taking vision tokens and learnable queries as input.
  }
  \label{tab:scaling-analysis}
  \begin{tabular}{r|cc|cc}
    \toprule
    Benchmark               & $c$        & $\alpha$ & $c$        & $\alpha$ \\
    \midrule
    Overall@MME             & 1.0676e+03 & -0.0516  & 9.0251e+02 & -0.0725  \\
    Overall@HallusionBench  & 1.9935e+01 & -0.0632  & 1.9491e+01 & -0.0653  \\
    Overall@POPE            & 6.5197e+01 & -0.0503  & 6.8697e+01 & -0.0377  \\
    BLEU-1@COCO VAL         & 3.0271e+01 & -0.0463  & 3.1687e+01 & -0.0826  \\
    BLEU-4@COCO VAL         & 5.2622e+00 & -0.1161  & 4.7702e+00 & -0.1968  \\
    ROUGE-L@COCO VAL        & 2.6441e+01 & -0.0413  & 2.7957e+01 & -0.0737  \\
    CIDEr@COCO VAL          & 4.0698e+00 & -0.1642  & 4.3223e+00 & -0.3937  \\
    Final Score@OCRBench    & 6.0272e+01 & -0.3727  & 5.0303e+01 & -0.4142  \\
    Overall@AI2D TEST       & 3.1651e+01 & -0.0359  & 2.0443e+01 & -0.0902  \\
    Overall@RealWorldQA     & 4.6232e+01 & -0.0030  & 4.6893e+01 & ~0.0335   \\
    Overall@MMStar          & 2.8996e+01 & -0.0272  & 2.3483e+01 & -0.0235  \\
    Overall@SEEDBench IMG   & 3.4562e+01 & -0.0710  & 2.8942e+01 & -0.0516  \\
    Overall@SEEDBench2      & 3.0322e+01 & -0.0498  & 2.3532e+01 & -0.0483  \\
    Overall@SEEDBench2 Plus & 2.4944e+01 & -0.0762  & 8.0294e+00 & -0.2512  \\
    Overall@ScienceQA TEST  & 4.6179e+01 & -0.0247  & 3.1462e+01 & -0.0695  \\
    Overall@OCRVQA TESTCORE & 1.7333e+01 & -0.1326  & 1.4473e+01 & -0.1734  \\
    Overall@ChartQA TEST    & 1.1750e+01 & -0.2370  & 1.1438e+01 & -0.2173  \\
    Overall@TextVQA VAL     & 1.4812e+01 & -0.2297  & 1.1842e+01 & -0.2538  \\
    \bottomrule
  \end{tabular}
\end{table}

The fitted curves reveal a clear logarithmic relationship between the number of vision tokens \(n_l\) and performance across most benchmarks, regardless of whether the models incorporate the user's questions as input.
This trend aligns well with the theoretical scaling behavior described in Equation \ref{equ:scaling-law-token-space}.
The parameter \(\alpha\) represents the slope of the logarithmic relationship, indicating the rate of performance improvement as the number of tokens increases.
Across the benchmarks, \(\alpha\) values are consistently negative, reflecting diminishing returns in performance gains as the token count grows larger.

In benchmarks such as OCRBench and ChartQA, steeper negative \(\alpha\) values highlight that performance is more sensitive to the number of vision tokens.
This indicates a stronger dependence on token count for these tasks, where reducing tokens could lead to substantial performance drops.
Conversely, benchmarks like ScienceQA TEST and MMStar exhibit relatively flat \(\alpha\) values, suggesting that performance is less sensitive to token scaling.

The constant \(c\), derived from the fitted curves, provides an additional perspective on baseline performance.
Higher values of \(c\) correspond to benchmarks with inherently stronger initial performance, even with fewer tokens.
For example, benchmarks such as HallusionBench and ScienceQA TEST exhibit larger \(c\) values, indicating robustness in initial model performance.

Through Figures \ref{fig:scaling-fit-vqq} and \ref{fig:scaling-fit-vq-ft}, a significant performance degradation can be observed in the RealWorldQA benchmark at sequence lengths \(n_l = 384\) and \(n_l = 512\), which results in a failure to fit the theoretical scaling law.
Analysis of the evaluation logs reveals that this performance drop stems from incorrect response formatting rather than fundamental model limitations.

For instance, when presented with the question ``Is there any flower in the picture?'', the expected response format requires selecting option ``A: Yes'' or ``B: No''. However, the model generates \verb*|<st>flowers<ed> <bbox>999,569,1009,609</bbox> </st>|, which constitutes an invalid answer format.
Similarly, for the question ``Which object is closer?'', where the correct response should be chosen from ``A: The stop sign.'' or ``B: The school zone sign.'', the model produces \verb*|<st>sign<ed> <bbox>899,359,999,419</bbox> Answer: B.|, which cannot be correctly parsed by the evaluation script despite containing the intended answer.

This formatting issue occurs most frequently in the ``Vision Queries (ft)'' configuration.
The underlying cause likely stems from the fact that the original model was trained with user questions, while this configuration underwent fine-tuning using only 10\% of the sampled data without the questions in fusion module.
Consequently, the model demonstrates a poor adaptation to evaluation formats that lack conversational context, resulting in responses that fail to conform to the required answer structure.

\begin{figure}[htb]
  \centering
  \includegraphics[width=0.7\linewidth]{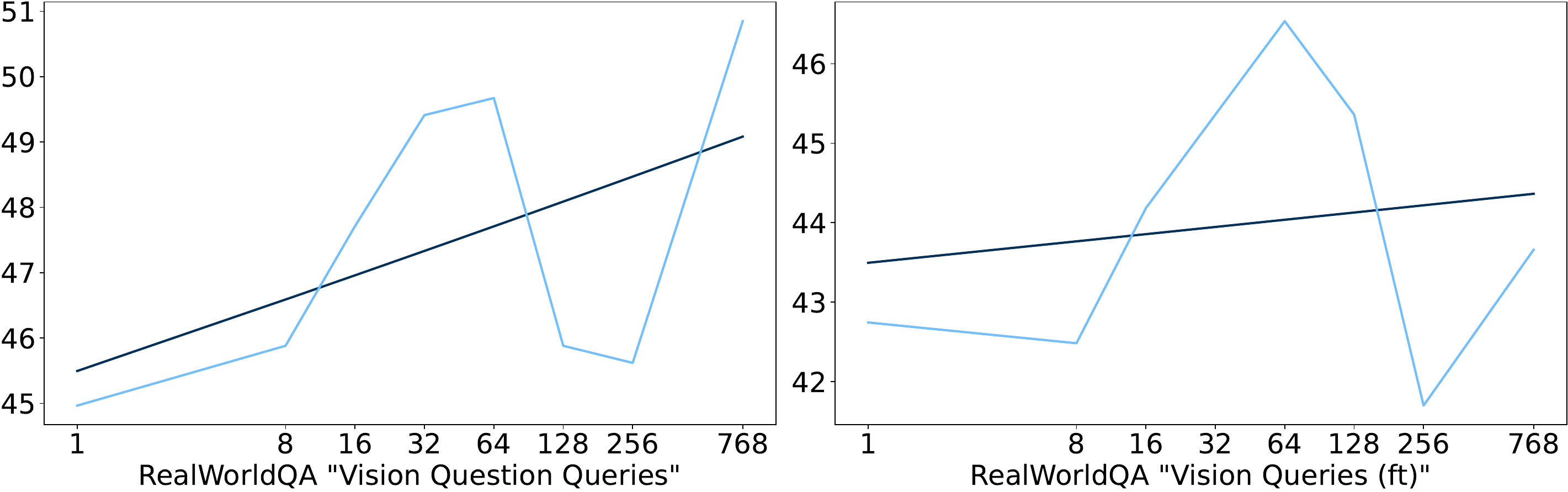}
  \caption{
    Fitted scaling curves for RealWorldQA benchmark with excluding anomalous data points at \(n_l = 384\) and \(n_l = 512\) excluded.
    For RealWorldQA ``Vision Question Queries'': \(c = 45.496627\), \(\alpha = -0.011420\).
    For RealWorldQA ``Vision Queries (ft)'': \(c = 43.495766\), \(\alpha = -0.002974\).
    The removal of these outliers demonstrates improved adherence to the theoretical scaling law.
  }
  \label{fig:scaling-fit-rwqa-drop-384-512}
\end{figure}

When these anomalous data points are excluded from the analysis, the fitted curves exhibit substantially improved consistency with the theoretical scaling behavior.
An illustration of the corrected scaling relationship, with data points at \(n_l = 384\) and \(n_l = 512\) removed, is presented in Figure \ref{fig:scaling-fit-rwqa-drop-384-512}.

While these fitted curves are not perfect, they sufficiently capture the overall scaling behavior across different benchmarks.
The variations in \(\alpha\) and \(c\) highlight task-specific differences in sensitivity to token scaling, offering insights for optimizing model architectures.

The fitted curves underscore the scaling capabilities of vision language models and provide a foundation for developing strategies that balance computational efficiency with task-specific performance.
The analysis reveals that: the model demonstrates scaling capability following \(S(n_l) \approx c / n_{l}^{\alpha}\), consistent with the theoretical scaling behavior.
This behavior is robust across different input configurations, and does not significantly be effected by the inclusion or exclusion of user questions in the input.

\subsection{Analysis on Fusing User Questions}
\label{subsec:exp-fusion-with-user-questions}

In subsection \ref{subsec:dtsa-input_pattern_formalization}, input patterns are formalized into a uniform representation:
\begin{equation*}
  \underbrace{\left[\tokenm{txt} \ldots \tokenm{txt}\right]}_{\text{non-referential}}
  \overbrace{
    \left[\tokenm{txt} \ldots \tokenm{txt}\right]
    \left[\tokenm{vis} \ldots \right]
  }^{\text{vision-referencing}}.
\end{equation*}
With this representation, text prompt tokens that relate to visual content are classified as vision-referencing components. In practical applications, the input text prompt corresponds to the user question. This formalization enables us to investigate how the presence of vision-referencing information in user questions influences model performance.

The impact of vision-referencing content in user questions can be analyzed from two complementary perspectives.

\textbf{Enhancing attention mechanisms.} On one hand, vision-referencing information enhances attention mechanisms. Vision-referencing content may improve the Vision-Question Fusion model's ability to understand user intent and focus on relevant image regions.
For instance, the question ``What is located in the left corner of the image?'' contains explicit spatial reference (``left corner'').
This spatial reference enabling the fusion model to direct attention to the corresponding image region while disregarding irrelevant areas, potentially yielding more accurate responses.

\textbf{Pseudo extending vision sequence.} On the other hand, vision-referencing tokens increase sequence length. According to the uniform representation framework, if the user's question contains information related to vision-referencing, it can be considered as a vision-referencing component.
This can also be considered as a pseudo expansion of the vision sequence, producing a longer vision token sequence.
According to the scaling relationship, the extending in sequence length should correspond to improvement in model performance.

Synthesizing these two mechanisms, it is possible to hypothesize that model performance will improve when user questions contain vision-referencing information, while remaining unaffected when questions lack visual grounding.
To empirically validate this hypothesis, an analysis is conducted by leveraging the experimental results from the Subsection \ref{subesc:exp-scaling_analysis}.

\begin{figure}[htb]
  \centering
  \hspace*{-1em}
  \begin{tikzpicture}[yscale=-1]
    \node[anchor=north west] at (0, 0.2) {
      \includegraphics[width=0.495\textwidth]{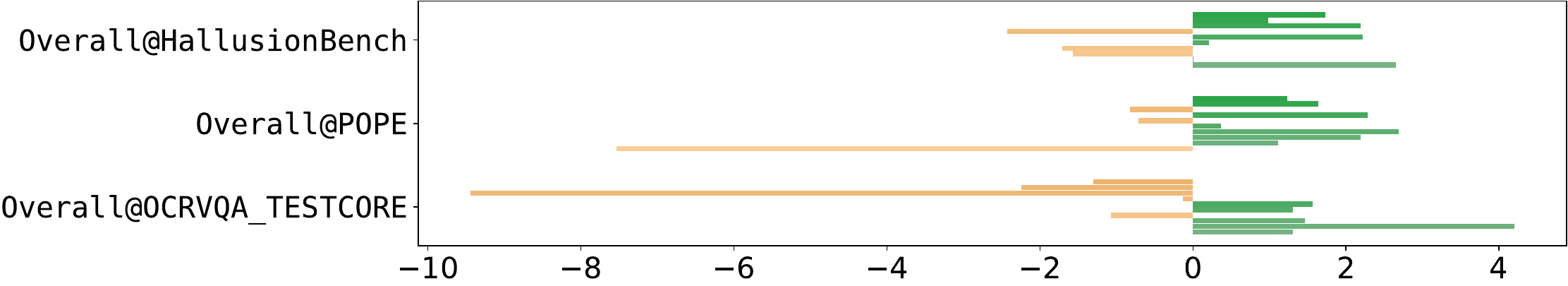}
    };
    \node[anchor=north west] at (0, 2.0) {
      \includegraphics[width=0.495\textwidth]{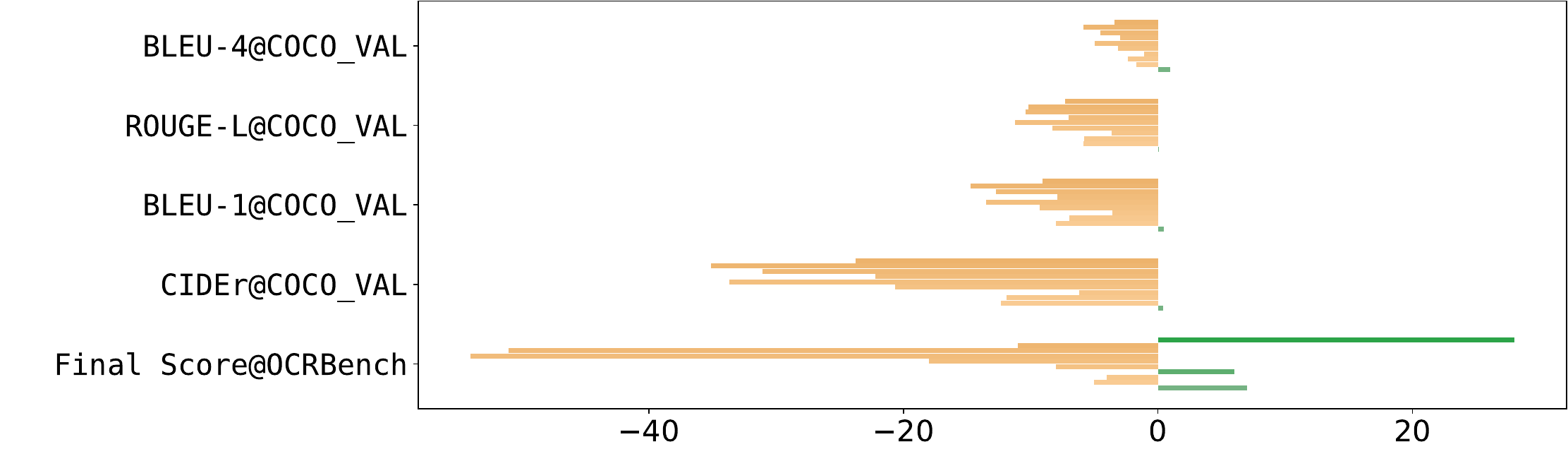}
    };
    \node[anchor=north west] at (0.495\textwidth,0.0) {
      \includegraphics[width=0.495\textwidth]{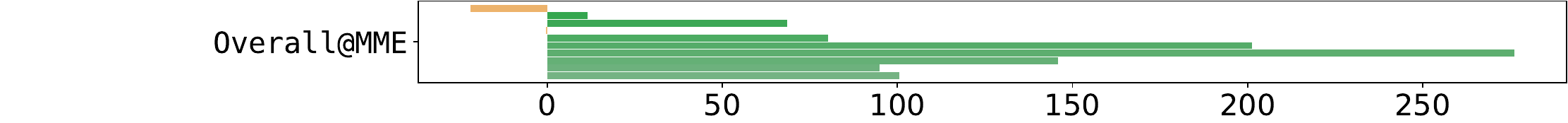}
    };
    \node[anchor=north west] at (0.495\textwidth,1.0) {
      \includegraphics[width=0.495\textwidth]{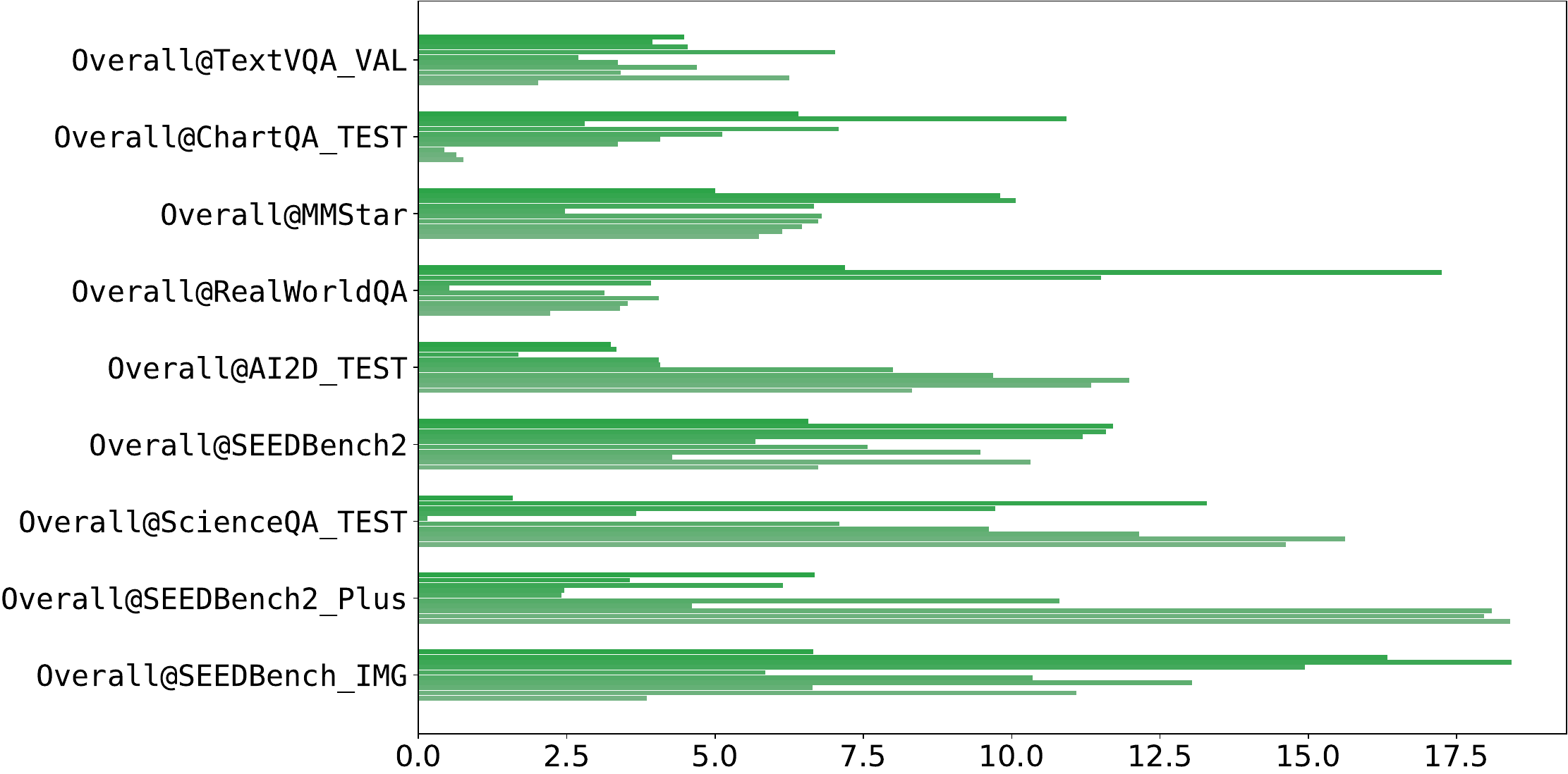}
    };
  \end{tikzpicture}
  \caption{
    A horizontal bar plot on the performance differences \(S_{\text{vqq}} - S_{\text{vq-ft}}\) for different benchmarks, where \(S_{\text{vqq}}\) is the model with user's questions and \(S_{\text{vq-ft}}\) is the further fine-tuned model without user's questions.
    Each benchmark is represented as a separate plot, and within each plot, the horizontal bars correspond to \(n_l = 1, 8, \dots, 768\), ordered from top to bottom.
  }
  \label{fig:tab_diff_vqq-vq_ft}
\end{figure}

Let \(S_{\text{vqq}}\) denote the performance value of the models that incorporate the user's question as part of the input.
For comparison, the model was further fine-tuned without user's question as part of the input, as outlined in Appendix \ref{appendix:implementation_details}.
Let \(S_{\text{vq-ft}}\) represent the performance value of the fine-tuned models that exclude the user's question.
The performance difference between the two models, \(S_{\text{vqq}} - S_{\text{vq-ft}}\), is visualized as horizontal bar plots in Figure \ref{fig:tab_diff_vqq-vq_ft}.
For each horizontal bar plot, the color represents the performance difference \(S_{\text{vqq}} - S_{\text{vq-ft}}\).
If the value is positive (\(S_{\text{vqq}} - S_{\text{vq-ft}} > 0\)), the bar is colored green, indicating that the model incorporating the user's question outperforms the fine-tuned model without the question.
Conversely, if the value is negative (\(S_{\text{vqq}} - S_{\text{vq-ft}} < 0\)), the bar is colored orange, indicating that the fine-tuned model without the question achieves better performance than the model with the question.

From Figure \ref{fig:tab_diff_vqq-vq_ft}, it can be observed that green areas (positive performance differences) predominate across most benchmarks, supporting our hypothesis that incorporating user questions with vision-referencing content improves model performance. However, notable exceptions exist where orange areas dominate, indicating that the fine-tuned model without user questions achieves superior performance.

A prominent counterexample is observed in the \textit{COCO VAL} benchmark, which exhibits more orange than green areas. This benchmark employs the question: ``Please describe this image in general.
Directly provide the description, do not include a prefix like `This image depicts'.''.
This question requires comprehensive image descriptions but lacks explicit vision-referencing content such as spatial references or object-specific queries.
According to the theoretical framework, such non-referential questions should not provide the attention-focusing benefits described earlier.
Instead, they may introduce unnecessary computational overhead without corresponding performance gains, explaining why the fine-tuned model without the question achieves better results in this specific case.
This finding aligns with the hypothesis that performance improvements are contingent upon the presence of meaningful vision-referencing information in user questions.

To provide a more direct assessment, the models are evaluated without user questions in their original state, and without additional fine-tuning.
This approach isolates the inherent impact of including user questions on model performance.

\begin{figure}[htb]
  \centering
  \hspace*{-2em}
  \begin{tikzpicture}[yscale=-1]
    \node[anchor=north west] at (0.0, 1.0) {
      \includegraphics[width=0.495\textwidth]{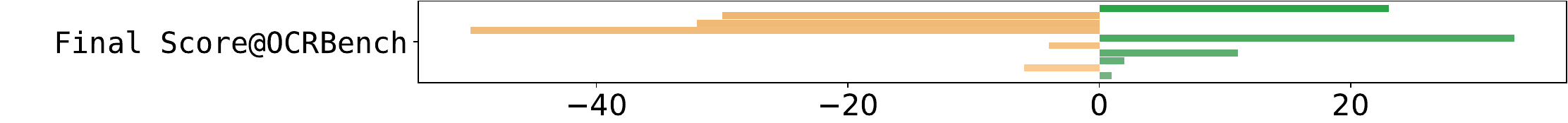}
    };
    \node[anchor=north west] at (0.0, 2.0) {
      \includegraphics[width=0.495\textwidth]{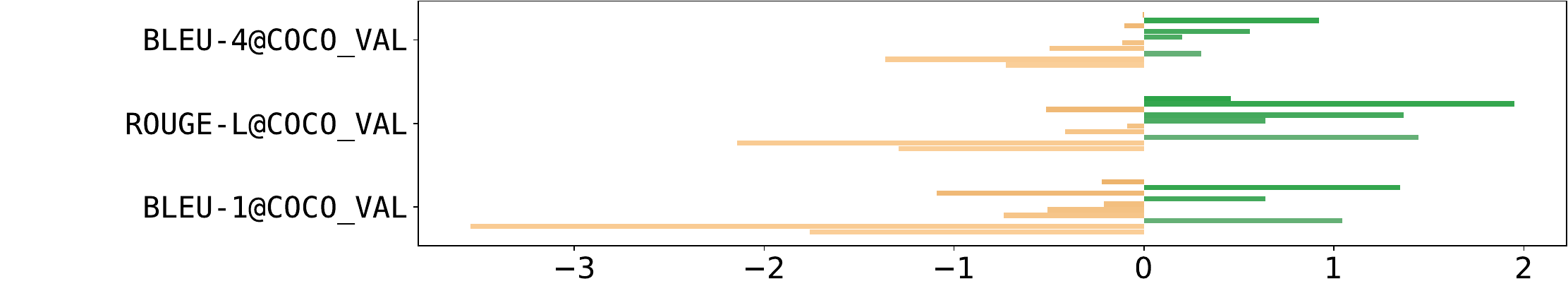}
    };
    \node[anchor=north west] at (0.0, 3.9) {
      \includegraphics[width=0.495\textwidth]{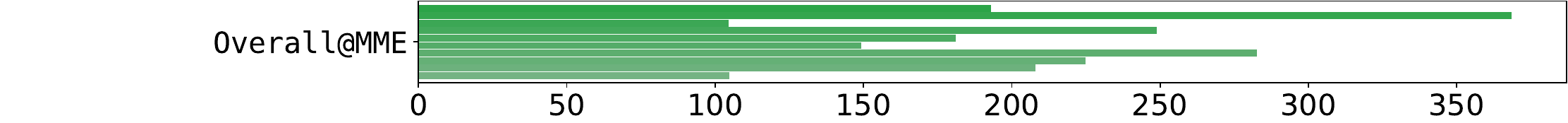}
    };
    \node[anchor=north west] at (0.495\textwidth, 0.0) {
      \includegraphics[width=0.495\textwidth]{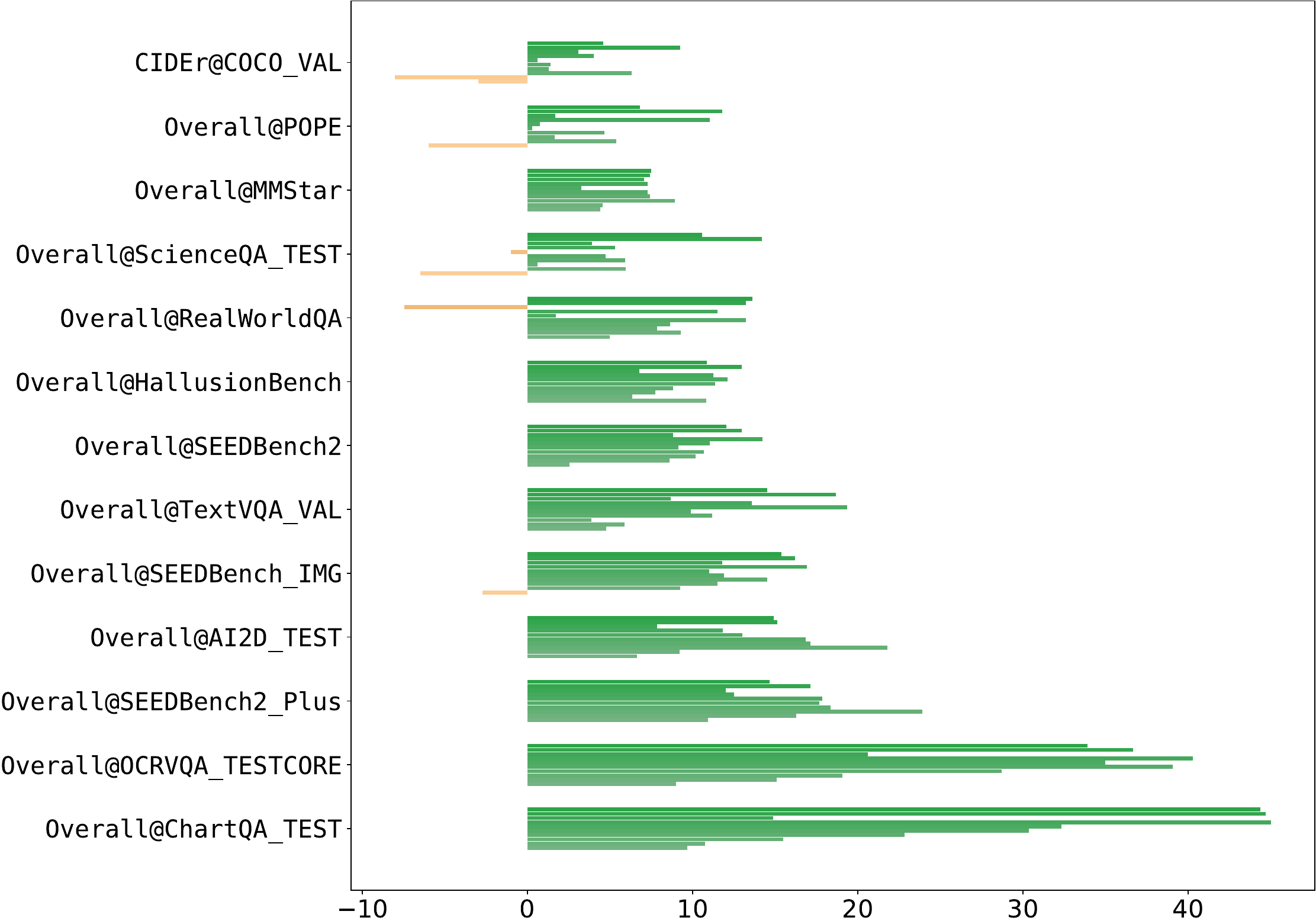}
    };
  \end{tikzpicture}
  \caption{
    A horizontal bar plot on the performance difference \(S_{\text{vqq}} - S_{\text{vq}}\) for each benchmark, where \(S_{\text{vqq}}\) is the model with user's questions and \(S_{\text{vq}}\) is the model without user's questions.
    Each benchmark is represented as a separate plot, and within each plot, the horizontal bars correspond to \(N_l = 1, 8, \dots, 768\), ordered from top to bottom.
  }
  \label{fig:tab_diff_vqq-vq}
\end{figure}

Let \(S_{\text{vq}}\) denote the performance of the model without incorporating the user's question as input and without any fine-tuning.
The performance difference, \(S_{\text{vqq}} - S_{\text{vq}}\), is computed for each benchmark and visualized in Figure \ref{fig:tab_diff_vqq-vq} with the same color scheme as Figure \ref{fig:tab_diff_vqq-vq_ft}.
It can be observed that more green bars are present compared to orange bars, suggesting that the model with the user's question generally outperforms the model without the user's question.
Additionally, the relative performance range shown on the right-hand side of Figure \ref{fig:tab_diff_vqq-vq} spans \([0, 40]\), which is significantly larger than the range observed in Figure \ref{fig:tab_diff_vqq-vq_ft} (\([0, 17]\)).
This indicates that in these benchmarks, the model relies heavily on the user's questions to achieve better performance.

Similar to Figure \ref{fig:tab_diff_vqq-vq_ft}, there are some outliers in Figure \ref{fig:tab_diff_vqq-vq}, proposing that under some specific conditions, the model without the user's question can slightly perform better than the model with the user's question.

This observation suggests that the presence of user questions is possible to improve the model's performance if the user's question is meaningful and relevant to the task. This finding can align with the hypothesis that relevant user questions can provide additional context and guidance to the model, or leading to pseudo extending vision sequences.

\section{Conclusion and Future Work}
\label{sec:conclusion}

\textbf{Conclusion.}
This research establishes a fundamental understanding of how vision token quantity influences model performance in vision-language tasks through both theoretical analysis and empirical validation.
The central question of characterizing the relationship between vision tokens and model discriminative capacity has been addressed through a comprehensive mathematical framework that reveals predictable scaling behaviors.

The theoretical analysis demonstrates that the expected divergence between vision-referencing sequences scales according to \(\mathbb{E}_{\mathcal{V}}\left[\mathcal{D}(n)\right] = \bigO{\sqrt{\Upsilon(n)}}\), exhibiting two distinct scaling regimes: sublinear \(\bigO{\sqrt{n}}\) scaling for small token counts and linear \(\bigO{n}\) scaling for large token counts.
This theoretical characterization is aligned with the model performance scaling relationships of the form \(S(n) \approx c / n^{\alpha(n)}\), where the scaling exponent \(\alpha(n)\) depends on the balance between these regimes.
Empirical validation across multiple vision-language benchmarks confirms that model performance closely follows these predicted scaling relationships, demonstrating the practical relevance of the theoretical framework.

These findings contribute to understanding vision token scaling through a mathematical framework that reveals predictable performance relationships.
The theoretical analysis may provide insights for vision token optimization and architectural design decisions in vision-language systems, complementing existing empirical approaches to multimodal model development.

The research acknowledges certain limitations in its current scope.
The theoretical analysis may not hold universally across all model architectures or datasets.
Additionally, the empirical validation focuses on a particular model architecture and benchmark suite, which may limit the generalizability of the specific scaling parameters observed.

\textbf{Future work.}
Several promising directions emerge from this foundation for future investigation. 
The theoretical framework can be extended to accommodate more general situations and correlation structures, potentially revealing scaling behaviors in broader classes of vision-language architectures.
Investigation of how different vision encoder designs, attention mechanisms, and training procedures influence the scaling parameters \(\psi_{\tequal}^{(AB)}(n)\), \(\psi_{\tcross}^{(AA)}(n)\), and \(\psi_{\tcross}^{(AB)}(n)\) would provide deeper insights into the sources of scaling behavior variations.
Broader empirical validation across different model families and architectural approaches would help establish the generalizability of the scaling relationships beyond the configurations examined in this study.
Extending the analysis to other modalities, such as video or audio tokens, could reveal whether similar scaling principles apply to multimodal integration more generally.

The finding also suggests practical applications worth exploring, including adaptive vision token allocation strategies and analysis of vision token compression techniques.
Wish that the mathematical framework can help to understand these approaches and their impact on model capabilities.

\acks{
  The authors would like to express their sincere gratitude to the anonymous reviewers and action editor for their insightful comments and valuable suggestions, which significantly improved the quality of this paper.
  We are particularly grateful to Dr. Andong Wang for his helpful discussions on the writing.

  This study was supported in part by Moonshot R\&D Grant Number JPMJMS2239.
  Tenghui Li was supported by RIKEN’s IPA Program.
}

\newpage

\appendix

\section{Additional Discussion for Divergence Token Analysis}
\label{appendix:additional_discussion_theory}

\subsection{Additional Derivative for Divergence Scaling}
\label{appendix:derivative_divergence_scaling}

The complete derivative for Equation \eqref{equ:exp_sum_diag_corss_split}, Inequality \eqref{equ:E_diff_hijAB_expand_cos_M} and \eqref{equ:E_diff_hijAB_bound}:
\begin{equation*}
  \begin{aligned}
    &\mathbb{E}_{\mathcal{V}} \left[ \left\| \sum_{i=1}^{n} \left( \tvar{h}_i^{(A)} - \tvar{h}_i^{(B)} \right) \right\|^2_F \right] \\
    &= \sum_{i=1}^{n} \mathbb{E}_{\mathcal{V}} \left[ \left\| \tvar{h}_i^{(A)} - \tvar{h}_i^{(B)} \right\|_F^2 \right] 
    + \sum_{i,j=1, i\ne j}^{n} \mathbb{E}_{\mathcal{V}} \left[ \left( \tvar{h}_i^{(A)} - \tvar{h}_i^{(B)} \right)^{\top} \left( \tvar{h}_j^{(A)} - \tvar{h}_j^{(B)} \right) \right] \\
    &= \sum_{i=1}^{n} \left(
      \mathbb{E}_{\mathcal{V}} \left[ \left\| \tvar{h}_i^{(A)} \right\|_F^2 \right]
      + \mathbb{E}_{\mathcal{V}} \left[ \left\| \tvar{h}_i^{(B)} \right\|_F^2 \right]
      - 2 \mathbb{E}_{\mathcal{V}} \left[ \left(\tvar{h}_i^{(A)}\right)^{\top} \tvar{h}_i^{(B)} \right]
    \right) \\
    &~ + \sum_{i,j=1, i \ne j}^{n} \left(
      \mathbb{E}_{\mathcal{V}} \left[ \left(\tvar{h}_i^{(A)}\right)^{\top} \tvar{h}_j^{(A)} \right]
      + \mathbb{E}_{\mathcal{V}} \left[ \left(\tvar{h}_i^{(B)}\right)^{\top} \tvar{h}_j^{(B)} \right]
    \right) \\
    &~ - \sum_{i,j=1, i \ne j}^{n} \left(
      \mathbb{E}_{\mathcal{V}} \left[ \left(\tvar{h}_i^{(A)}\right)^{\top} \tvar{h}_j^{(B)} \right]
      + \mathbb{E}_{\mathcal{V}} \left[ \left(\tvar{h}_i^{(B)}\right)^{\top} \tvar{h}_j^{(A)} \right]
    \right) \\
    &= \sum_{i=1}^{n} \left(
      \mathbb{E}_{\mathcal{V}} \left[ \left\| \tvar{h}_i^{(A)} \right\|_F^2 \right]
      + \mathbb{E}_{\mathcal{V}} \left[ \left\| \tvar{h}_i^{(B)} \right\|_F^2 \right]
      - 2 \mathbb{E}_{\mathcal{V}} \left[
        \left\| \tvar{h}_i^{(A)} \right\|_F
        \left\| \tvar{h}_i^{(B)} \right\|_F
        \cos\left( \Theta_{ii}^{(AB)} \right)
      \right]
    \right) \\
    &~ + \sum_{i,j=1, i \ne j}^{n} \left(
      \mathbb{E}_{\mathcal{V}} \left[
        \left\| \tvar{h}_i^{(A)} \right\|_F
        \left\| \tvar{h}_j^{(A)} \right\|_F
        \cos\left( \Theta_{ij}^{(AA)} \right)
      \right]
      + \mathbb{E}_{\mathcal{V}} \left[
        \left\| \tvar{h}_i^{(B)} \right\|_F
        \left\| \tvar{h}_j^{(B)} \right\|_F
        \cos\left( \Theta_{ij}^{(BB)} \right)
      \right]
    \right) \\
    &~ - \sum_{i,j=1, i \ne j}^{n} \left(
      \mathbb{E}_{\mathcal{V}} \left[
        \left\| \tvar{h}_i^{(A)} \right\|_F
        \left\| \tvar{h}_j^{(B)} \right\|_F
        \cos\left( \Theta_{ij}^{(AB)} \right)
      \right]
      + \mathbb{E}_{\mathcal{V}} \left[
        \left\| \tvar{h}_i^{(B)} \right\|_F
        \left\| \tvar{h}_j^{(A)} \right\|_F
        \cos\left( \Theta_{ij}^{(BA)} \right)
      \right]
    \right). \\
    &\leq 2 M^{2} \sum_{i=1}^{n} \left(
      1
      - 2 \mathbb{E}_{\mathcal{V}} \left[
        \cos\left( \Theta_{ii}^{(AB)} \right)
      \right]
    \right) \\
    &~ + M^{2} \sum_{i,j=1, i \ne j}^{n} \left(
      \mathbb{E}_{\mathcal{V}} \left[
        \cos\left( \Theta_{ij}^{(AA)} \right)
      \right]
      + \mathbb{E}_{\mathcal{V}} \left[
        \cos\left( \Theta_{ij}^{(BB)} \right)
      \right]
      - 2 \mathbb{E}_{\mathcal{V}} \left[
        \cos\left( \Theta_{ij}^{(AB)} \right)
      \right]
    \right) \\
    &\leq 2 M^2 \sum_{i=1}^{n} \left( 1 - \psi_{\tequal}^{(AB)}(n) \right) 
      + M^2 \sum_{i,j=1, i \ne j}^{n} \left(
      \psi_{\tcross}^{(AA)}(n)
      + \psi_{\tcross}^{(BB)}(n)
      - 2 \psi_{\tcross}^{(AB)}(n)
    \right) \\
    &\leq 2 M^2 \left(
      n \left(1 - \psi_{\tequal}^{(AB)}(n)\right)
      + (n^2-n) \left( \psi_{\tcross}^{(AA)}(n) - \psi_{\tcross}^{(AB)}(n) \right)
    \right) \\
    &= \bigO{
      n \left(1 - \psi_{\tequal}^{(AB)}(n)\right)
      + (n^2-n) \left( \psi_{\tcross}^{(AA)}(n) - \psi_{\tcross}^{(AB)}(n) \right)
    } \\
    &:= \bigO{\Upsilon(n)}.
  \end{aligned}
\end{equation*}

\subsection{Analysis of Dependency Measures}
\label{appendix:dependency_constraints}

This appendix provides a detailed analysis of the constraints imposed by the dependency measures \(\psi_{\tequal}^{(AB)}\), \(\psi_{\tcross}^{(AA)}\), and \(\psi_{\tcross}^{(AB)}\) on the sequence length \(n\) and examines the conditions under which meaningful divergence bounds can be established.

Recall the upper bound proposed in Inequality \ref{equ:expected_divergence_bound_sqrt}, which is given by:
\begin{equation*}
  \sqrt{
    n \left(1 - \psi_{\tequal}^{(AB)}\right) + (n^2-n) \left( \psi_{\tcross}^{(AA)} - \psi_{\tcross}^{(AB)} \right)
  }.
\end{equation*}
The fundamental constraint arises from requiring the expression within the square root remain non-negative (in real numbers). Rearranging the fundamental constraint yields:
\begin{equation}
  \label{equ:rearranged_constraint}
  n \left(
    \psi_{\tcross}^{(AA)} - \psi_{\tcross}^{(AB)}
  \right) \geq 
  \psi_{\tequal}^{(AB)} + \psi_{\tcross}^{(AA)} - \psi_{\tcross}^{(AB)} - 1,
\end{equation}
where \(n \geq 1\) by definition and all dependency measures are bounded within \([0,1]\).
There are three distinct cases based on the relationship between \(\psi_{\tcross}^{(AA)}\) and \(\psi_{\tcross}^{(AB)}\).

1. \(\psi_{\tcross}^{(AA)} = \psi_{\tcross}^{(AB)}\):
When \(\psi_{\tcross}^{(AA)}\) and \(\psi_{\tcross}^{(AB)}\) are equal, the quadratic term in the divergence bound vanishes. In this scenario, constraint \eqref{equ:rearranged_constraint} is always satisfied for any \(n \geq 1\), and the divergence bound is simplified to \(\bigO{\sqrt{n}}\).

2. \(\psi_{\tcross}^{(AA)} > \psi_{\tcross}^{(AB)}\):
This case represents the most comment scenario where temporal similarity within the same branch is greater than that between different branches.
The constraint becomes:
\begin{equation}
  n \geq \frac{
    \psi_{\tequal}^{(AB)} + \psi_{\tcross}^{(AA)} - \psi_{\tcross}^{(AB)} - 1
  }{
    \psi_{\tcross}^{(AA)} - \psi_{\tcross}^{(AB)}
  }.
\end{equation}

Since \(\psi_{\tequal}^{(AB)} \in [0, 1]\) and \(\psi_{\tcross}^{(AA)} - \psi_{\tcross}^{(AB)} \geq 0\), it can be shown that:
\begin{equation}
  \frac{
    \psi_{\tequal}^{(AB)} + \psi_{\tcross}^{(AA)} - \psi_{\tcross}^{(AB)} - 1
  }{
    \psi_{\tcross}^{(AA)} - \psi_{\tcross}^{(AB)}
  }
  = 1 + \frac{
    \psi_{\tequal}^{(AB)} - 1
  }{
    \psi_{\tcross}^{(AA)} - \psi_{\tcross}^{(AB)}
  }
  \leq 1.
\end{equation}
Therefore, the constraint is automatically satisfied for all \(n \geq 1\), ensuring a meaningful bound can be established for any sequence length.

3. \(\psi_{\tcross}^{(AA)} < \psi_{\tcross}^{(AB)}\):
This case represents an unusual scenario where the temporal similarity between different branches is greater than that within the same branch. While this is theoretically possible, it is highly unlikely in practical applications because the similarity of different tokens in different branches should not be higher than that in the same branch.
Although this case is practically unlikely, it is still valid to analyze here, and the constraint becomes,
\begin{equation}
  n \leq \frac{
    \psi_{\tequal}^{(AB)} + \psi_{\tcross}^{(AA)} - \psi_{\tcross}^{(AB)} - 1
  }{
    \psi_{\tcross}^{(AA)} - \psi_{\tcross}^{(AB)}
  } = 1 + \frac{
    \psi_{\tequal}^{(AB)} - 1
  }{
    \psi_{\tcross}^{(AA)} - \psi_{\tcross}^{(AB)}
  }.
\end{equation}
Since \(\psi_{\tequal}^{(AB)} - 1 \leq 0\) and \(\psi_{\tcross}^{(AA)} - \psi_{\tcross}^{(AB)} < 0\), the fraction \(\frac{\psi_{\tequal}^{(AB)} - 1}{\psi_{\tcross}^{(AA)} - \psi_{\tcross}^{(AB)}} \geq 0\).
It shows that the sequence length \(n\) should be less than a threshold.
As \(\psi_{\tcross}^{(AA)} - \psi_{\tcross}^{(AB)}\) approaches zero, this fraction can become arbitrarily large, leading to:
\begin{equation}
  0 \leq \frac{
    \psi_{\tequal}^{(AB)} - 1
  }{
    \psi_{\tcross}^{(AA)} - \psi_{\tcross}^{(AB)}
  } \leq +\infty.
\end{equation}

This imposes an upper bound on the sequence length \(n\) for which meaningful divergence bounds can be established. If the actual sequence length exceeds this bound, the theoretical framework cannot guarantee sequence separability, regardless of the length of the context. However, this scenario is considered unlikely in practical applications, as it would require cross-branch correlations to systematically exceed intra-sequence temporal correlations.

\begin{figure}[htbp]
  \centering
  \includegraphics[width=0.98\linewidth]{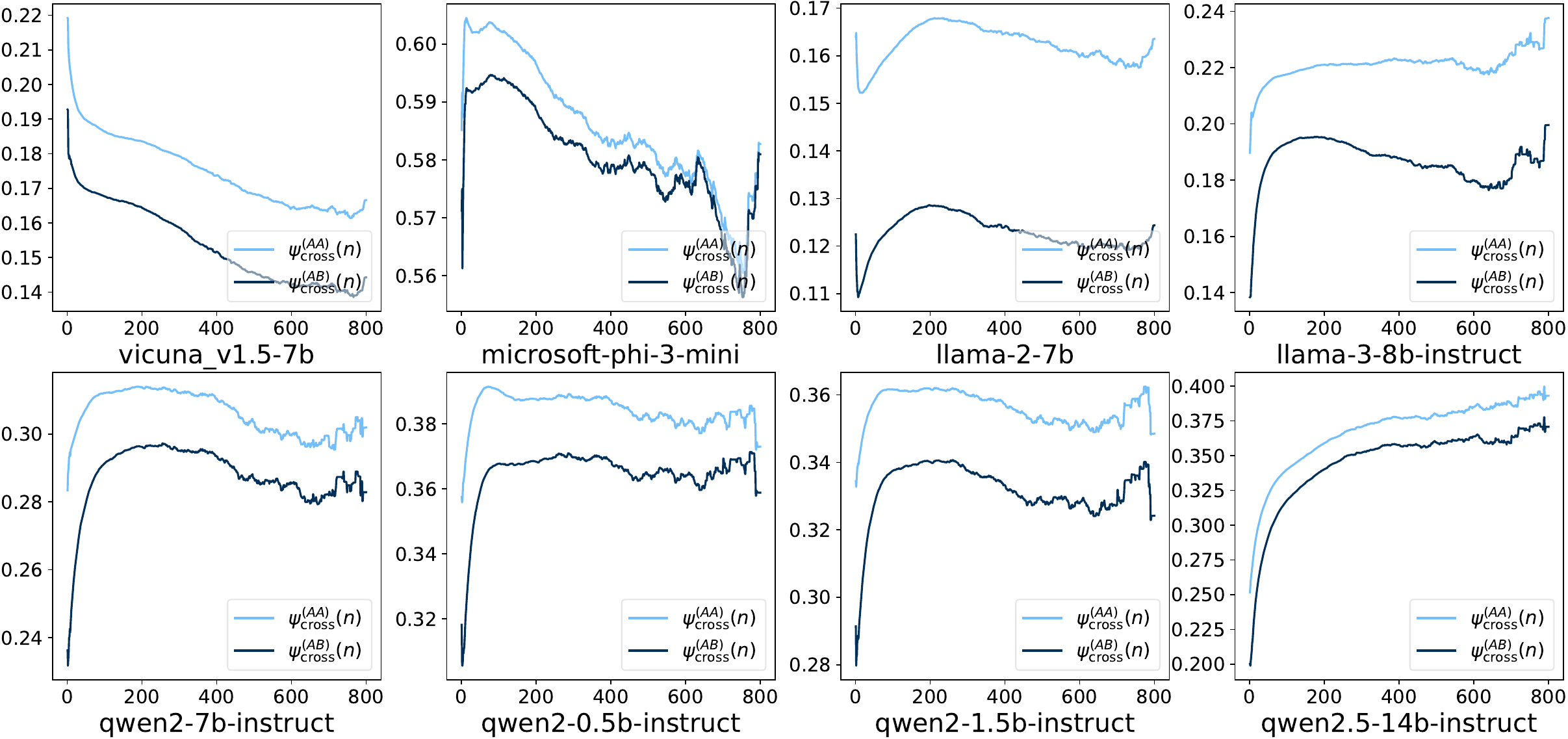}
  \caption{Estimated \(\psi_{\tcross}^{(AA)}(n)\) and \(\psi_{\tcross}^{AB}(n)\) for various large language models.}
  \label{fig:compare_psi_aabb_gt_ab}
\end{figure}

To provide a direct empirical validation of the theoretical relationships, the estimated values of \(\psi_{\tcross}^{(AA)}(n)\) and \(\psi_{\tcross}^{(AB)}(n)\) are compared in Figure \ref{fig:compare_psi_aabb_gt_ab}. 
The experimental configuration follows the same settings established in Subsection \ref{subsec:dtsa-empirical-estimation}, utilizing the identical results.
As demonstrated in Figure \ref{fig:compare_psi_aabb_gt_ab}, the empirically estimated \(\psi_{\tcross}^{(AA)}(n)\) consistently exceeds \(\psi_{\tcross}^{(AB)}(n)\) across all experimental conditions, thereby confirming the validity of the theoretical framework presented above.

\subsection{Expression for Scaling Exponent}
\label{appendix:scaling_exponent}

This appendix provides additional analysis for the scaling relationship \(\gamma \Upsilon(n)^{\beta/2} = c / n^{\alpha(n)}\), and express a closed-form expression for scaling constant \(c\) and scaling exponent \(\alpha(n)\).
\(\gamma\) is the scaling factor for matching \(S(n)\) and \(\Upsilon(n)^{\beta/2}\),
and \(\beta > 0\) measures the exponent relationship between \(\sqrt{\Upsilon(n)}\) and \(c/n^{\alpha(n)}\).

A special situation arises when \(n=1\), where the quadratic component in \(\Upsilon(n)\) vanishes,
therefore, without loss of generality, it can be considered that 
\begin{equation}
  c = \gamma \Upsilon(1)^{\beta/2} = \gamma \left(1 - \psi_{\tequal}^{(AB)}(1)\right)^{\frac{\beta}{2}}.
\end{equation}

1. If \(\psi_{\tcross}^{(AA)}(n) - \psi_{\tcross}^{(AB)}(n) = 0\) for all \(n > 1\), the scaling exponent is derived as follows:
\begin{equation*}
  \begin{aligned}
  \dfrac{c}{n^{\alpha(n)}}
  &= \gamma \left( n \left(1 - \psi_{\tequal}^{(AB)}(n)\right) \right)^{\frac{\beta}{2}}, \\
  \log \dfrac{c}{\gamma} - \alpha(n) \log n 
  &= \dfrac{\beta}{2}\left(
    \log n + \log\left(1 - \psi_{\tequal}^{(AB)}(n)\right)
  \right). \\
  \end{aligned}
\end{equation*}
The solution is therefore given by
\begin{equation}
  \label{equ:alpha_n_cross_aa_bb_zero_wrt_n}
  \alpha(n)
  = \log_{n} \dfrac{c}{\gamma} 
  - \dfrac{\beta}{2}
  - \dfrac{\beta}{2} \log_{n} \left(1 - \psi_{\tequal}^{(AB)}(n)\right).
\end{equation}

2. If \(\psi_{\tcross}^{(AA)}(n) - \psi_{\tcross}^{(AB)}(n) > 0\) for all \(n > 1\), the following relationships hold,
\begin{equation*}
  \begin{aligned}
    \dfrac{c}{n^{\alpha(n)}}
    &= \gamma \left(
      n \left(1 - \psi_{\tequal}^{(AB)}(n)\right)
      + (n^2 - n) \left( \psi_{\tcross}^{(AA)}(n) - \psi_{\tcross}^{(AB)}(n) \right)
    \right)^{\frac{\beta}{2}}, \\
    \log\dfrac{c}{\gamma} - \alpha(n) \log n
    &= \dfrac{\beta}{2} \log n + \dfrac{\beta}{2} \log\left(
      1 - \psi_{\tequal}^{(AB)}(n)
      + (n - 1) \left( \psi_{\tcross}^{(AA)}(n) - \psi_{\tcross}^{(AB)}(n) \right)
    \right). \\
  \end{aligned}
\end{equation*}
Therefore, \(\alpha(n)\) becomes,
\begin{equation*}
  \alpha(n) 
  = \log_{n} \frac{c}{\gamma}
  - \dfrac{\beta}{2} 
  - \dfrac{\beta}{2} \log_{n} \left(
    1 - \psi_{\tequal}^{(AB)}(n)
    + (n - 1) \left( \psi_{\tcross}^{(AA)}(n) - \psi_{\tcross}^{(AB)}(n) \right)
  \right).
\end{equation*}
The logarithmic term in the expression can be reformulated by factoring out the quadratic coefficient,
\begin{equation*}
  \begin{aligned}
    &\log_{n} \left(
      1 - \psi_{\tequal}^{(AB)}(n)
      + (n - 1) \left( \psi_{\tcross}^{(AA)}(n) - \psi_{\tcross}^{(AB)}(n) \right)
    \right) \\
    &= \log_{n} \left( \psi_{\tcross}^{(AA)}(n) - \psi_{\tcross}^{(AB)}(n) \right)
    + \log_{n} \left(
      n + \dfrac{
        1 - \psi_{\tequal}^{(AB)}(n)
      }{
        \psi_{\tcross}^{(AA)}(n) - \psi_{\tcross}^{(AB)}(n)
      } - 1 
    \right) \\
    &= \log_{n} \left( \psi_{\tcross}^{(AA)}(n) - \psi_{\tcross}^{(AB)}(n) \right)
    + \log_{n} \left(
      n + \left. n \right|_{\rho(n) = 1}
    \right),
  \end{aligned}
\end{equation*}
where the final equality comes from the definition of the critical balance point in Equation \eqref{equ:expected_divergence_bound_sqrt_balance} with slightly abused notation.
Based on this decomposition, the scaling exponent \(\alpha(n)\) can be expressed in a closed form as:
\begin{equation}
  \label{equ:alpha_n_cross_aa_bb_positive_wrt_n}
  \alpha(n)
  = \log_{n} \dfrac{c}{\gamma}
  - \dfrac{\beta}{2}
  - \dfrac{\beta}{2}
   \left(
    \log_{n} \left( \psi_{\tcross}^{(AA)}(n) - \psi_{\tcross}^{(AB)}(n) \right)
    + \log_{n} \left( n + \left. n \right|_{\rho(n) = 1} \right)
  \right).
\end{equation}

To simply the analysis, the same constant setting introduced in Subsection \ref{subsec:dtsa-properties-divergence-scaling} is applied: \(\psi_{\tequal}^{(AB)}(n) \rightarrow \psi_{\tequal}^{(AB)}\), \(\psi_{\tcross}^{(AA)}(n) \rightarrow \psi_{\tcross}^{(AA)}\), and \(\psi_{\tcross}^{(AB)}(n) \rightarrow \psi_{\tcross}^{(AB)}\).
With this simplification, 
\begin{equation*}
  c = \gamma\left(1 - \psi_{\tequal}^{(AB)}\right)^{\frac{\beta}{2}},
\end{equation*}
and the scaling exponent \(\alpha(n)\) can also be simplified.

For \(n > 1\) and \(\psi_{\tcross}^{(AA)} - \psi_{\tcross}^{(AB)} = 0\), the scaling exponent becomes,
\begin{equation}
  \label{equ:sim_scaling_exponent_constant_psi_cross_aa_ab_zero}
  \begin{aligned}
    \alpha(n)
    &= \log_{n} \dfrac{c}{\gamma} - \dfrac{\beta}{2} - \dfrac{\beta}{2} \log_{n} \left(1 - \psi_{\tequal}^{(AB)}\right) \\
    &= \dfrac{\beta}{2} \log_{n} \left(1 - \psi_{\tequal}^{(AB)}\right) - \dfrac{\beta}{2} - \dfrac{\beta}{2} \log_{n} \left(1 - \psi_{\tequal}^{(AB)}\right) \\
    &= - \dfrac{\beta}{2}.
  \end{aligned}
\end{equation}
For \(n > 1\) and \(\psi_{\tcross}^{(AA)} - \psi_{\tcross}^{(AB)} > 0\), the scaling exponent is given by,
\begin{equation}
  \label{equ:sim_scaling_exponent_constant_psi_cross_aa_ab_positive}
  \begin{aligned}
    \alpha(n)
    &= \log_{n} \dfrac{c}{\gamma}
      - \dfrac{\beta}{2}
      - \dfrac{\beta}{2}
      \left(
        \log_{n} \left( \psi_{\tcross}^{(AA)} - \psi_{\tcross}^{(AB)} \right)
        + \log_{n} \left( n + \left. n \right|_{\rho(n) = 1} \right)
      \right) \\
    &= \dfrac{\beta}{2} \log_{n} \left(1 - \psi_{\tequal}^{(AB)}\right)
    - \dfrac{\beta}{2}
    - \dfrac{\beta}{2}
    \left(
      \log_{n} \left( \psi_{\tcross}^{(AA)} - \psi_{\tcross}^{(AB)} \right)
      + \log_{n} \left( n + \left. n \right|_{\rho(n) = 1} \right)
    \right) \\
    &= \dfrac{\beta}{2} \left(
      \log_{n} \left(1 - \psi_{\tequal}^{(AB)}\right)
      - \log_{n} \left( \psi_{\tcross}^{(AA)} - \psi_{\tcross}^{(AB)} \right)
      - \log_{n} \left( n + \left. n \right|_{\rho(n) = 1} \right)
      - 1 
    \right) \\
    &= \dfrac{\beta}{2} \left(
      \log_{n} \left(1 + \left. n \right|_{\rho(n) = 1}\right) 
      - \log_{n} \left( n + \left. n \right|_{\rho(n) = 1} \right)
      - 1 
    \right) \\
    &= \dfrac{\beta}{2} 
      \log_{n} \left( \frac{
        1 + \left. n \right|_{\rho(n) = 1}
      }{
        n + \left. n \right|_{\rho(n) = 1}
      } \right) - \dfrac{\beta}{2}. \\
  \end{aligned}
\end{equation}
This \(\alpha(n)\) is a monotonically decreasing function of \(n\), and its value will approach to \(-\beta\) as \(n\) increase.
The gradient of \(\alpha(n)\) has the following form,
\begin{equation*}
  \begin{aligned}
  \dfrac{d \alpha(n)}{d n}
  &= \dfrac{\beta}{2} \dfrac{
    - \frac{1}{n + \left. n \right|_{\rho(n) = 1}} \log n 
    - \frac{1}{n} \log\left(
      \frac{1+\left. n \right|_{\rho(n) = 1}}{n +\left. n \right|_{\rho(n) = 1}}
    \right)
  }{
    \left(\log n\right)^{2}
  } \\
  &\le \dfrac{- \beta}{
    2 \left(\log n\right)^{2} \left(n + \left. n \right|_{\rho(n) = 1}\right)
  }
  \log \left(
    \frac{n \left(1+\left. n \right|_{\rho(n) = 1}\right)}{n +\left. n \right|_{\rho(n) = 1}}
  \right) \\
  &= \dfrac{- \beta}{
    2 \left(\log n\right)^{2} \left(n + \left. n \right|_{\rho(n) = 1}\right)
  }
  \log \left(
    \frac{
      1 + \left. n \right|_{\rho(n) = 1}
    }{
      1 + \frac{\left. n \right|_{\rho(n) = 1}}{n}
    }
  \right). \\
  \end{aligned}
\end{equation*}
Since \(n > 1\) and \(1 + \left. n \right|_{\rho(n) = 1} > 1 + \frac{\left. n \right|_{\rho(n) = 1}}{n}\), the logarithmic term is positive. Therefore, \(\frac{d \alpha(n)}{d n} < 0\) and \(\alpha(n)\) is monotonically decreasing.
The asymptotic behavior as \(n \rightarrow \infty\) is characterized by
\begin{equation*}
  \begin{aligned}
    \lim_{n \rightarrow \infty} \alpha(n)
    &= -\frac{\beta}{2}
      + \lim_{n \rightarrow \infty} 
      \frac{\beta}{2 \log n} \log \left(
        1 + \left. n \right|_{\rho(n) = 1}
      \right)
      - \lim_{n \rightarrow \infty} 
      \frac{\beta}{2 \log n} \log \left(
        n + \left. n \right|_{\rho(n) = 1}
      \right) \\
    &= -\frac{\beta}{2}
      - \lim_{n \rightarrow \infty} 
      \frac{\beta}{2} \frac{1}{1 + \frac{\left. n \right|_{\rho(n) = 1}}{n}} 
      \\
    &= -\beta.
  \end{aligned}
\end{equation*}

\subsection{Empirical Distribution of Cosine Similarities}
\label{appendix:empirical-distribution-cosine-similarities}

To provide empirical validation for the theoretical framework, particularly the dependency assumptions based on cosine similarity distributions, the empirical distributions of cosine similarities are analyzed. The experimental configurations follow those established in Subsection \ref{subsec:dtsa-empirical-estimation}.

The cosine similarities \(\cos(\Theta)_{ii}^{(AB)}\) and \(\cos(\Theta)_{ij}^{(AB)}\) are computed, and their distributions are examined through histogram analysis.
As discussed in Subsection \ref{subsec:dtsa-empirical-estimation}, the inherent structure of the dataset results in branch A containing all `chosen' responses and branch B containing all `rejected' responses.
Consequently, \(\cos(\Theta)_{ij}^{(AA)}\) is estimated as the average of cosine similarities within the respective response categories,
\begin{equation*}
  \cos(\Theta)_{ij}^{(AA)} \leftarrow \frac{1}{2} \left(\cos(\Theta)_{ij}^{(++)} + \cos(\Theta)_{ij}^{(--)}\right),
\end{equation*}
where \(\cos(\Theta)_{ij}^{(++)}\) represents cosine similarities between `chosen' responses and \(\cos(\Theta)_{ij}^{(--)}\) represents cosine similarities between `rejected' responses.

\begin{figure}[tbp]
  \centering
  \begin{subfigure}{0.86\linewidth}
    \centering
    \includegraphics[width=\linewidth]{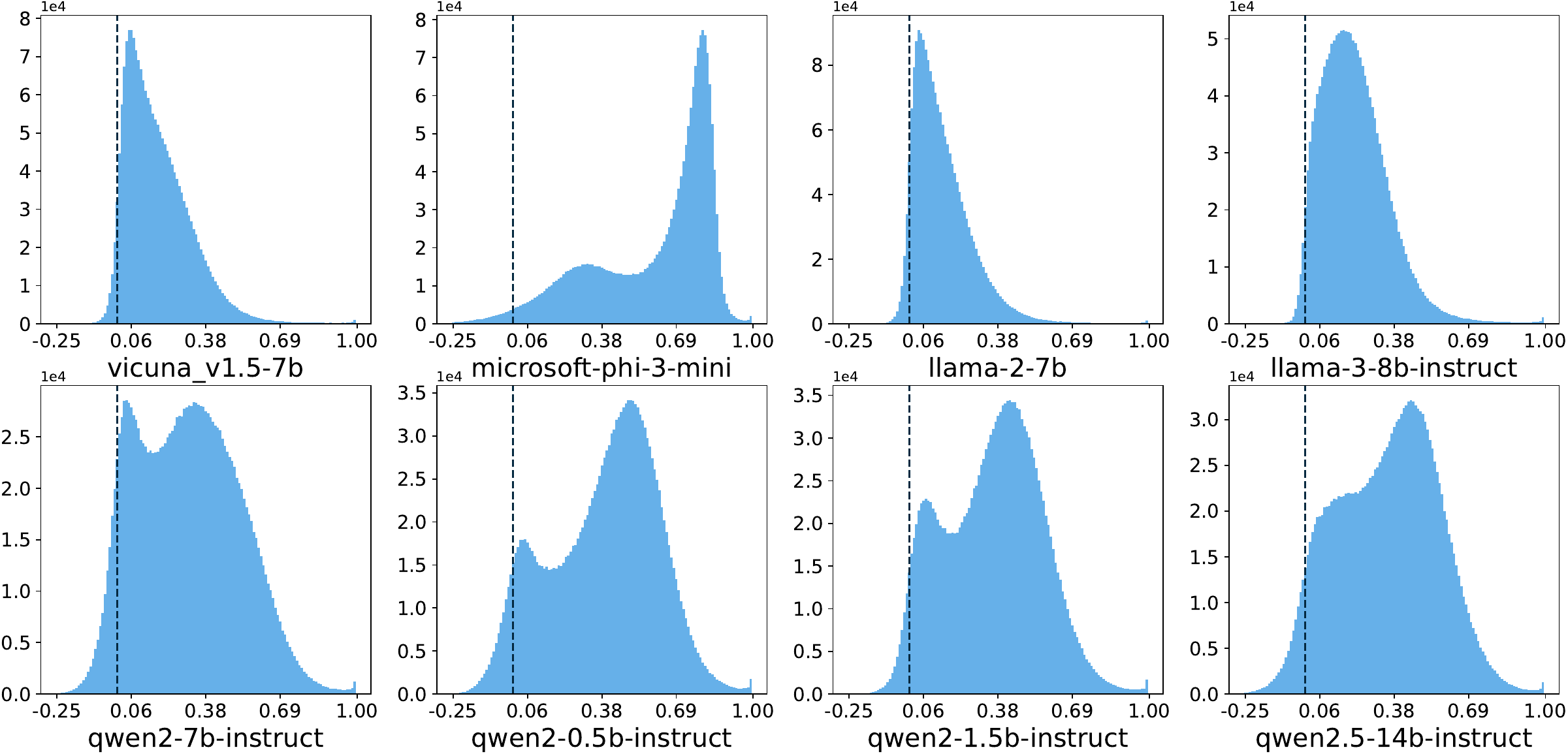}
    \caption{Histogram of \(\cos(\Theta)_{ii}^{(AB)}\)}
    \label{fig:hist_E_cos_theta_equal_ab}
  \end{subfigure}
  \begin{subfigure}{0.86\linewidth}
    \centering
    \includegraphics[width=\linewidth]{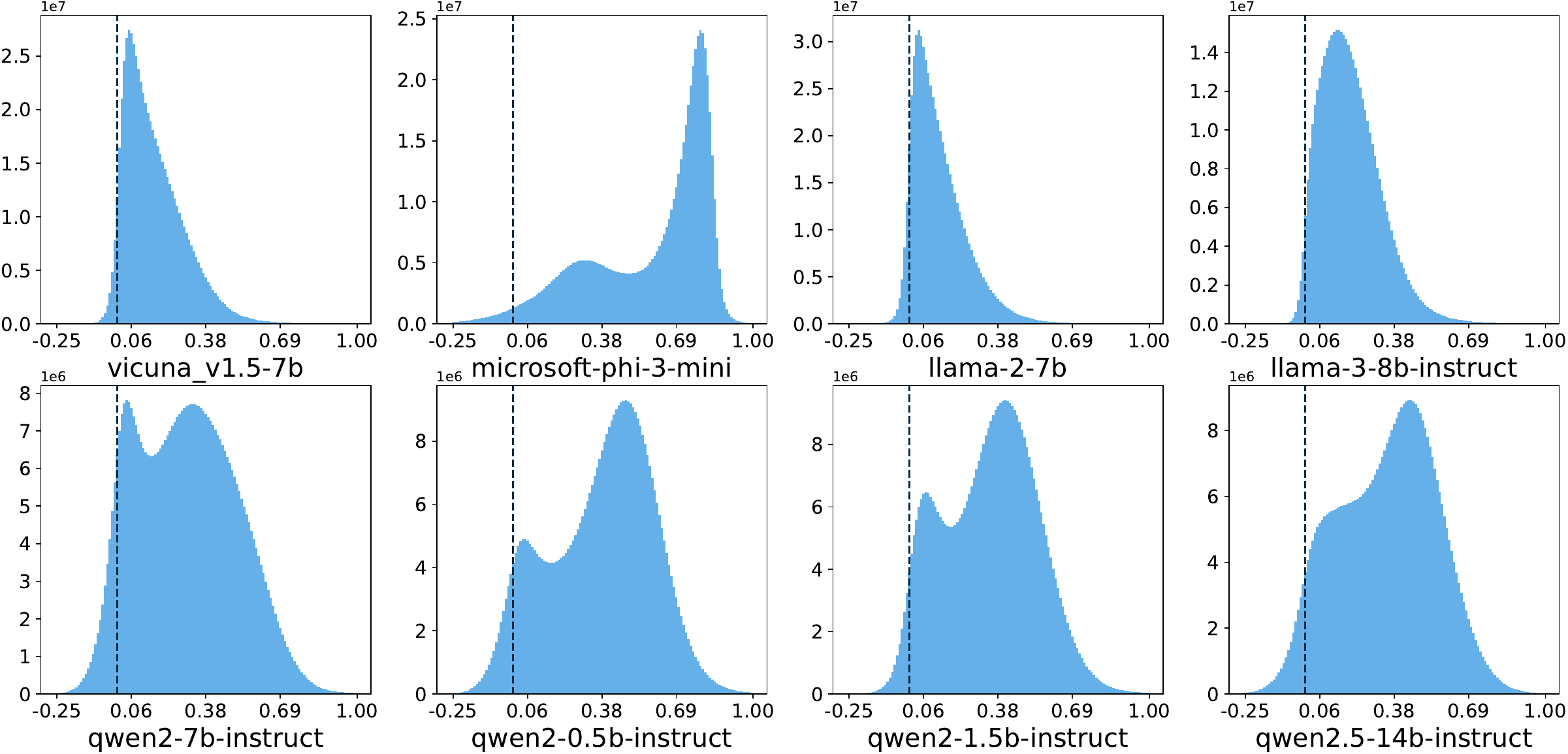}
    \caption{Histogram of \(\cos(\Theta)_{ij}^{(AB)}\)}
    \label{fig:hist_E_cos_theta_cross_ab}
  \end{subfigure}
  \begin{subfigure}{0.86\linewidth}
    \centering
    \includegraphics[width=\linewidth]{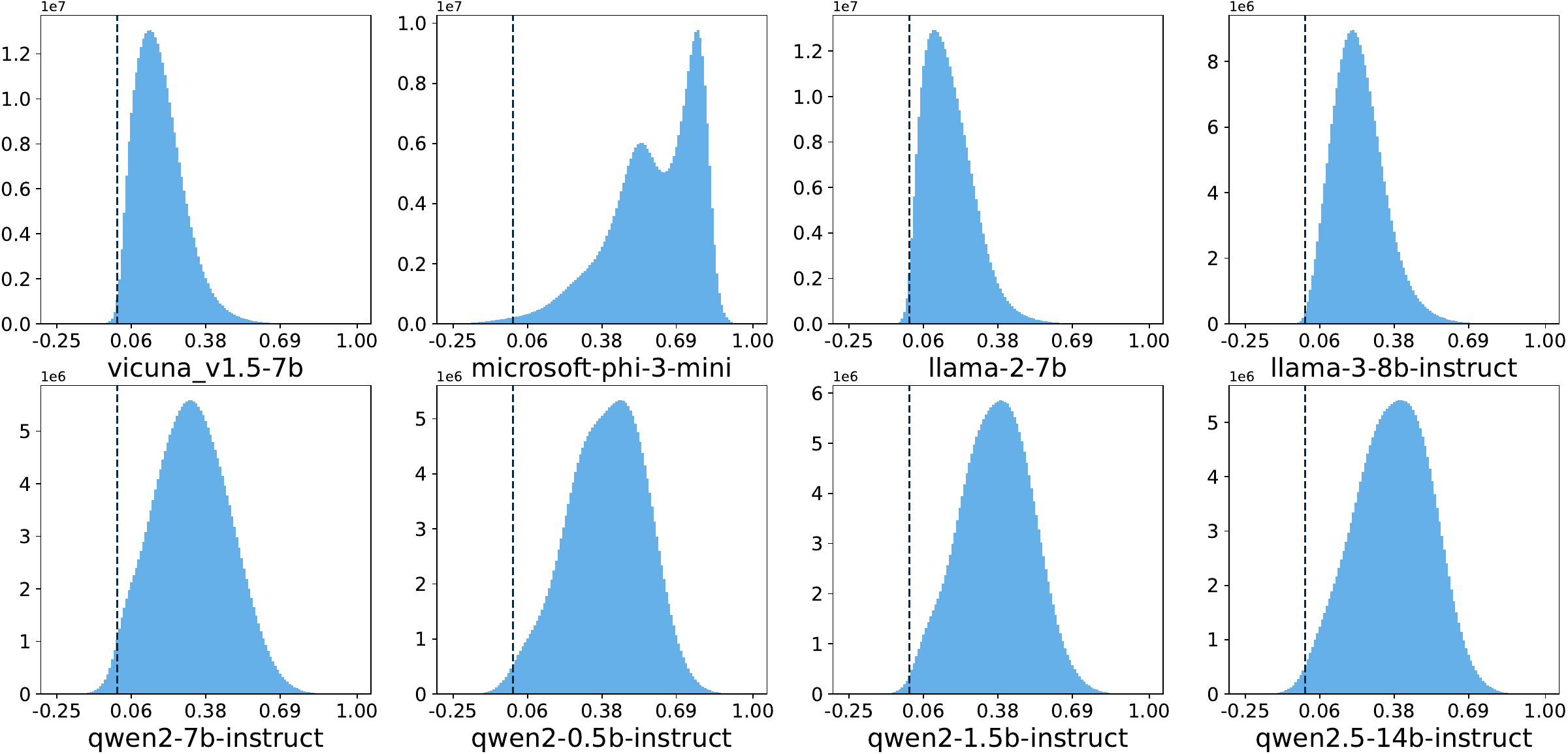}
    \caption{Histogram of \(\frac{1}{2} \left(\cos(\Theta)_{ij}^{(AA)} + \cos(\Theta)_{ij}^{(BB)}\right)\)}
    \label{fig:hist_E_cos_theta_cross_aabb}
  \end{subfigure}
  \caption{Histograms of cosine from Inequality~\eqref{equ:E_diff_hijAB_expand_cos_M}. The dashed lines represent \(\cos \theta = 0\).}
  \label{fig:hist_E_cos_theta}
\end{figure}

Figure \ref{fig:hist_E_cos_theta} presents the empirical distributions of these cosine similarities. The analysis reveals several key observations:
First, the vast majority of cosine similarity values are positive, supporting the theoretical assumption that representations maintain directional coherence.
Second, different base models exhibit distinct distributional characteristics, suggesting model-specific dependency patterns.
Third, the distributions provide empirical evidence for the validity of the dependency measures defined in Definition \ref{def:dependency_measures}, as the observed similarity patterns align with the theoretical expectations.

\section{Datasets}
\label{appendix:datasets}

The training utilizes a combination of existing datasets and newly generated data to enhance performance across various tasks.
The primary datasets employed include LLAVA V1.5 MIX665K \citep{liu2024llava}, BAAI-SVIT \citep{zhao2023svit}, and mPLUG DocDownstream 1.0 \citep{ye2023mplug-docowl}.

To further augment the training data, we generated new question-and-answer pairs using detailed descriptions from BAAI-SVIT, with the Llama-3 8B model serving as the agent.
The generated questions were of two types: yes/no questions and multiple-choice questions (e.g., selecting one option from A, B, C, or D).
To maintain data quality, any generated text that did not conform to the correct format was excluded from the dataset.

Another source of augmented training data comes from the structure-aware parsing task in CCpdf, derived from mPLUG DocDownstream 1.0.
To enhance data quality, we utilized the nougat-base model \citep{blecher2023nougat} to convert a subset of images into better-formatted text.
To ensure the validity of the outputs, we applied several filtering criteria:
(1) generated texts that were too short were discarded,
(2) cases where the length difference between the original and generated text was excessively large were excluded,
(3) generated texts containing endless repeated patterns were removed, and
(4) texts that lacked sufficient similarity to the original under the embedding space, as measured using all-mpnet-base-v2 \citep{reimers-2019-sentence-bert}, were filtered out.
These steps ensured that the final dataset maintained high quality and relevance for training.

\section{Implementation Details}
\label{appendix:implementation_details}

Our implementation is built upon two widely used open-source pretrained models: CLIP ViT-H/14 \citep{radford2021CLIP} and Llama-2 7B \citep{touvron2023llama-2}.
The CLIP ViT-H/14 vision encoder processes image inputs with a resolution of \((336, 336)\), using a stride of 14, resulting in \(24 \times 24 = 576\) tokens, each with a hidden size of 1024.
The Llama-2 7B model, serving as the backbone language model, features a hidden size of 4096.

The experiments were conducted on high-performance hardware comprising 8 NVIDIA A100 GPUs, each with 40 GB of memory.
For evaluation on more accessible hardware, we utilized NVIDIA RTX A6000 GPUs, each with 48 GB of memory.

The image preprocessing pipeline involves several steps to adapt raw pixel images for input to the CLIP vision encoder.
First, as introduced in subsection \ref{subsec:main_architecture-proposed_method}, the images are processed into global and local views, with each view independently passed through the CLIP encoder to generate vision tokens.

Based on the available hardware resources, we adopt an HD-9 cropping strategy, which generates nine local image views and one global image view as input to the CLIP vision encoder.
Each view has a resolution of \((336, 336)\), resulting in a maximum input image resolution of \((1008 \times 1008)\).
This setup initially produces \(5760\) tokens, which is computationally expensive to process.
To address this, we apply a local merging operation that reduces the token count by a factor of four.
Neighboring vision tokens are merged in a \(4 \times 4 \Rightarrow 1\) manner using a 2D convolution with a kernel size of 3, padding of 1, and a stride of 2.
This reduces the number of vision tokens to \(1440\).
After appending \token{new line} tokens to mark the end of each view, the total token count becomes \(1450\).
The tokens for all views are then concatenated, ensuring efficient processing while retaining critical information.

As introduced in subsection \ref{subsec:main_architecture-proposed_method}, the fused model integrates the vision tokens from the global and local views with the text tokens from the user's question and the learnable queries.
For the first question, \(n_l\) learnable queries are appended, while subsequent questions are each followed by \(n_s\) learnable queries.
The value of \(n_l\) ranges from 1 to 768, with \(n_s\) set to \(8\) when \(n_l \ge 8\).
For smaller values of \(n_l\) (i.e., \(n_l < 8\)), \(n_s\) is set equal to \(n_l\).
For example, when only one learnable query is used, both \(n_l\) and \(n_s\) are set to \(1\).
The text tokenizer is based on the CLIP text tokenizer, and \(n_l\) new special tokens are added to the tokenizer to support the learnable queries.

The hidden size of the fused model is set to 1024, matching the hidden size of the vision tokens generated by CLIP ViT-H/14.
This is significantly smaller than the hidden size of Llama-2 7B, which is 4096.
Adopting a smaller hidden size helps reduce both memory consumption and computational cost, making the model more efficient for training and inference.

To align the fused vision tokens generated by the fused model with the Llama-2 7B backbone, a linear projection layer is used to map the fused vision tokens to the hidden size of the Llama-2 model.
We consider a three-step training strategy for optimizing the model.

In the first step, we perform preliminary training using the contrastive loss introduced in Equation \ref{equ:loss_contrastive} to align the fused vision tokens with the CLIP text encoder.
Since the maximum sequence length of the CLIP text encoder is 77 and most questions exceed this limit, we extend the sequence length to 512 to accommodate longer inputs.
The modules involved in this step include the vision encoder, the 2D convolution layer for merging neighboring vision tokens, the fused model, and the CLIP text encoder.
This step requires only a few training steps, using a batch size of 32 per device, gradient accumulation of 1, an equivalent batch size of 256, and a learning rate of 2e-5.
The training employs a cosine learning rate scheduler with a warm-up ratio of 0.1 and is performed over 1000 steps.

The second step involves fine-tuning on the full dataset.
The required modules include the vision encoder, the 2D convolution layer, the fused model, a linear projection layer to map the fused vision tokens to the Llama-2 hidden size, and the Llama-2 7B backbone.
In this stage, the Llama-2 backbone is frozen, and the remaining modules are updated.
The training uses a batch size of 5 per device, gradient accumulation of 64, and an equivalent batch size of 2560, with a learning rate of 2e-5.
A cosine learning rate scheduler with a warm-up ratio of 0.03 is employed.
The training spans two epochs and uses the generation loss described in Equation \ref{equ:loss_generative_ce}.
The fusion model consists of 20 layers, with \(n_l = 256\) learnable queries for the first question and \(n_s = 8\) for subsequent questions.
Given the considerable size of the dataset, the fine-tuning process is estimated to require more than 200 hours.

The third step focuses on further fine-tuning to evaluate various configurations of \(n_l\) and \(n_s\), such as \(n_l = 128, n_s = 8\) and \(n_l = 64, n_s = 8\).
To reduce the substantial time required for full-data fine-tuning, we reuse the model trained in the second step (\(n_l = 256, n_s = 8\)) and fine-tune it using only 10\% of randomly sampled training data, which will take approximately 10 hours.
This includes fine-tuning for alternative configurations as well as the original \(n_l = 256, n_s = 8\).
This strategy significantly reduces computational cost while allowing for the exploration of multiple settings.
The loss function remains the same as in the second step (Equation \ref{equ:loss_generative_ce}), with the Llama-2 backbone frozen and updates applied only to the other modules.
All other hyperparameters and configurations are consistent with those used in the second step.

Another critical aspect of the fusion model is assessing the impact of incorporating the user's question on the model's performance.
To evaluate this, we introduce an alternative setup where the model processes inputs without including the user's question.
In this configuration, the input is limited to vision tokens and learnable queries, excluding any text-based queries provided by the user.
To determine the validity of fusing the user's question, we conduct experiments both with or without further fine-tuning the model, keeping the Llama-2 backbone frozen in both cases.
For this evaluation, we reuse the 10\% randomly sampled training data previously employed for tuning different configurations of \(n_l\).
All other hyperparameters and configurations are kept consistent with those used in the second step.
This approach allows us to efficiently analyze the influence of the user's question on the fusion model's performance while maintaining computational efficiency.

\section{Detailed Results}
\label{appendix:detailed-results}

\begin{figure}[htb]
  \centering
  \begin{subfigure}{0.46\linewidth}
    \centering
    \includegraphics[width=\linewidth]{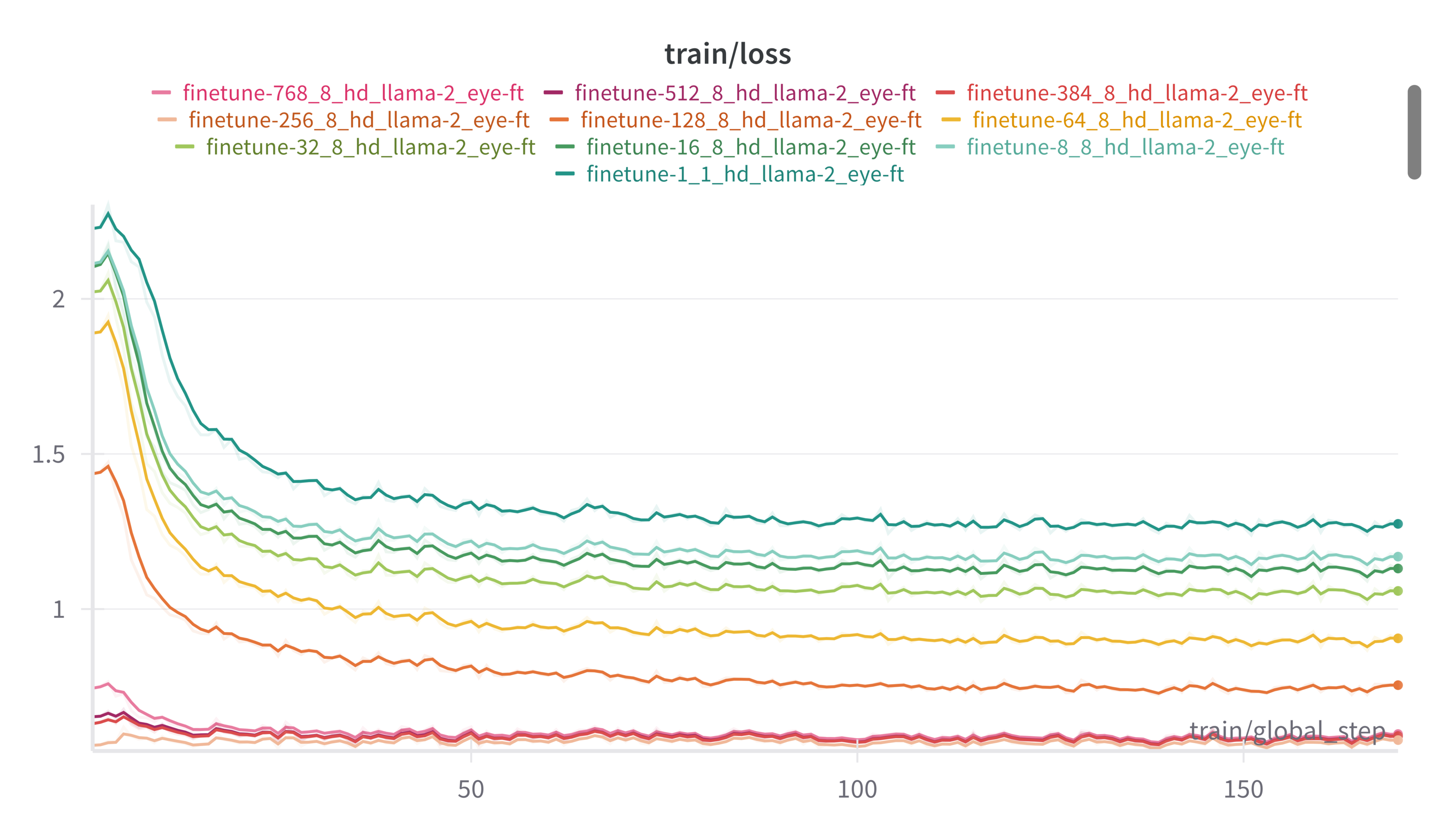}
    \caption{Losses of ``Vision Query Queries''.}
    \label{fig:finetune-llama2-loss}
  \end{subfigure}
  \begin{subfigure}{0.46\linewidth}
    \centering
    \includegraphics[width=\linewidth]{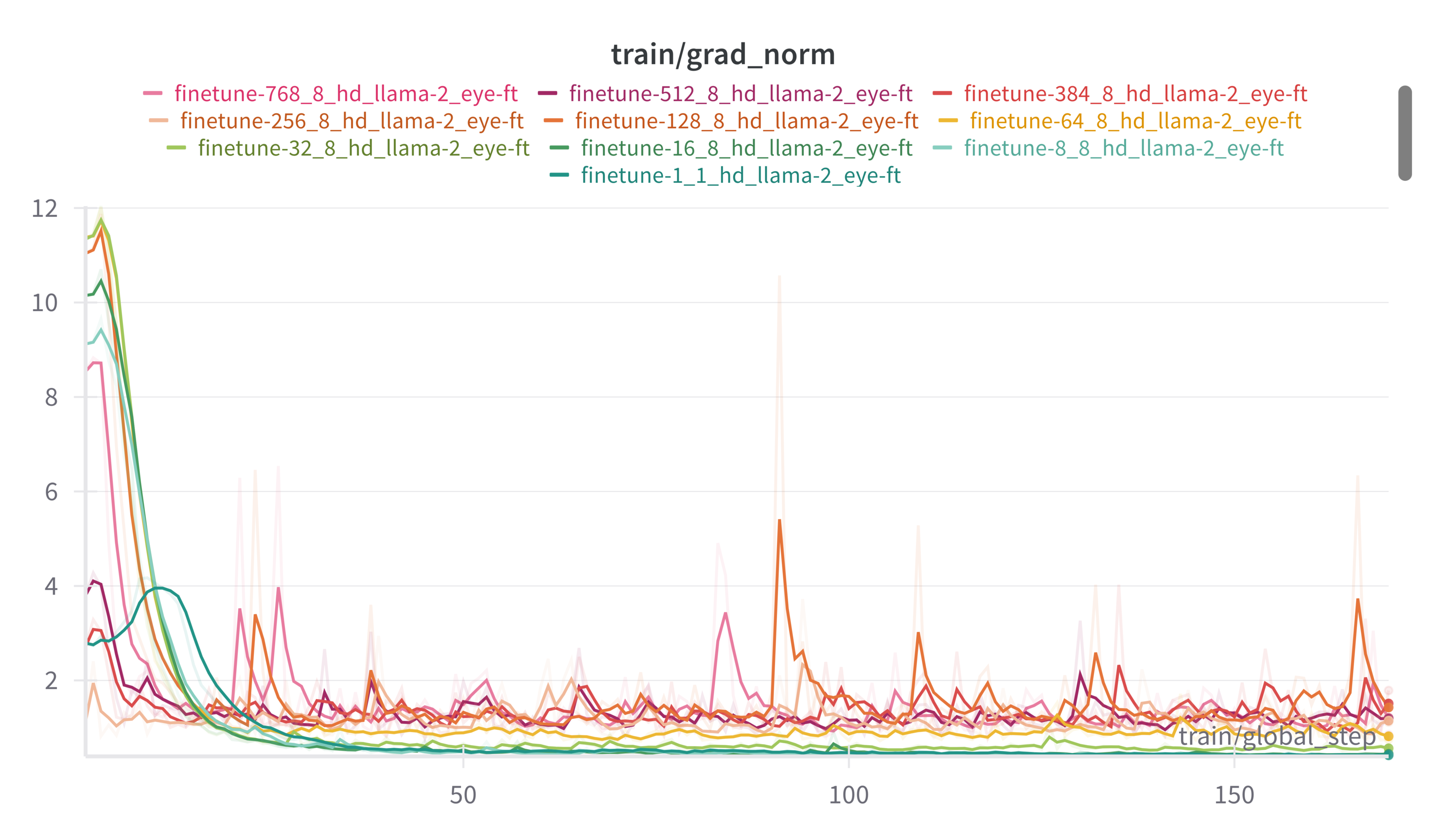}
    \caption{Gradient norms of ``Vision Query Queries''.}
    \label{fig:finetune-llama2-grad_norm}
  \end{subfigure}
  \begin{subfigure}{0.46\linewidth}
    \centering
    \includegraphics[width=\linewidth]{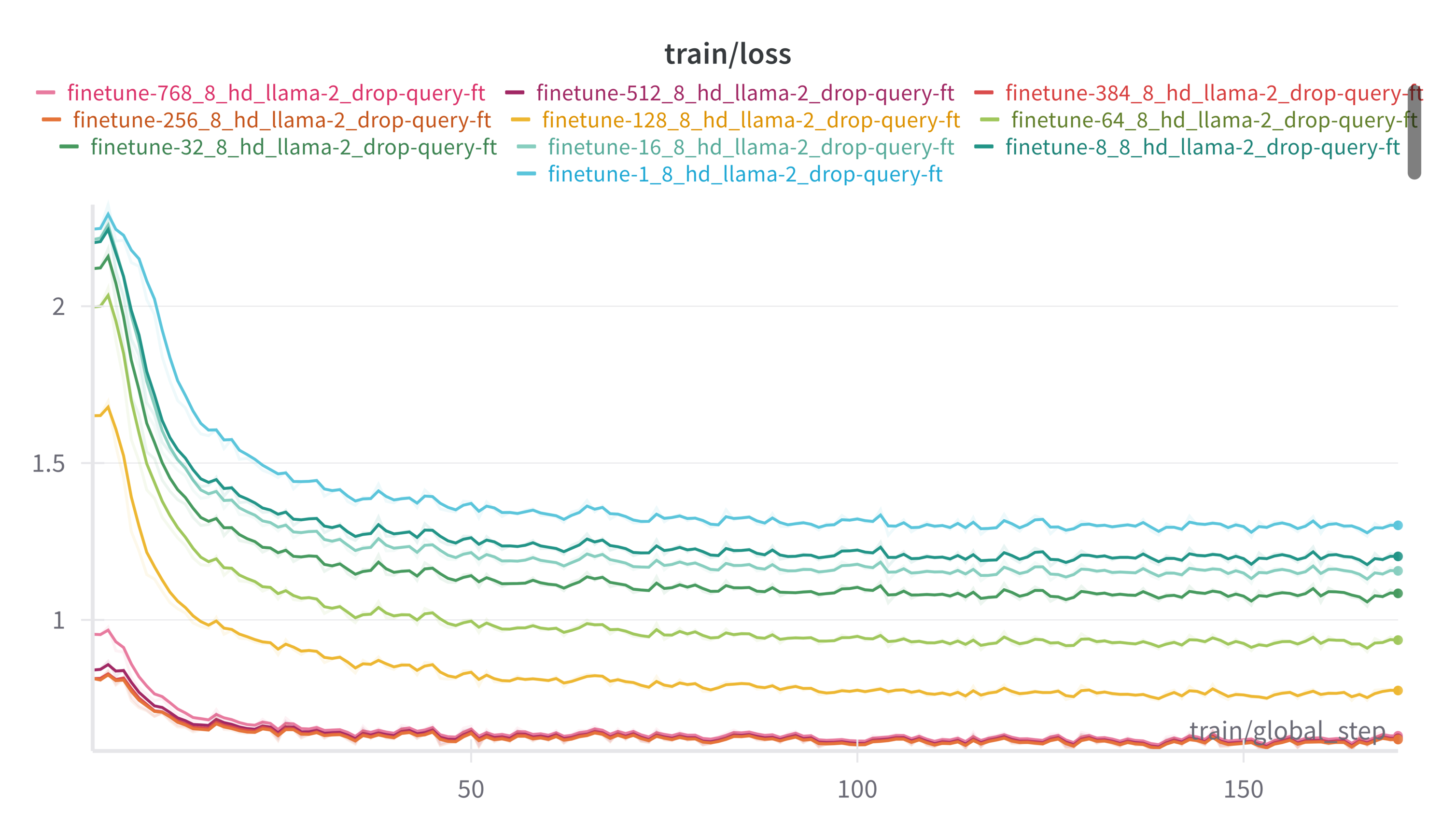}
    \caption{Losses of ``Vision Query (ft)''.}
    \label{fig:finetune-llama2-drop-query-loss}
  \end{subfigure}
  \begin{subfigure}{0.46\linewidth}
    \centering
    \includegraphics[width=\linewidth]{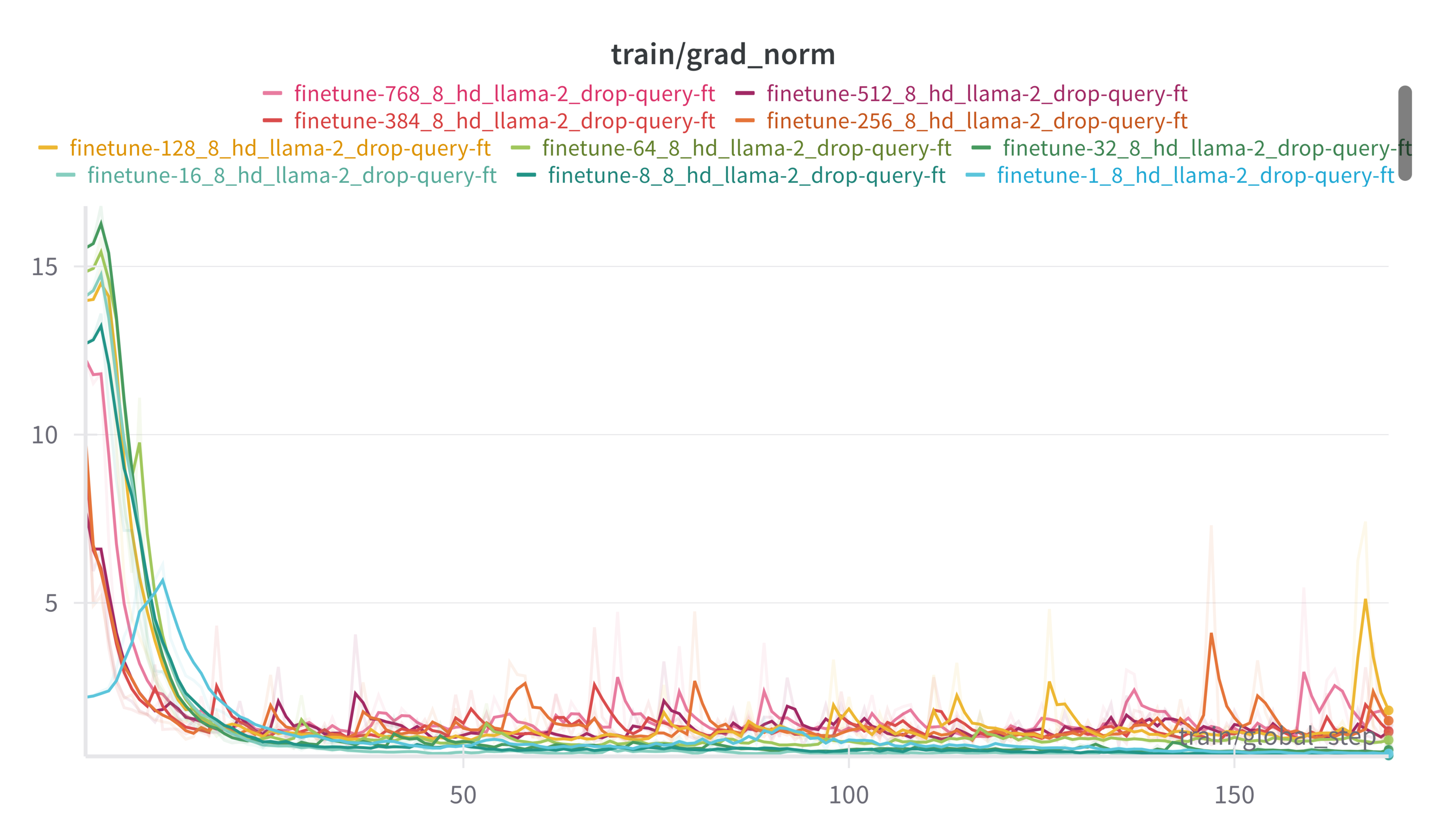}
    \caption{Gradient norms of ``Vision Query (ft)''.}
    \label{fig:finetune-llama2-drop-query-grad_norm}
  \end{subfigure}
  \caption{Loss and gradient norm of further fine-tuned model with different configurations.}
  \label{fig:finetune-llama2-loss-grad_norm}
\end{figure}

For evaluation, we utilize the VLMEvalKit \citep{duan2024vlmevalkit} to compute scores across various benchmarks.
Different configurations of \(n_l\) and \(n_s\) are tested, including: \(768(8)\), \(512(8)\), \(384(8)\), \(256(8)\), \(128(8)\), \(64(8)\), \(32(8)\), \(16(8)\), \(8(8)\), \(1(1)\), where the first number represents \(n_l\) and the second number in parentheses denotes \(n_s\).
The losses and gradient norms of the further fine-tuned models are shown in Figure \ref{fig:finetune-llama2-loss-grad_norm}.

The evaluation spans a diverse set of benchmarks, including MME \citep{fu2023mme}, HallusionBench \citep{HallusionBench}, POPE \citep{POPE}, OCRBench \citep{ocrbench}, COCO VAL \citep{COCO}, RealWorldQA\footnote{\url{https://x.ai/blog/grok-1.5v}}, MMStar \citep{MMStart}, SEEDBench IMG \citep{li2023seed}, SEEDBench2 \citep{li2023seed2}, SEEDBench2 Plus \citep{li2024seed2plus}, ScienceQA TEST \citep{ScienceQA}, AI2D TEST\footnote{\url{https://allenai.org/data/diagrams}}, OCRVQA TESTCORE \citep{OCRVQA}, ChartQA TEST \citep{masry-etal-2022-chartqa}, and TextVQA VAL \citep{TextVQA}.

The baseline models are:
360VL-70B \footnote{\url{https://github.com/360CVGroup/360VL}},
InstructBLIP-7B \citep{Dai2023InstructBLIPTG},
InternLM-XComposer2-4KHD \citep{dong2024InternLM-xcomposer2-4khd},
LLaVA-v1-7B, LLaVA-v1.5-13B, LLaVA-v1.5-7B \citep{liu2024llava},
MiniGPT-4-v1-7B, MiniGPT-4-v2 \citep{zhu2023minigpt-4},
mPLUG-Owl2 \citep{ye2024mplug2},
OpenFlamingo v2 \citep{alayrac2022flamingo},
Phi-3-Vision \citep{abdin2024phi-3},
Qwen-VL-Chat \citep{Qwen-vl},
Fuyu-8B \citep{fuyu-8b} (jointly pretrained with vision data),
GPT-4o (0513, detail-low) \citep{openai2024gpt4ocard}.
Results are summarized in the following tables, with all benchmark data for the models obtained from VLMEvalKit \citep{duan2024vlmevalkit}.
The blank cells indicate that the scores are not available from VLMEvalKit.

In the column titled ``Fused Model Inputs,'' the following terms are used to describe the input configurations of the model:
\begin{itemize}
  \item ``Vision Query Queries'' refers to the model that takes vision tokens, text tokens from the user's question, and learnable queries as inputs.
  \item ``Vision Query (ft)'' refers to the model that takes vision tokens and learnable queries as inputs, with the model further fine-tuned without the user's question.
  \item ``Vision Query'' refers to the model that takes vision tokens and learnable queries as inputs, but remains the same as the original model, without any additional fine-tuning.
\end{itemize}

It is important to note that the training datasets for these models differ, and as such, the results are provided for reference purposes only.

\begin{table}[tbp]
  \setlength{\tabcolsep}{2pt}
  \renewcommand{\arraystretch}{0.8}
  \small
  \centering
  \caption{Score comparison of POPE.}
  \noindent\makebox[\textwidth]{
    \begin{tabular}{lcrrrr}
      \toprule
      Fused Model Inputs                             &
      $n_l$                                          &
      Overall                                        &
      acc                                            &
      precision                                      &
      recall
      \\
      \midrule
      Vision Question Queries                        & 768    & 86.977 & 87.033 & 87.357 & 86.600 \\
      Vision Queries (ft)                            & 768    & 85.742 & 85.456 & 84.085 & 87.467 \\
      Vision Queries                                 & 768    & 80.158 & 76.500 & 69.362 & 94.933 \\
      Vision Question Queries                        & 512    & 86.626 & 86.444 & 85.482 & 87.800 \\
      Vision Queries (ft)                            & 512    & 84.990 & 84.400 & 81.891 & 88.333 \\
      Vision Queries                                 & 512    & 74.835 & 66.978 & 60.451 & 98.200 \\
      Vision Question Queries                        & 384    & 85.254 & 84.433 & 80.984 & 90.000 \\
      Vision Queries (ft)                            & 384    & 86.080 & 86.233 & 87.048 & 85.133 \\
      Vision Queries                                 & 384    & 83.572 & 82.178 & 77.508 & 90.667 \\
      Vision Question Queries                        & 256    & 86.734 & 87.233 & 90.267 & 83.467 \\
      Vision Queries (ft)                            & 256    & 84.447 & 83.289 & 78.975 & 90.733 \\
      Vision Queries                                 & 256    & 75.719 & 68.767 & 61.933 & 97.400 \\
      Vision Question Queries                        & 128    & 84.016 & 82.878 & 78.778 & 90.000 \\
      Vision Queries (ft)                            & 128    & 84.727 & 84.822 & 85.261 & 84.200 \\
      Vision Queries                                 & 128    & 83.255 & 82.522 & 80.331 & 86.400 \\
      Vision Question Queries                        & 64     & 82.761 & 81.489 & 77.440 & 88.867 \\
      Vision Queries (ft)                            & 64     & 82.398 & 80.756 & 76.029 & 89.933 \\
      Vision Queries                                 & 64     & 82.476 & 81.756 & 79.343 & 85.867 \\
      Vision Question Queries                        & 32     & 81.232 & 80.300 & 77.562 & 85.267 \\
      Vision Queries (ft)                            & 32     & 78.539 & 74.278 & 67.377 & 94.133 \\
      Vision Queries                                 & 32     & 76.593 & 78.933 & 86.167 & 68.933 \\
      Vision Question Queries                        & 16     & 79.533 & 77.200 & 72.150 & 88.600 \\
      Vision Queries (ft)                            & 16     & 77.338 & 72.944 & 66.533 & 92.333 \\
      Vision Queries                                 & 16     & 77.879 & 75.800 & 71.717 & 85.200 \\
      Vision Question Queries                        & 8      & 77.749 & 74.167 & 68.280 & 90.267 \\
      Vision Queries (ft)                            & 8      & 76.634 & 71.933 & 65.734 & 91.867 \\
      Vision Queries                                 & 8      & 72.383 & 63.422 & 58.140 & 95.867 \\
      Vision Question Queries                        & 1      & 58.113 & 49.689 & 49.778 & 69.800 \\
      Vision Queries (ft)                            & 1      & 65.648 & 52.033 & 51.134 & 91.667 \\
      Vision Queries                                 & 1      & 64.091 & 58.678 & 56.679 & 73.733 \\
      \midrule
      \multicolumn{2}{c}{Model Name}                                                              \\
      \midrule
      \multicolumn{2}{c|}{LLaVA-v1.5-13B}            & 88.400 & 88.600 & 89.600 & 87.300          \\
      \multicolumn{2}{c|}{360VL-70B}                 & 87.300 & 88.200 & 94.700 & 81.100          \\
      \multicolumn{2}{c|}{InstructBLIP-7B}           & 86.100 & 86.000 & 85.700 & 86.500          \\
      \multicolumn{2}{c|}{LLaVA-v1.5-7B}             & 86.100 & 87.000 & 92.100 & 80.900          \\
      \multicolumn{2}{c|}{GPT-4o (0513, detail-low)} & 85.000 & 86.200 & 93.300 & 78.100          \\
      \multicolumn{2}{c|}{mPLUG-Owl2}                & 84.600 & 85.400 & 89.800 & 79.900          \\
      \multicolumn{2}{c|}{Mantis-8B-Fuyu}            & 84.000 & 84.000 & 84.400 & 83.500          \\
      \multicolumn{2}{c|}{Phi-3-Vision}              & 83.700 & 85.600 & 96.100 & 74.100          \\
      \multicolumn{2}{c|}{LLaVA-v1-7B}               & 75.900 & 69.200 & 62.300 & 96.900          \\
      \multicolumn{2}{c|}{Qwen-VL-Chat}              & 74.900 & 67.300 & 67.800 & 83.600          \\
      \multicolumn{2}{c|}{MiniGPT-4-v2}              & 60.000 & 48.400 & 49.900 & 75.300          \\
      \multicolumn{2}{c|}{OpenFlamingo v2}           & 52.600 & 42.400 & 50.000 & 55.500          \\
      \multicolumn{2}{c|}{MiniGPT-4-v1-7B}           & 34.600 & 36.400 & 58.400 & 24.600          \\
      \multicolumn{2}{c|}{InternLM-XComposer2-4KHD}  & 2.900  & 10.100 & 53.700 & 1.500           \\
      \bottomrule
    \end{tabular}
  }
\end{table}

\begin{table}[tbp]
  \setlength{\tabcolsep}{2pt}
  \renewcommand{\arraystretch}{0.8}
  \small
  \centering
  \caption{Score comparison of MME, HallusionBench, COCO VAL.}
  \noindent\makebox[\textwidth]{
    \begin{tabular}{lc|rrr|rrrr|rrrr}
      \toprule
                                                     &
                                                     &
      \multicolumn{3}{c|}{MME}                       &
      \multicolumn{4}{c|}{HallusionBench}            &
      \multicolumn{4}{c}{COCO VAL}
      \\
      Fused Model Inputs                             &
      $n_l$                                          &
      \multicolumn{1}{c}{\rotatebox{90}{Overall}}    &
      \multicolumn{1}{c}{\rotatebox{90}{Perception}} &
      \multicolumn{1}{c|}{\rotatebox{90}{Cognition}} &
      \multicolumn{1}{c}{\rotatebox{90}{aAcc}}       &
      \multicolumn{1}{c}{\rotatebox{90}{fAcc}}       &
      \multicolumn{1}{c}{\rotatebox{90}{qAcc}}       &
      \multicolumn{1}{c|}{\rotatebox{90}{Overall}}   &
      \multicolumn{1}{c}{\rotatebox{90}{BLEU-1}}     &
      \multicolumn{1}{c}{\rotatebox{90}{BLEU-4}}     &
      \multicolumn{1}{c}{\rotatebox{90}{ROUGE-L}}    &
      \multicolumn{1}{c}{\rotatebox{90}{CIDEr}}
      \\
      \midrule
      Vision Question Queries                        & 768      & 1461.206 & 1202.635 & 258.571 & 50.683 & 22.254 & 17.143 & 30.027 & 40.892 & 10.282 & 34.858 & 12.897 \\
      Vision Queries (ft)                            & 768      & 1483.059 & 1244.845 & 238.214 & 50.053 & 18.786 & 16.044 & 28.294 & 49.980 & 13.713 & 42.126 & 36.652 \\
      Vision Queries                                 & 768      & 1268.119 & 1072.405 & 195.714 & 34.805 & 12.428 & 10.330 & 19.188 & 41.113 & 10.291 & 34.400 & 8.308  \\
      Vision Question Queries                        & 512      & 1414.270 & 1165.699 & 248.571 & 50.578 & 21.387 & 16.703 & 29.556 & 40.707 & 10.570 & 34.702 & 15.256 \\
      Vision Queries (ft)                            & 512      & 1402.772 & 1182.058 & 220.714 & 51.314 & 17.052 & 17.363 & 28.576 & 55.407 & 16.420 & 44.891 & 50.358 \\
      Vision Queries                                 & 512      & 1045.693 & 860.693  & 185.000 & 29.232 & 10.694 & 9.890  & 16.605 & 39.360 & 9.647  & 32.752 & 6.008  \\
      Vision Question Queries                        & 384      & 1409.437 & 1152.294 & 257.143 & 54.784 & 21.676 & 21.978 & 32.813 & 38.037 & 9.740  & 32.416 & 7.763  \\
      Vision Queries (ft)                            & 384      & 1340.833 & 1112.976 & 227.857 & 52.156 & 20.809 & 18.901 & 30.622 & 50.765 & 14.282 & 42.812 & 38.826 \\
      Vision Queries                                 & 384      & 1304.939 & 1085.296 & 219.643 & 46.057 & 15.607 & 16.484 & 26.049 & 39.129 & 9.844  & 32.932 & 4.672  \\
      Vision Question Queries                        & 256      & 1472.529 & 1209.315 & 263.214 & 48.791 & 18.786 & 14.945 & 27.507 & 38.787 & 9.838  & 33.043 & 9.108  \\
      Vision Queries (ft)                            & 256      & 1472.806 & 1225.663 & 247.143 & 50.473 & 21.098 & 18.242 & 29.938 & 46.695 & 12.835 & 40.083 & 31.333 \\
      Vision Queries                                 & 256      & 1223.633 & 1018.990 & 204.643 & 30.284 & 10.116 & 8.352  & 16.250 & 38.147 & 9.280  & 31.678 & 5.103  \\
      Vision Question Queries                        & 128      & 1396.146 & 1157.574 & 238.571 & 51.104 & 18.786 & 18.022 & 29.304 & 37.189 & 9.383  & 31.461 & 5.491  \\
      Vision Queries (ft)                            & 128      & 1315.989 & 1095.632 & 220.357 & 49.527 & 15.896 & 15.824 & 27.082 & 50.710 & 14.330 & 42.714 & 39.148 \\
      Vision Queries                                 & 128      & 1215.086 & 1001.872 & 213.214 & 32.492 & 10.983 & 8.132  & 17.202 & 37.400 & 9.181  & 30.822 & 4.903  \\
      Vision Question Queries                        & 64       & 1328.092 & 1095.235 & 232.857 & 48.160 & 16.185 & 15.604 & 26.650 & 37.055 & 9.207  & 31.348 & 6.703  \\
      Vision Queries (ft)                            & 64       & 1126.880 & 889.023  & 237.857 & 48.686 & 15.029 & 15.604 & 26.440 & 46.324 & 12.352 & 39.644 & 27.363 \\
      Vision Queries                                 & 64       & 1178.733 & 1003.733 & 175.000 & 30.599 & 8.960  & 6.374  & 15.311 & 37.565 & 9.321  & 31.435 & 5.299  \\
      Vision Question Queries                        & 32       & 1321.832 & 1099.689 & 222.143 & 42.587 & 10.405 & 9.451  & 20.814 & 35.466 & 8.110  & 29.854 & 5.743  \\
      Vision Queries (ft)                            & 32       & 1045.688 & 830.331  & 215.357 & 45.321 & 12.139 & 10.110 & 22.523 & 39.049 & 9.191  & 33.465 & 11.910 \\
      Vision Queries                                 & 32       & 1039.101 & 835.887  & 203.214 & 25.026 & 6.358  & 4.615  & 12.000 & 36.205 & 8.608  & 30.269 & 4.443  \\
      Vision Question Queries                        & 16       & 1291.082 & 1066.082 & 225.000 & 43.113 & 13.006 & 7.692  & 21.270 & 37.606 & 8.754  & 32.330 & 15.509 \\
      Vision Queries (ft)                            & 16       & 1145.157 & 938.729  & 206.429 & 44.585 & 12.717 & 11.209 & 22.837 & 44.580 & 11.090 & 38.103 & 27.403 \\
      Vision Queries                                 & 16       & 1066.146 & 889.003  & 177.143 & 27.340 & 8.382  & 4.835  & 13.519 & 36.562 & 8.452  & 30.886 & 9.226  \\
      Vision Question Queries                        & 8        & 1228.312 & 992.954  & 235.357 & 42.482 & 13.873 & 9.451  & 21.935 & 35.793 & 8.040  & 31.152 & 13.552 \\
      Vision Queries (ft)                            & 8        & 1133.447 & 905.590  & 227.857 & 42.376 & 13.295 & 10.110 & 21.927 & 43.824 & 9.748  & 37.006 & 25.907 \\
      Vision Queries                                 & 8        & 1020.312 & 841.383  & 178.929 & 34.490 & 5.491  & 6.813  & 15.598 & 39.338 & 9.401  & 33.294 & 21.574 \\
      Vision Question Queries                        & 1        & 992.141  & 734.284  & 257.857 & 44.059 & 11.561 & 12.747 & 22.789 & 27.612 & 4.022  & 24.384 & 1.899  \\
      Vision Queries (ft)                            & 1        & 891.690  & 679.190  & 212.500 & 41.956 & 10.116 & 8.352  & 20.141 & 27.142 & 3.086  & 24.364 & 1.516  \\
      Vision Queries                                 & 1        & 887.341  & 690.556  & 196.786 & 25.342 & 5.780  & 4.835  & 11.986 & 29.373 & 4.749  & 25.674 & 4.887  \\
      \midrule
      \multicolumn{2}{c}{Model Name}                                                                                                                                    \\
      \midrule
      \multicolumn{2}{c|}{360VL-70B}                 & 2009.700 & 1646.200 & 363.600  & 54.600  & 26.900 & 23.100 & 34.800 & 71.000 & 28.100 & 53.000 & 86.600          \\
      \multicolumn{2}{c|}{GPT-4o (0513, detail-low)} & 2328.700 & 1609.400 & 719.300  & 67.500  & 46.500 & 41.100 & 51.700 &        &        &        &                 \\
      \multicolumn{2}{c|}{InternLM-XComposer2-4KHD}  & 2130.400 & 1581.500 & 548.900  & 60.700  & 33.800 & 33.000 & 42.500 & 20.200 & 4.900  & 17.700 & 6.000           \\
      \multicolumn{2}{c|}{LLaVA-v1.5-7B}             & 1808.400 & 1506.200 & 302.100  & 48.800  & 20.500 & 13.600 & 27.600 & 19.800 & 4.700  & 20.000 & 0.000           \\
      \multicolumn{2}{c|}{LLaVA-v1.5-13B}            & 1780.800 & 1502.600 & 278.200  & 45.300  & 17.100 & 11.000 & 24.500 & 20.700 & 5.100  & 21.400 & 0.400           \\
      \multicolumn{2}{c|}{Qwen-VL-Chat}              & 1860.000 & 1467.800 & 392.100  & 56.400  & 27.700 & 26.400 & 36.800 & 75.800 & 34.000 & 54.900 & 98.900          \\
      \multicolumn{2}{c|}{mPLUG-Owl2}                & 1786.400 & 1436.000 & 350.400  & 48.900  & 22.500 & 16.700 & 29.400 & 25.800 & 7.100  & 33.600 & 35.000          \\
      \multicolumn{2}{c|}{Phi-3-Vision}              & 1508.000 & 1205.100 & 302.900  & 56.800  & 29.500 & 30.800 & 39.000 & 15.800 & 2.900  & 16.100 & 0.000           \\
      \multicolumn{2}{c|}{InstructBLIP-7B}           & 1391.400 & 1137.100 & 254.300  & 53.600  & 20.200 & 19.800 & 31.200 & 56.800 & 20.900 & 39.900 & 58.100          \\
      \multicolumn{2}{c|}{Mantis-8B-Fuyu}            & 1321.600 & 1057.700 & 263.900  & 50.800  & 20.800 & 17.800 & 29.800 & 18.500 & 3.700  & 18.200 & 0.000           \\
      \multicolumn{2}{c|}{LLaVA-v1-7B}               & 1075.500 & 832.000  & 243.600  & 43.600  & 13.000 & 8.100  & 21.600 & 27.000 & 6.700  & 26.400 & 5.500           \\
      \multicolumn{2}{c|}{MiniGPT-4-v1-7B}           & 1047.400 & 770.600  & 276.800  & 52.400  & 17.300 & 25.900 & 31.900 & 19.600 & 4.300  & 17.500 & 0.800           \\
      \multicolumn{2}{c|}{MiniGPT-4-v2}              & 968.400  & 708.400  & 260.000  & 52.600  & 16.500 & 21.100 & 30.000 & 12.600 & 1.400  & 13.300 & 0.100           \\
      \multicolumn{2}{c|}{OpenFlamingo v2}           & 607.200  & 535.000  & 72.100   & 52.700  & 17.600 & 18.000 & 29.400 & 6.400  & 1.300  & 15.800 & 14.900          \\
      \bottomrule
    \end{tabular}
  }
\end{table}

\begin{table}[tbp]
  \setlength{\tabcolsep}{2pt}
  \renewcommand{\arraystretch}{0.8}
  \small
  \centering
  \caption{Score comparison of OCRBench, AI2D TEST, RealWorldQA, MMStar, SEEDBench IMG, SEEDBench2, SEEDBench2 Plus, ScienceQA TEST, OCRVQA TESTCORE, ChartQA TEST, TextVQA VAL.}
  \noindent\makebox[\textwidth]{
    \begin{tabular}{lcrrrrrrrrrrr}
      \toprule
      Fused Model Inputs                                 &
      $n_l$                                              &
      \multicolumn{1}{c}{\rotatebox{90}{
      \parbox{3.9cm}{Final Score \newline OCRBench}}}    &
      \multicolumn{1}{c}{\rotatebox{90}{
      \parbox{3.9cm}{Overall \newline AI2D TEST}}}       &
      \multicolumn{1}{c}{\rotatebox{90}{
      \parbox{3.9cm}{Overall \newline RealWorldQA}}}     &
      \multicolumn{1}{c}{\rotatebox{90}{
      \parbox{3.9cm}{Overall \newline MMStar}}}          &
      \multicolumn{1}{c}{\rotatebox{90}{
      \parbox{3.9cm}{Overall \newline SEEDBench IMG}}}   &
      \multicolumn{1}{c}{\rotatebox{90}{
      \parbox{3.9cm}{Overall \newline SEEDBench2}}}      &
      \multicolumn{1}{c}{\rotatebox{90}{
      \parbox{3.9cm}{Overall \newline SEEDBench2 Plus}}} &
      \multicolumn{1}{c}{\rotatebox{90}{
      \parbox{3.9cm}{Overall \newline ScienceQA TEST}}}  &
      \multicolumn{1}{c}{\rotatebox{90}{
      \parbox{3.9cm}{Overall \newline OCRVQA TESTCORE}}} &
      \multicolumn{1}{c}{\rotatebox{90}{
      \parbox{3.9cm}{Overall \newline ChartQA TEST}}}    &
      \multicolumn{1}{c}{\rotatebox{90}{
          \parbox{3.9cm}{Overall \newline TextVQA VAL}}}
      \\
      \midrule
      Vision Question Queries                            & 768     & 493.000 & 41.192 & 50.850 & 35.200 & 53.598 & 41.804 & 45.059 & 55.776 & 36.068 & 48.160 & 52.998 \\
      Vision Queries (ft)                                & 768     & 465.000 & 37.953 & 43.660 & 30.200 & 46.944 & 35.230 & 38.384 & 54.189 & 37.370 & 41.760 & 48.522 \\
      Vision Queries                                     & 768     & 470.000 & 26.295 & 37.255 & 27.733 & 38.245 & 29.761 & 30.391 & 45.216 & 2.181  & 3.800  & 38.474 \\
      Vision Question Queries                            & 512     & 472.000 & 40.835 & 49.150 & 35.400 & 53.338 & 41.775 & 43.390 & 57.164 & 38.021 & 51.480 & 53.550 \\
      Vision Queries (ft)                                & 512     & 483.000 & 37.500 & 31.895 & 25.600 & 37.001 & 30.064 & 39.833 & 43.877 & 40.267 & 40.560 & 49.612 \\
      Vision Queries                                     & 512     & 502.000 & 25.712 & 35.948 & 28.000 & 37.149 & 28.825 & 26.263 & 42.985 & 1.367  & 6.800  & 34.884 \\
      Vision Question Queries                            & 384     & 445.000 & 39.411 & 40.000 & 35.467 & 52.066 & 39.806 & 44.049 & 53.842 & 30.794 & 43.200 & 53.400 \\
      Vision Queries (ft)                                & 384     & 496.000 & 37.727 & 28.497 & 25.400 & 33.635 & 28.222 & 37.901 & 44.125 & 40.234 & 40.400 & 48.856 \\
      Vision Queries                                     & 384     & 477.000 & 31.574 & 47.451 & 28.400 & 40.289 & 31.004 & 32.060 & 49.926 & 10.189 & 28.320 & 44.720 \\
      Vision Question Queries                            & 256     & 448.000 & 39.475 & 45.621 & 34.133 & 52.108 & 40.072 & 39.438 & 52.900 & 43.294 & 50.600 & 54.396 \\
      Vision Queries (ft)                                & 256     & 502.000 & 35.427 & 41.699 & 27.467 & 37.170 & 28.874 & 36.978 & 49.232 & 43.424 & 43.520 & 47.370 \\
      Vision Queries                                     & 256     & 498.000 & 27.655 & 34.118 & 26.867 & 35.216 & 25.838 & 26.921 & 47.595 & 3.027  & 5.600  & 40.820 \\
      Vision Question Queries                            & 128     & 438.000 & 33.614 & 45.882 & 30.333 & 47.625 & 37.713 & 27.800 & 48.438 & 38.607 & 41.320 & 49.472 \\
      Vision Queries (ft)                                & 128     & 456.000 & 29.534 & 45.359 & 27.867 & 41.779 & 32.034 & 25.384 & 48.290 & 37.044 & 36.200 & 46.774 \\
      Vision Queries                                     & 128     & 405.000 & 20.596 & 44.183 & 27.067 & 36.615 & 26.683 & 9.969  & 49.430 & 3.646  & 9.000  & 30.138 \\
      Vision Question Queries                            & 64      & 386.000 & 35.039 & 49.673 & 31.733 & 47.829 & 37.914 & 28.371 & 49.281 & 41.699 & 38.160 & 45.802 \\
      Vision Queries (ft)                                & 64      & 394.000 & 27.040 & 46.536 & 24.933 & 37.472 & 30.343 & 17.567 & 42.191 & 40.397 & 34.080 & 42.438 \\
      Vision Queries                                     & 64      & 390.000 & 18.199 & 36.471 & 24.467 & 35.954 & 28.797 & 10.716 & 44.571 & 2.637  & 7.800  & 35.918 \\
      Vision Question Queries                            & 32      & 320.000 & 35.460 & 49.412 & 31.800 & 45.025 & 36.507 & 30.347 & 50.273 & 30.924 & 29.680 & 42.600 \\
      Vision Queries (ft)                                & 32      & 314.000 & 25.777 & 45.359 & 25.067 & 31.984 & 27.032 & 25.736 & 40.654 & 31.999 & 26.320 & 37.910 \\
      Vision Queries                                     & 32      & 309.000 & 18.329 & 40.784 & 24.400 & 30.516 & 25.830 & 11.989 & 44.373 & 2.214  & 6.840  & 31.434 \\
      Vision Question Queries                            & 16      & 268.000 & 36.302 & 47.712 & 30.800 & 42.854 & 35.066 & 31.796 & 47.248 & 21.810 & 20.320 & 34.112 \\
      Vision Queries (ft)                                & 16      & 272.000 & 24.320 & 44.183 & 24.333 & 36.207 & 30.795 & 13.702 & 35.102 & 20.345 & 19.880 & 30.708 \\
      Vision Queries                                     & 16      & 266.000 & 14.508 & 39.869 & 21.867 & 31.366 & 24.890 & 7.905  & 46.653 & 2.734  & 4.840  & 30.236 \\
      Vision Question Queries                            & 8       & 221.000 & 37.144 & 45.882 & 30.467 & 42.938 & 35.370 & 33.114 & 50.273 & 16.895 & 15.520 & 27.886 \\
      Vision Queries (ft)                                & 8       & 226.000 & 25.810 & 42.484 & 24.333 & 31.844 & 25.058 & 15.152 & 34.655 & 12.695 & 14.880 & 21.636 \\
      Vision Queries                                     & 8       & 227.000 & 27.947 & 36.601 & 25.933 & 33.699 & 26.765 & 16.864 & 44.323 & 1.823  & 4.760  & 22.032 \\
      Vision Question Queries                            & 1       & 24.000  & 30.635 & 44.967 & 30.133 & 32.153 & 29.022 & 26.131 & 47.744 & 18.424 & 12.320 & 9.708  \\
      Vision Queries (ft)                                & 1       & 17.000  & 22.312 & 42.745 & 24.400 & 28.309 & 22.289 & 7.729  & 33.118 & 17.122 & 11.560 & 7.694  \\
      Vision Queries                                     & 1       & 23.000  & 24.028 & 40.000 & 25.733 & 34.865 & 26.474 & 15.195 & 54.239 & 9.440  & 2.640  & 4.930  \\
      \midrule
      \multicolumn{2}{c}{Model Name}                                                                                                                                   \\
      \midrule
      \multicolumn{2}{c|}{InternLM-XComposer2-4KHD}      & 675.000 &         & 63.300 & 55.300 & 74.600 & 57.400 & 65.200 & 96.300 &        &        &                 \\
      \multicolumn{2}{c|}{GPT-4o (0513, detail-low)}     & 663.000 &         & 68.600 & 61.600 & 76.400 &        & 64.000 & 90.100 &        &        &                 \\
      \multicolumn{2}{c|}{Phi-3-Vision}                  & 637.000 &         & 58.800 & 47.700 & 70.900 & 55.700 & 64.200 & 90.000 & 61.900 & 81.800 & 72.400          \\
      \multicolumn{2}{c|}{Qwen-VL-Chat}                  & 488.000 &         & 49.300 & 34.500 & 64.800 &        & 46.000 & 68.800 & 58.600 & 49.800 & 60.700          \\
      \multicolumn{2}{c|}{360VL-70B}                     & 397.000 &         & 62.400 & 48.100 & 73.100 &        & 52.300 & 87.400 &        &        &                 \\
      \multicolumn{2}{c|}{Mantis-8B-Fuyu}                & 366.000 &         & 43.700 & 34.400 & 59.300 &        & 33.600 & 56.800 & 54.200 & 53.300 & 49.000          \\
      \multicolumn{2}{c|}{LLaVA-v1.5-13B}                & 337.000 &         & 55.300 & 34.300 & 68.200 &        & 44.600 & 72.600 & 63.400 & 18.200 & 48.900          \\
      \multicolumn{2}{c|}{LLaVA-v1.5-7B}                 & 318.000 &         & 54.800 & 33.100 & 65.800 & 43.400 & 41.300 & 69.200 & 60.600 & 17.800 & 45.500          \\
      \multicolumn{2}{c|}{InstructBLIP-7B}               & 276.000 &         & 36.900 & 32.700 & 44.500 &        & 29.500 & 54.100 & 50.200 & 10.900 & 33.600          \\
      \multicolumn{2}{c|}{LLaVA-v1-7B}                   & 269.000 &         & 45.800 & 27.100 & 50.400 &        & 31.200 & 61.800 &        &        &                 \\
      \multicolumn{2}{c|}{mPLUG-Owl2}                    & 255.000 &         & 50.800 & 34.800 & 64.500 &        & 44.400 & 69.500 & 65.200 & 22.800 & 56.400          \\
      \multicolumn{2}{c|}{MiniGPT-4-v1-7B}               & 172.000 &         & 21.300 & 16.300 & 31.600 &        & 15.200 & 39.600 & 0.300  &        & 0.600           \\
      \multicolumn{2}{c|}{OpenFlamingo v2}               & 149.000 &         & 35.200 & 26.900 & 28.800 &        & 28.700 & 44.800 & 8.900  &        & 16.300          \\
      \multicolumn{2}{c|}{MiniGPT-4-v2}                  & 31.000  &         & 30.700 & 21.300 & 29.400 &        & 23.300 & 54.700 & 0.000  &        & 0.000           \\
      \bottomrule
    \end{tabular}
  }
\end{table}

\section{Comparison with Other Models And Computation Costs}
\label{appendix:compare_and_computation_costs}

\subsection{Comparison with Benchmark Models}

To evaluate the effectiveness of the proposed approach, comprehensive comparisons are conducted against established baseline models across multiple benchmarks. The baseline models considered for comparison include LLaVA-v1.5-7B \citep{liu2024llava}, mPLUG-Owl2 \citep{ye2024mplug2}, InstructBLIP-7B \citep{Dai2023InstructBLIPTG}, and OpenFlamingo v2 \citep{alayrac2022flamingo}.

It should be noted that these baseline models were trained on different datasets than the proposed model, rendering the comparisons indicative rather than directly equivalent. The performance evaluations are therefore presented for reference and contextual understanding of the model's capabilities.

\begin{figure}[tb]
  \centering
  \hspace*{-1em}%
  \begin{subfigure}{0.49\textwidth}
    \centering
    \includegraphics[width=\linewidth]{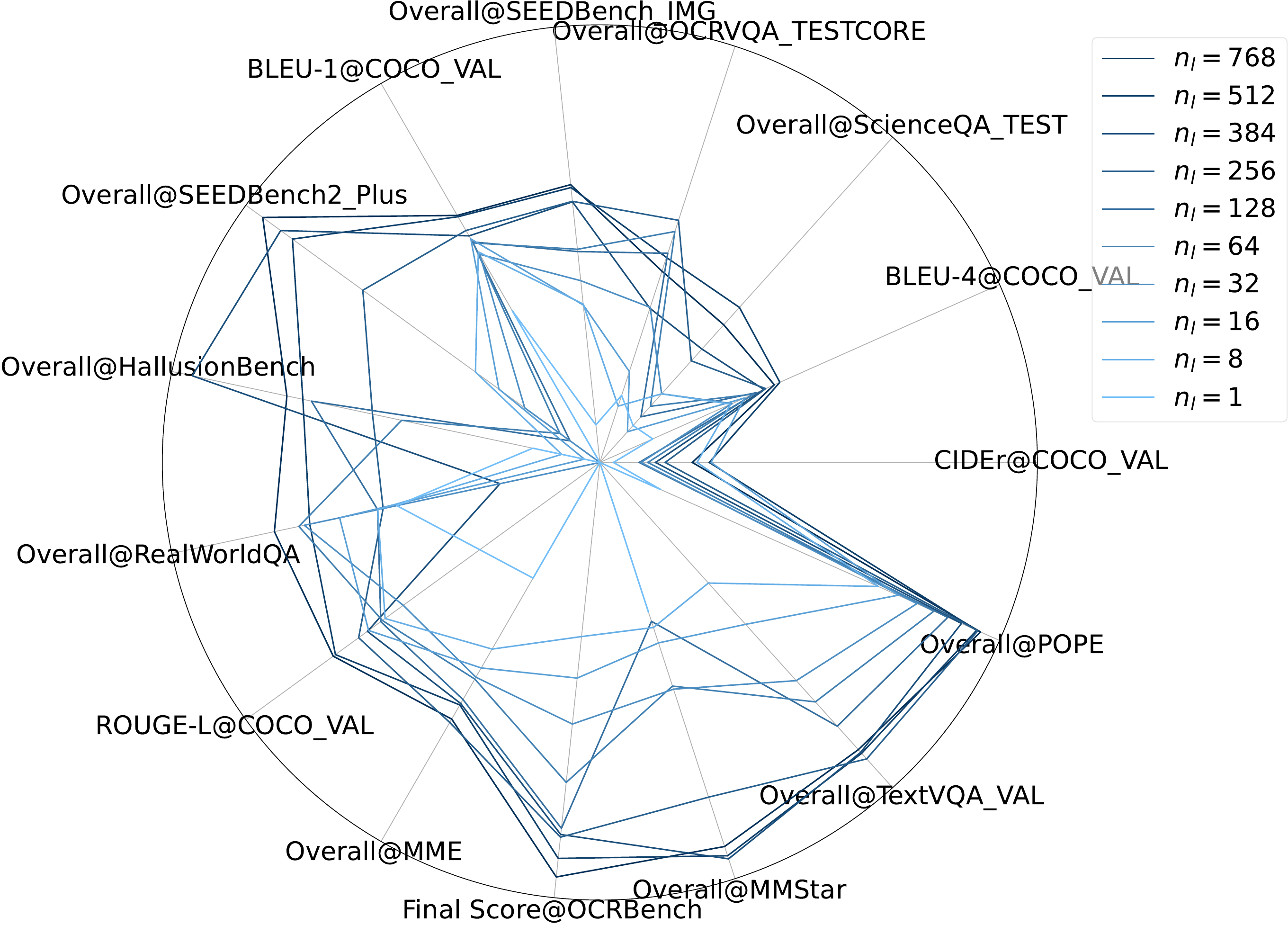}
    \caption{Performance variation across \(n_l\)}
    \label{fig:tab_radar_chart_nl}
  \end{subfigure}%
  \begin{subfigure}{0.49\textwidth}
    \centering
    \includegraphics[width=\linewidth]{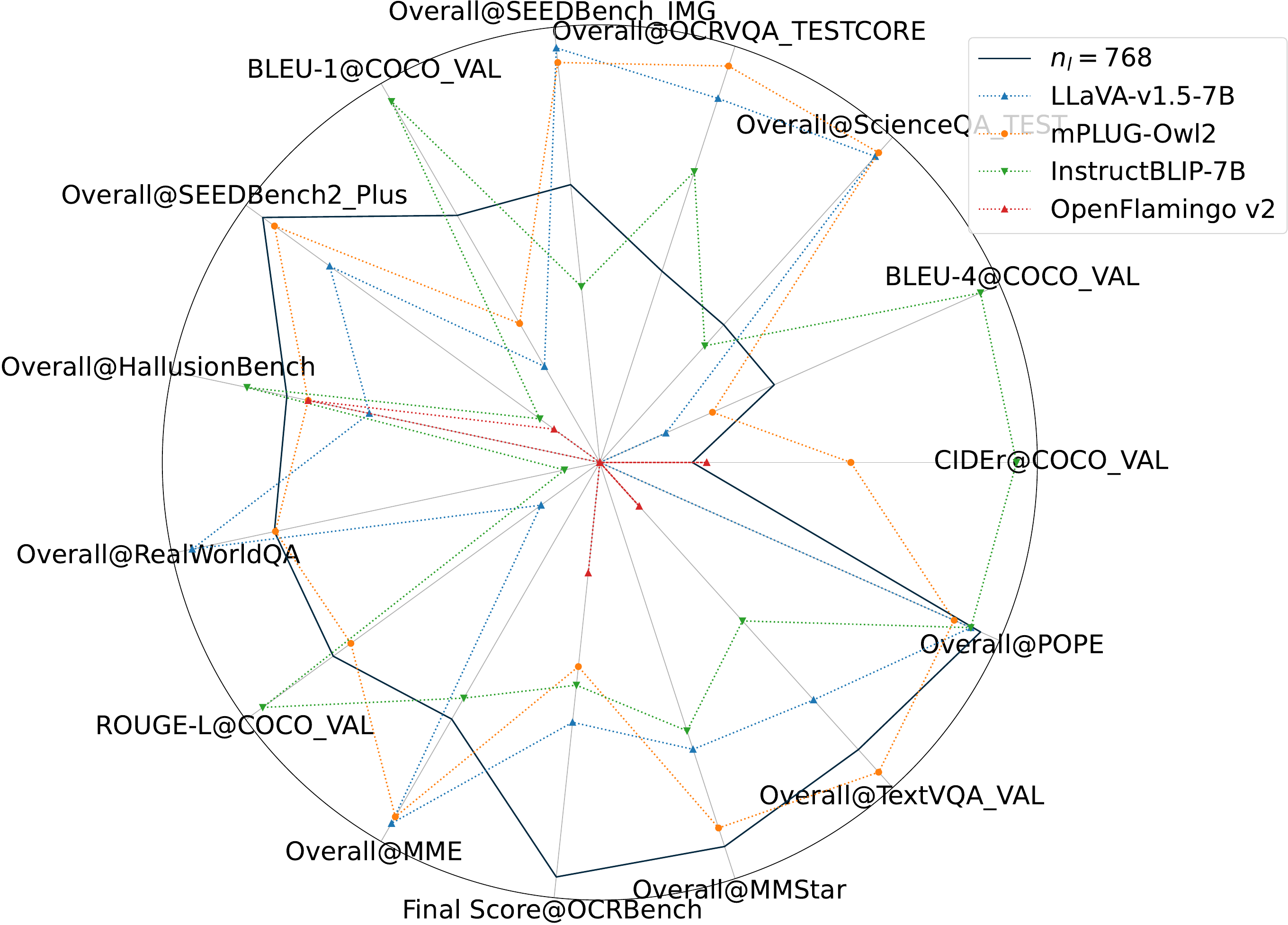}
    \caption{Comparison of \(n_l=768\) with baseline}
    \label{fig:tab_radar_chart_picked}
  \end{subfigure}
  \caption{
    Performance comparison between the proposed model and baseline models across various benchmarks.
    For normalization purposes, each benchmark's performance scores are scaled to the interval \([0,1]\) using the transformation \(\tvar{x} \leftarrow (\tvar{x} - \min(\tvar{x})) / (\max(\tvar{x}) - \min(\tvar{x}))\).
    Baseline model results are obtained from VLMEvalKit evaluations.
  }
  \label{fig:tab_radar_chart}
\end{figure}

Figure \ref{fig:tab_radar_chart_nl} shows the performance variation of the proposed model with different values of \(n_l\). The light blue color corresponds to the proposed model with smaller \(n_l\), and the dark blue color corresponds to the proposed model with larger \(n_l\). The radar chart shows that increasing \(n_l\) improves the model's performance across most benchmarks, consisting with the theoretical analysis presented earlier.

The comparative analysis, as illustrated in Figure \ref{fig:tab_radar_chart_picked}, reveals the performance characteristics of both the proposed model (represented by solid line) and the baseline models (represented in dashed lines).
The proposed model demonstrates competitive performance relative to baseline models on several evaluation benchmarks, including OCRBench, MMStart POPE, SEEDBench2 Plus, and HallusionBench. These results suggest that the model effectively captures the underlying patterns required for these specific tasks.
However, performance gaps are observed on other benchmarks, including ScienceQA Test, SeedBench IMG, RealWorld QA, and MME, where the proposed model underperforms compared to established baselines. These disparities may be attributed to differences in training data, model architecture design choices, or task-specific optimization strategies employed by the baseline models.

\subsection{Comparison with Token Reduction Methods}

The experimental comparison evaluates the proposed method against two established token reduction approaches: VisPruner \citep{zhang2024FasterVLM}, which utilizes visual cues from the vision encoder for token pruning, and VisionZip \citep{yang2025visionzip}, which selects a set of informative tokens for the language model.
A comprehensive comparison across multiple benchmarks is conducted to assess the performance of the proposed method relative to these baseline methods.
Results are presented in Table \ref{tab:compare_token_reduction_methods}.

The experimental results reveal that each method exhibits distinct strengths and weaknesses across different task categories.
The proposed method shows relatively lower performance on certain benchmarks such as MME and RealWorldQA, where visual perception capabilities are primarily evaluated.
However, the proposed method demonstrates better performance on text-intensive tasks, including OCRBench, ChartQA, and TextVQA, suggesting more effective preservation of textual information during token reduction.

\begin{table}[htb]
    \centering
    \setlength{\tabcolsep}{1.0pt}
    \renewcommand{\arraystretch}{1.0}
    \footnotesize
    \caption{
        Comparison of token reduction methods.
        VisPruner \citep{zhang2024FasterVLM} is configured with dominant \(n_l - 10\), contextual token 10, and base model LLaVA-v1.5-7B.
        VisionZip \citep{yang2025visionzip} is configured with visual token num \(n_l\), important ratio \(0.5\), and base model LLaVA-v1.5-7B.
        The proposed method is configured with \(n_l\), and use ``Vision Question Queries'' as input.
        The largest value for each setting is marked \textbf{bold}.
    }
    \label{tab:compare_token_reduction_methods}
    \hspace*{-3.5em}
    \begin{tabular}{crrrrrrrrrrrrrrrrr}
        \toprule
        Methods       &
        \multicolumn{1}{c}{
            \rotatebox{90}{MME}
        }             &
        \multicolumn{1}{c}{
            \rotatebox{90}{OCRBench}
        }             &
        \multicolumn{1}{c}{
            \rotatebox{90}{HallusionBench}
        }             &
        \multicolumn{1}{c}{
            \rotatebox{90}{POPE}
        }             &
        \multicolumn{1}{c}{
            \rotatebox{90}{AI2D TEST}
        }             &
        \multicolumn{1}{c}{
            \rotatebox{90}{RealWorldQA}
        }             &
        \multicolumn{1}{c}{
            \rotatebox{90}{MMStar}
        }             &
        \multicolumn{1}{c}{
            \rotatebox{90}{ScienceQA TEST}
        }             &
        \multicolumn{1}{c}{
            \rotatebox{90}{OCRVQA TESTCORE}
        }             &
        \multicolumn{1}{c}{
            \rotatebox{90}{ChartQA TEST}
        }             &
        \multicolumn{1}{c}{
            \rotatebox{90}{TextVQA VAL}
        }             &
        \multicolumn{1}{c}{
            \rotatebox{90}{SEEDBench IMG}
        }             &
        \multicolumn{1}{c}{
            \rotatebox{90}{SEEDBench2 Plus}
        }             &
        \multicolumn{1}{c}{
            \rotatebox{90}{BLEU-1 COCO VAL}
        }             &
        \multicolumn{1}{c}{
            \rotatebox{90}{BLEU-4 COCO VAL}
        }             &
        \multicolumn{1}{c}{
            \rotatebox{90}{ROUGE-L COCO VAL}
        }             &
        \multicolumn{1}{c}{
            \rotatebox{90}{CIDEr COCO VAL}
        }
        \\
        \midrule
        \(n_l = 512\) &                                                                                                                                                                                                                                                                                                   \\
        VisPruner     & \textbf{1691.82} & 308.00          & 23.11          & 81.19          & 51.88          & 48.10          & 32.00          & \textbf{68.02} & \textbf{52.90} & 14.84          & 22.76          & \textbf{60.15} & 38.87          & 21.19          & 5.13           & 21.30          & 0.95           \\
        VisionZip     & 1527.86          & 303.00          & 19.76          & 69.75          & \textbf{51.98} & 48.24          & 31.67          & 67.58          & 41.08          & 12.88          & 20.26          & 59.63          & 38.95          & 21.36          & 5.19           & 21.56          & 1.18           \\
        Proposed      & 1414.27          & \textbf{472.00} & \textbf{29.56} & \textbf{86.63} & 40.84          & \textbf{49.15} & \textbf{35.40} & 57.16          & 38.02          & \textbf{51.48} & \textbf{53.55} & 53.34          & \textbf{43.39} & \textbf{40.71} & \textbf{10.57} & \textbf{34.70} & \textbf{15.26} \\
        \hline
        \(n_l = 384\) &                                                                                                                                                                                                                                                                                                   \\
        VisPruner     & \textbf{1682.13} & 312.00          & 24.39          & 84.59          & 51.30          & \textbf{49.28} & 31.93          & \textbf{68.17} & \textbf{54.65} & 14.88          & 25.29          & \textbf{59.89} & 38.91          & 21.32          & 5.21           & 21.65          & 1.49           \\
        VisionZip     & 1590.55          & 295.00          & 20.43          & 77.56          & \textbf{51.65} & 48.63          & 32.73          & 67.77          & 47.49          & 13.16          & 20.77          & 59.46          & 38.47          & 21.30          & 5.21           & 21.55          & 1.19           \\
        Proposed      & 1409.44          & \textbf{445.00} & \textbf{32.81} & \textbf{85.25} & 39.41          & 40.00          & \textbf{35.47} & 53.84          & 30.79          & \textbf{43.20} & \textbf{53.40} & 52.07          & \textbf{44.05} & \textbf{38.04} & \textbf{9.74}  & \textbf{32.42} & \textbf{7.76}  \\
        \hline
        \(n_l = 256\) &                                                                                                                                                                                                                                                                                                   \\
        VisPruner     & \textbf{1551.58} & 302.00          & 26.33          & 84.88          & 51.04          & 47.71          & 32.00          & \textbf{67.72} & \textbf{52.51} & 13.84          & 23.61          & \textbf{59.14} & 38.30          & 22.05          & 5.44           & 22.70          & 2.85           \\
        VisionZip     & 1535.73          & 299.00          & 20.74          & 80.91          & \textbf{52.27} & \textbf{49.02} & 31.73          & 67.43          & 47.14          & 12.88          & 19.84          & 58.62          & 38.91          & 22.28          & 5.47           & 22.87          & 2.85           \\
        Proposed      & 1472.53          & \textbf{448.00} & \textbf{27.51} & \textbf{86.73} & 39.48          & 45.62          & \textbf{34.13} & 52.90          & 43.29          & \textbf{50.60} & \textbf{54.40} & 52.11          & \textbf{39.44} & \textbf{38.79} & \textbf{9.84}  & \textbf{33.04} & \textbf{9.11}  \\
        \hline
        \(n_l = 128\) &                                                                                                                                                                                                                                                                                                   \\
        VisPruner     & \textbf{1605.53} & 291.00          & 24.19          & \textbf{84.61} & 51.00          & \textbf{47.71} & \textbf{30.40} & \textbf{68.32} & \textbf{42.02} & 12.48          & 19.77          & \textbf{57.66} & 38.30          & 25.87          & 6.57           & 28.26          & 15.76          \\
        VisionZip     & 1529.75          & 286.00          & 21.97          & 82.53          & \textbf{51.59} & 46.93          & 29.33          & 68.02          & 34.11          & 11.64          & 17.20          & 56.89          & \textbf{38.60} & 26.92          & 6.78           & 29.26          & \textbf{17.30} \\
        Proposed      & 1396.15          & \textbf{438.00} & \textbf{29.30} & 84.02          & 33.61          & 45.88          & 30.33          & 48.44          & 38.61          & \textbf{41.32} & \textbf{49.47} & 47.63          & 27.80          & \textbf{37.19} & \textbf{9.38}  & \textbf{31.46} & 5.49           \\
        \hline
        \(n_l = 64\)  &                                                                                                                                                                                                                                                                                                   \\
        VisPruner     & \textbf{1536.95} & 283.00          & 21.54          & 80.32          & 49.68          & 45.62          & 29.27          & 67.28          & 25.52          & 10.72          & 16.30          & \textbf{55.54} & 37.99          & 30.25          & 7.76           & 32.85          & 27.24          \\
        VisionZip     & 1509.86          & 273.00          & 20.45          & 77.80          & \textbf{50.42} & 47.06          & 28.67          & \textbf{67.63} & 21.81          & 11.04          & 15.87          & 55.27          & \textbf{38.95} & 33.46          & 8.78           & \textbf{34.80} & \textbf{32.47} \\
        Proposed      & 1328.09          & \textbf{386.00} & \textbf{26.65} & \textbf{82.76} & 35.04          & \textbf{49.67} & \textbf{31.73} & 49.28          & \textbf{41.70} & \textbf{38.16} & \textbf{45.80} & 47.83          & 28.37          & \textbf{37.05} & \textbf{9.21}  & 31.35          & 6.70           \\
        \hline
        \(n_l = 32\)  &                                                                                                                                                                                                                                                                                                   \\
        VisPruner     & 1379.74          & 239.00          & 19.92          & 73.10          & 48.15          & 44.84          & 26.87          & \textbf{67.43} & 14.94          & 10.16          & 13.03          & 51.24          & 38.56          & 38.56          & 9.77           & 36.75          & 37.08          \\
        VisionZip     & \textbf{1419.03} & 224.00          & 20.12          & 66.95          & \textbf{49.09} & 43.14          & 28.00          & 66.63          & 21.42          & 11.08          & 13.07          & \textbf{51.66} & \textbf{38.65} & \textbf{45.69} & \textbf{12.52} & \textbf{39.02} & \textbf{44.58} \\
        Proposed      & 1321.83          & \textbf{320.00} & \textbf{20.81} & \textbf{81.23} & 35.46          & \textbf{49.41} & \textbf{31.80} & 50.27          & \textbf{30.92} & \textbf{29.68} & \textbf{42.60} & 45.03          & 30.35          & 35.47          & 8.11           & 29.85          & 5.74           \\
        \hline
        \(n_l = 16\)  &                                                                                                                                                                                                                                                                                                   \\
        VisPruner     & 1203.98          & 150.00          & 18.65          & 49.75          & 46.89          & 41.70          & 26.20          & \textbf{66.14} & 16.44          & 9.40           & 9.36           & \textbf{46.72} & \textbf{38.16} & \textbf{45.12} & 10.69          & \textbf{35.76} & \textbf{34.66} \\
        VisionZip     & 1076.22          & 64.00           & 17.77          & 36.26          & \textbf{47.22} & 38.43          & 24.93          & 64.70          & \textbf{25.10} & 10.68          & 7.21           & 44.70          & 37.07          & 44.02          & \textbf{11.36} & 30.48          & 28.54          \\
        Proposed      & \textbf{1291.08} & \textbf{268.00} & \textbf{21.27} & \textbf{79.53} & 36.30          & \textbf{47.71} & \textbf{30.80} & 47.25          & 21.81          & \textbf{20.32} & \textbf{34.11} & 42.85          & 31.80          & 37.61          & 8.75           & 32.33          & 15.51          \\
        \bottomrule
    \end{tabular}
\end{table}

\subsection{Computational Cost Analysis}

The main computational overhead arises from high-dimensional processing and additional fusion modules, which are computed exclusively during the prefill phase.
Therefore, it is better to measure the time to first token (TTFT).
To address the trade-off between token count and computational efficiency, TTFT and GPU resource utilization are measured.

The evaluations are conducted on 500 randomly selected images from the COCO test set \citep{COCO} using the prompt ``Please describe the given image into detail.'' .
The results in Table \ref{tab:ttft} demonstrate a clear computational trade-off pattern.
TTFT decreases substantially from approximately 0.3 seconds at larger \(n_l\) to around 0.2 seconds at smaller \(n_l\).
GPU utilization follows a similar trend, dropping from over 80\% to approximately 73-77\%.
Memory usage remains relatively stable across all configurations.
It is notable that introducing two additional modules, the high-dimensional processing modules and fusion modules, will incur additional computational costs, therefore more computational resources are required than the methods without these modules.

\begin{table}[htbp]
    \centering
    \setlength{\tabcolsep}{3.0pt}
    \renewcommand{\arraystretch}{1.0}
    \small
    \caption{
        Comparison of time to first token (TTFT), GPU utilization, and GPU memory usage.
        The proposed method is configured with \(n_l\), and use ``Vision Question Queries'' as input.
        The GPU utilization and memory usage are measured on a single NVIDIA RTX 3090 GPU (24G).
        Note: TTFTs below are measured from image loading, tokenization to first token generation.
    }
    \label{tab:ttft}
    \begin{tabular}{crrrr}
        \toprule
        \(n_l\) & TTFT (s) & GPU util (\%) & GPU peak (GiB) & GPU average (GiB) \\
        \midrule
        768     & 0.3038   & 81.9693       & 15.4844        & 15.4832           \\
        512     & 0.3024   & 83.8838       & 15.4844        & 15.4830           \\
        256     & 0.3013   & 83.9863       & 15.4844        & 15.4830           \\
        128     & 0.2554   & 79.9462       & 15.4258        & 15.4244           \\
        64      & 0.2228   & 73.8923       & 15.4786        & 15.4770           \\
        32      & 0.2203   & 74.1203       & 15.4454        & 15.4439           \\
        16      & 0.2211   & 76.7657       & 15.5391        & 15.5373           \\
        8       & 0.2181   & 75.9544       & 15.4005        & 15.3990           \\
        1       & 0.2137   & 73.0568       & 15.4395        & 15.4338           \\
        \bottomrule
    \end{tabular}
\end{table}

To further quantify the computational trade-offs associated with varying vision token counts, the number of floating-point operations (FLOPs) is measured across different configurations.
The measurements are conducted on 500 randomly selected images from the COCO test set using the prompt ``Describe the image in detail.'' with a maximum of 256 newly generated tokens per image.
The time cost of \texttt{model.forward} is recorded for both prefill and decode phases, and the results for prefill time, decode throughput (tokens per second), and FLOPs are presented in Table~\ref{tab:inf_time_flops}.

The results reveal that computational cost scales approximately linearly with vision token count. 
Specifically, FLOPs\footnote{FLOPs are measured using the library \texttt{thop}: \url{https://github.com/Lyken17/pytorch-OpCounter}} increase from 2.380 TFLOPs to 4.151 TFLOPs as \(n_l\) increases from 1 to 768, and the same trend is observed in prefill time.
In contrast, decode throughput remains relatively stable across different \(n_l\) values due to the KV cache mechanism.
During the decode phase, attention computations are substantially reduced because only the newly generated token attends to the cached key-value pairs from previous tokens, including vision tokens.
Consequently, variations in \(n_l\) from 1 to 768 do not significantly impact attention computation costs during autoregressive generation, resulting in consistent tokens-per-second rates across configurations.

Balancing the trade-off between computational efficiency and performance is crucial for practical deployment, particularly in resource-constrained environments or latency-sensitive applications.
Measuring the time cost can provide valuable insights into the computational trade-offs associated with different vision token counts, helping to inform the selection of an optimal configuration for specific use cases.

\begin{table}[htbp]
    \centering
    \setlength{\tabcolsep}{3.0pt}
    \renewcommand{\arraystretch}{1.0}
    \small
    \caption{
        Comparison of prefill time, tokens per second in decode phase and TFLOPs of the prefill.
        The proposed method is configured with \(n_l\), and use ``Vision Question Queries'' as input.
        The GPU utilization and memory usage are measured on a single NVIDIA RTX 3090 GPU (24G).
        Note: the data below are measured on \texttt{model.forward};
        FLOPs will be different due to different size of input images (A high resolution image will be cropped into nine patches. If its size is not big enough (\(336 \le \text{width or height} \le 336 \times 3\)), the patches without any pixels will be ignored.);
        all data cells are format as (mean \(\pm\) std).
    }
    \label{tab:inf_time_flops}
    \begin{tabular}{cccc}
        \toprule
        \(n_l\) & Prefill Time (s)     & Tokens/s             & TFLOPs (\(1e^{12}\)) \\
        \midrule
        768     & \(0.255 \pm 0.023\) & \(23.114 \pm 0.103\) & \(4.151 \pm 0.326\)         \\
        512     & \(0.255 \pm 0.023\) & \(23.092 \pm 0.113\) & \(4.151 \pm 0.326\)         \\
        384     & \(0.285 \pm 0.020\) & \(23.032 \pm 0.088\) & \(5.040 \pm 0.326\)         \\
        256     & \(0.255 \pm 0.023\) & \(23.100 \pm 0.099\) & \(4.151 \pm 0.326\)         \\
        128     & \(0.211 \pm 0.022\) & \(23.049 \pm 0.083\) & \(3.262 \pm 0.326\)         \\
        64      & \(0.178 \pm 0.022\) & \(23.051 \pm 0.130\) & \(2.817 \pm 0.326\)         \\
        32      & \(0.177 \pm 0.022\) & \(23.067 \pm 0.115\) & \(2.595 \pm 0.326\)         \\
        16      & \(0.177 \pm 0.022\) & \(23.141 \pm 0.124\) & \(2.484 \pm 0.326\)         \\
        8       & \(0.175 \pm 0.022\) & \(23.188 \pm 0.121\) & \(2.428 \pm 0.326\)         \\
        1       & \(0.172 \pm 0.029\) & \(23.080 \pm 0.145\) & \(2.380 \pm 0.326\)         \\
        \bottomrule
    \end{tabular}
\end{table}

\section{\texorpdfstring{Further Analysis on \(n_l\)}{Further Analysis on nl}}
\label{appendix:further-analysis-Nl}

The following analysis focuses on the general trends and patterns observed across different experimental configurations.

To facilitate comprehensive comparison across benchmarks with varying scales, the results are normalized using min-max scaling: \(\tvar{x} \leftarrow (\tvar{x} - \min(\tvar{x})) / (\max(\tvar{x}) - \min(\tvar{x}))\), which transforms all results to the range \([0, 1]\).
This normalization enables consistent comparison of performance trends across diverse benchmarks while preserving the relative relationships within each dataset.
The normalized results reveal several distinct behavioral patterns when analyzed as functions of \(n_l\).
To characterize these patterns and better illustration, K-means clustering is applied to partition the results into four clusters, providing a better visualization for observing how \(n_l\) influences model performance and identifying optimal configurations across different scenarios.

\begin{figure}[htb]
  \centering
  \begin{subfigure}{0.45\textwidth}
    \centering
    \includegraphics[width=\linewidth]{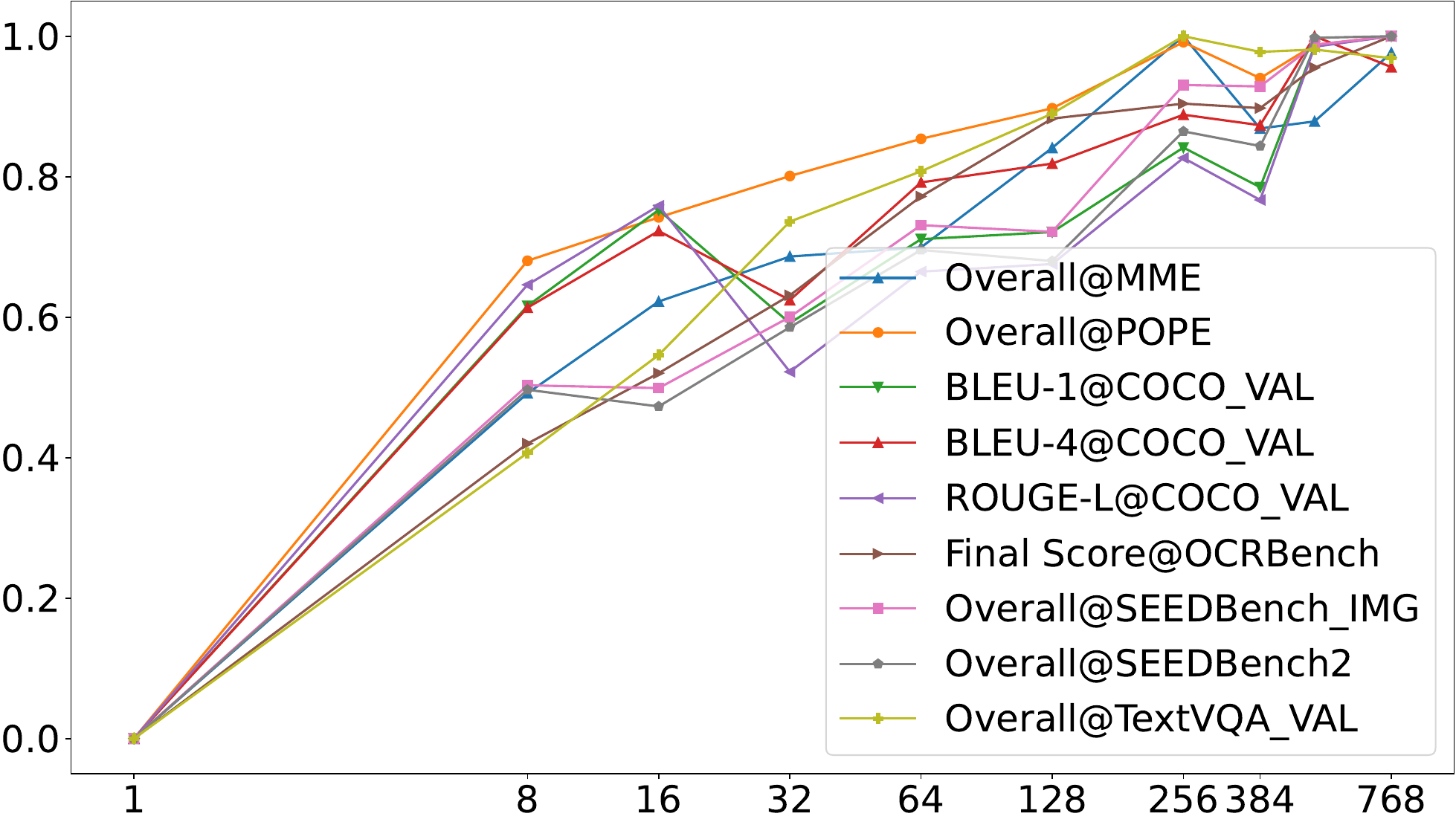}
    \caption{}
    \label{fig:tab_merged_vqq_c0}
  \end{subfigure}%
  \hspace*{1em}
  \begin{subfigure}{0.45\textwidth}
    \centering
    \includegraphics[width=\linewidth]{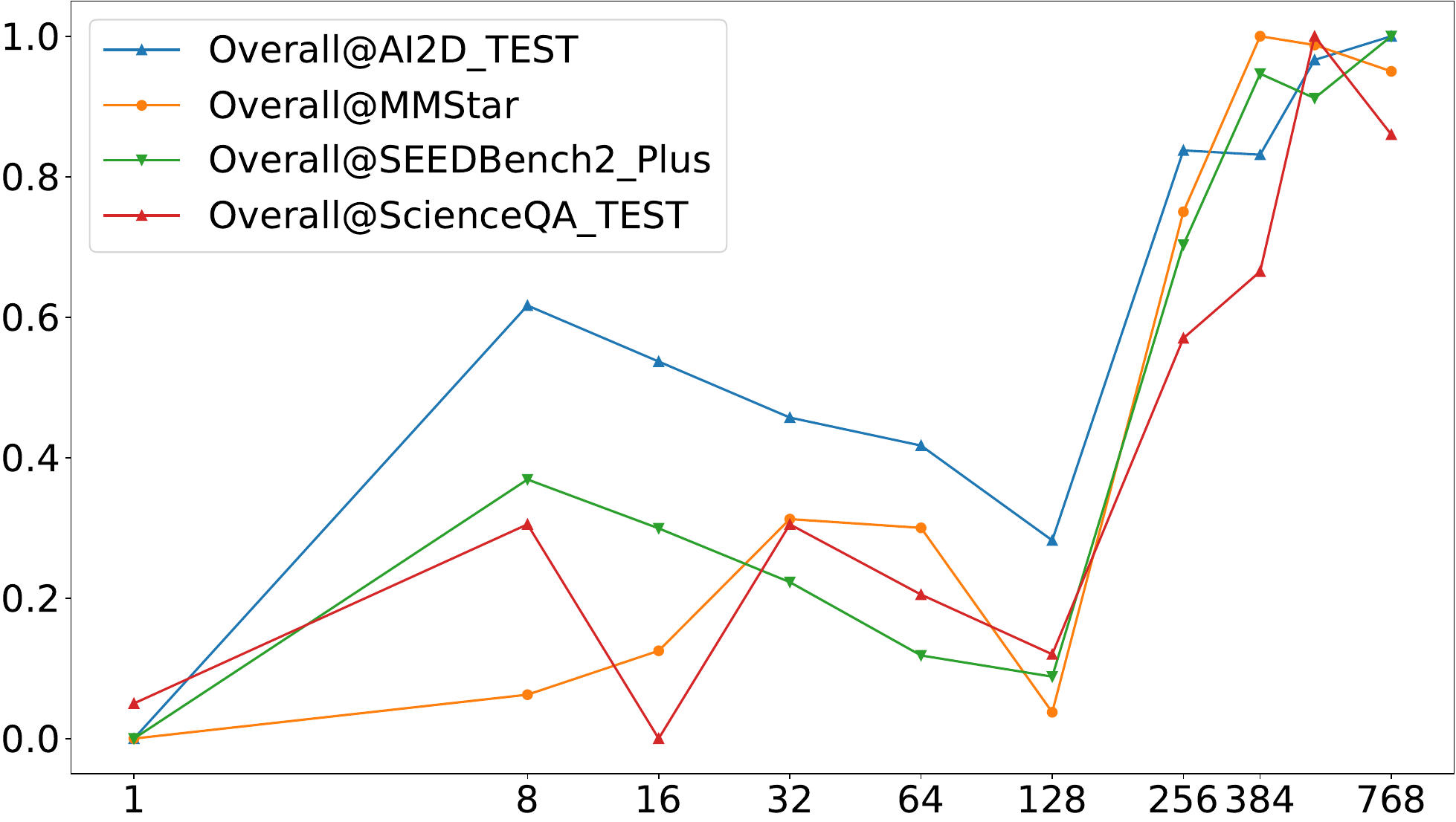}
    \caption{}
    \label{fig:tab_merged_vqq_c1}
  \end{subfigure}
  \begin{subfigure}{0.45\textwidth}
    \centering
    \includegraphics[width=\linewidth]{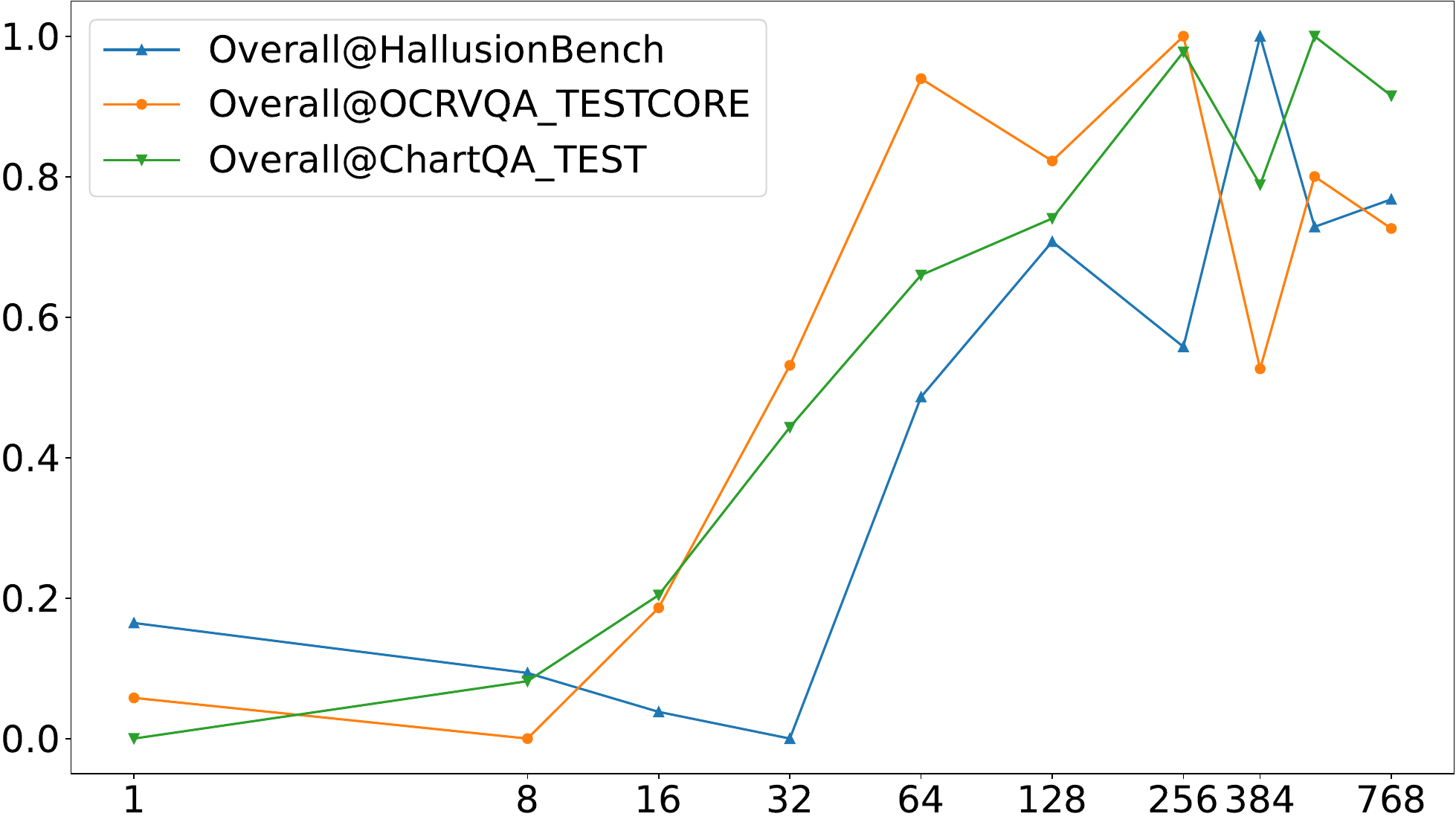}
    \caption{}
    \label{fig:tab_merged_vqq_c3}
  \end{subfigure}
  \hspace*{1em}
  \begin{subfigure}{0.45\textwidth}
    \centering
    \includegraphics[width=\linewidth]{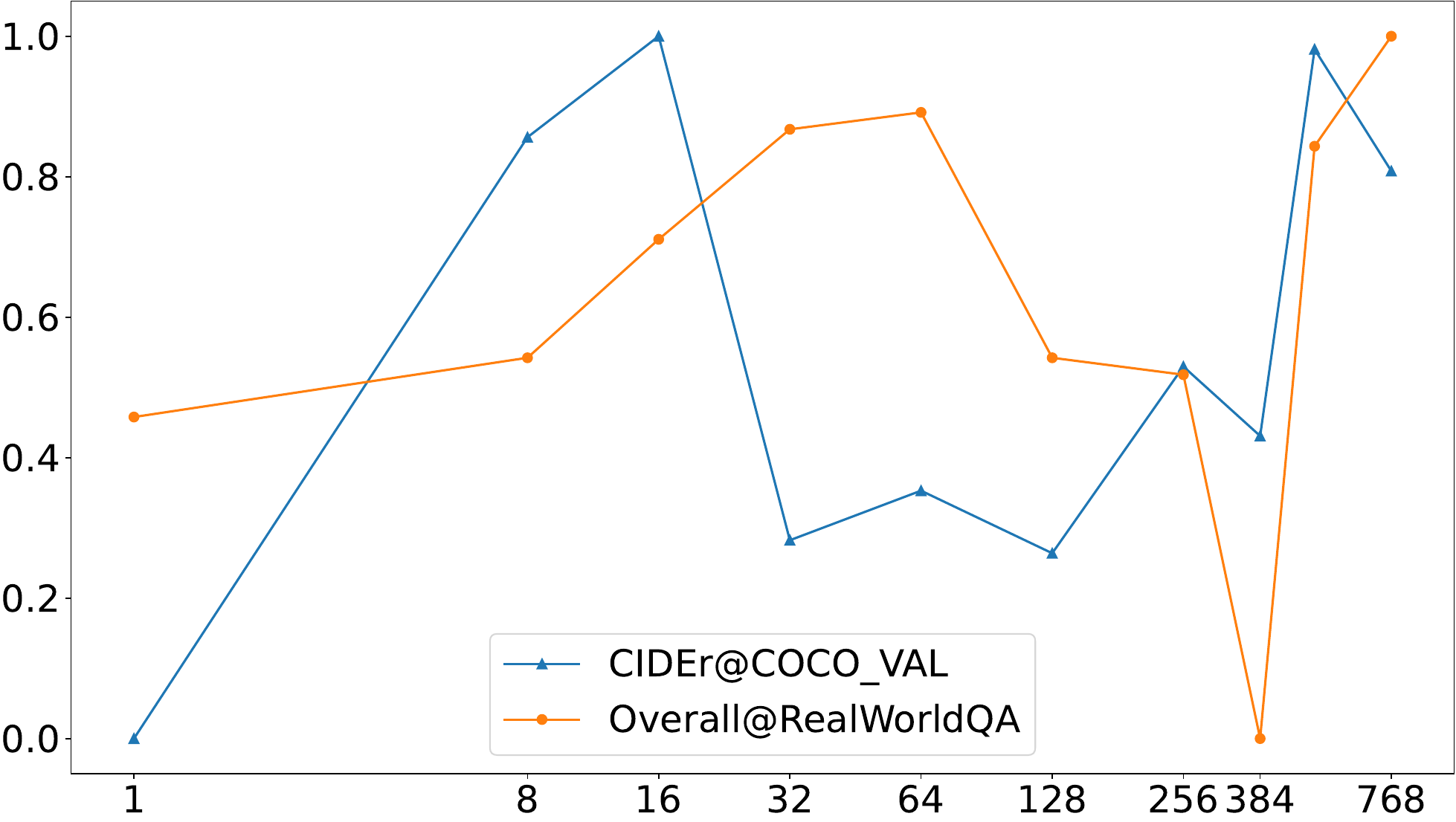}
    \caption{}
    \label{fig:tab_merged_vqq_c2}
  \end{subfigure}
  \caption{
    Clustered view of performances.
    Inputs: vision tokens, user questions, and learnable queries.
    Values are normalized using min-max scaling. X-axis is \(\log_2(n_l)\).
  }
  \label{fig:tab_merged_vqq}
\end{figure}

\textbf{Analysis with vision tokens, questions, and learnable queries.}
The analysis begins with the model configuration that processes vision tokens, user questions, and learnable queries as inputs.
The results, illustrated in Figure \ref{fig:tab_merged_vqq}, reveal distinct clusters that characterize the relationship between the number of learnable queries (\(n_l\)) and model performance.

A consistent observation across all clusters is that model performance remains suboptimal when \(n_l\) is small.
As \(n_l\) increases, performance improves substantially before eventually stabilizing at higher values.
This pattern demonstrates that \(n_l\) serves as a critical factor in enabling the model to achieve optimal performance.

Figure \ref{fig:tab_merged_vqq_c0} demonstrates an approximately linear relationship between performance and \(\log_2(n_l)\), indicating consistent logarithmic scaling behavior within this cluster.
In Figure \ref{fig:tab_merged_vqq_c1}, the model exhibits degraded performance for \(n_l < 128\), followed by rapid improvement as \(n_l\) increases, ultimately achieving stability at higher values.
Similarly, Figure \ref{fig:tab_merged_vqq_c3} shows poor performance for \(n_l < 64\), with subsequent rapid improvement and stabilization as \(n_l\) grows. These observations reinforce the critical importance of sufficient \(n_l\) for achieving optimal model performance.
However, the behavior observed in Figure \ref{fig:tab_merged_vqq_c2} exhibits relatively noisy behavior. There is a sharp drop in performance at specific \(n_l\) values, followed by a gradual recovery.
This pattern suggests that the under specific tasks or datasets, there could be unexpected noise or instability in the model performance.

\begin{figure}[htbp]
  \centering
  \begin{subfigure}{0.45\textwidth}
    \centering
    \includegraphics[width=\linewidth]{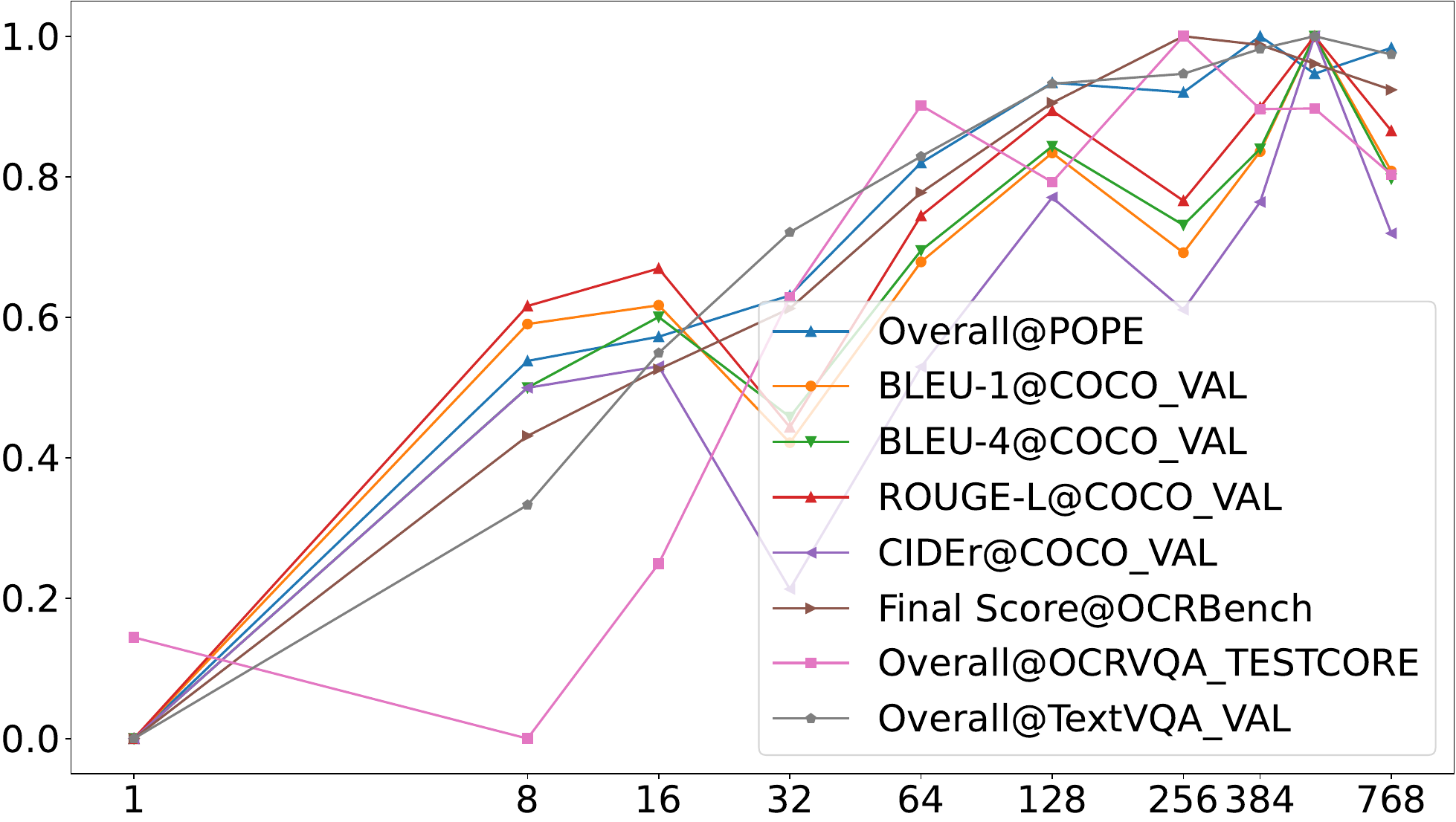}
    \caption{}
    \label{fig:tab_merged_vq_ft_c3}
  \end{subfigure}%
  \hspace*{1em}
  \begin{subfigure}{0.45\textwidth}
    \centering
    \includegraphics[width=\linewidth]{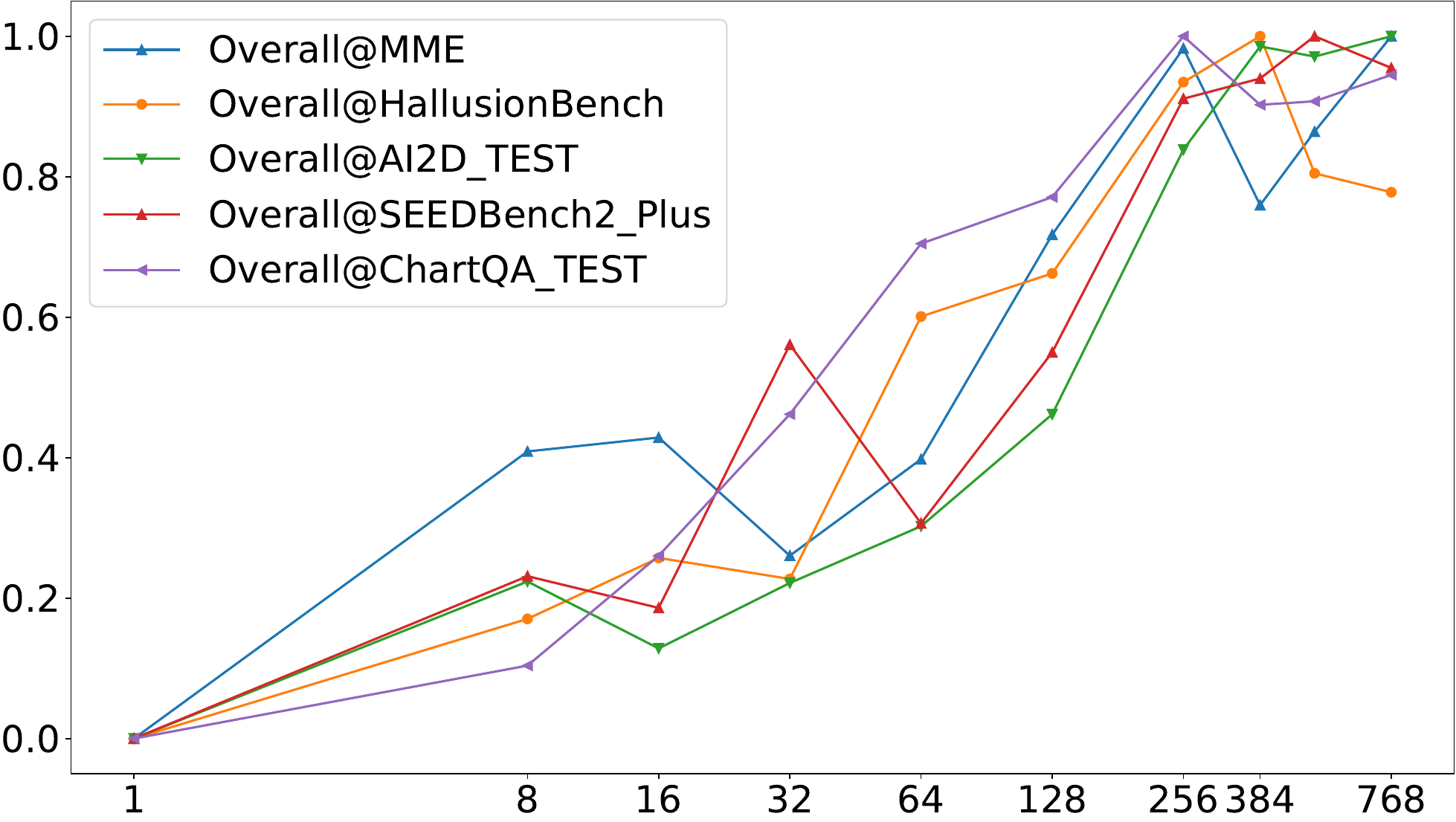}
    \caption{}
    \label{fig:tab_merged_vq_ft_c0}
  \end{subfigure}
  \begin{subfigure}{0.45\textwidth}
    \centering
    \includegraphics[width=\linewidth]{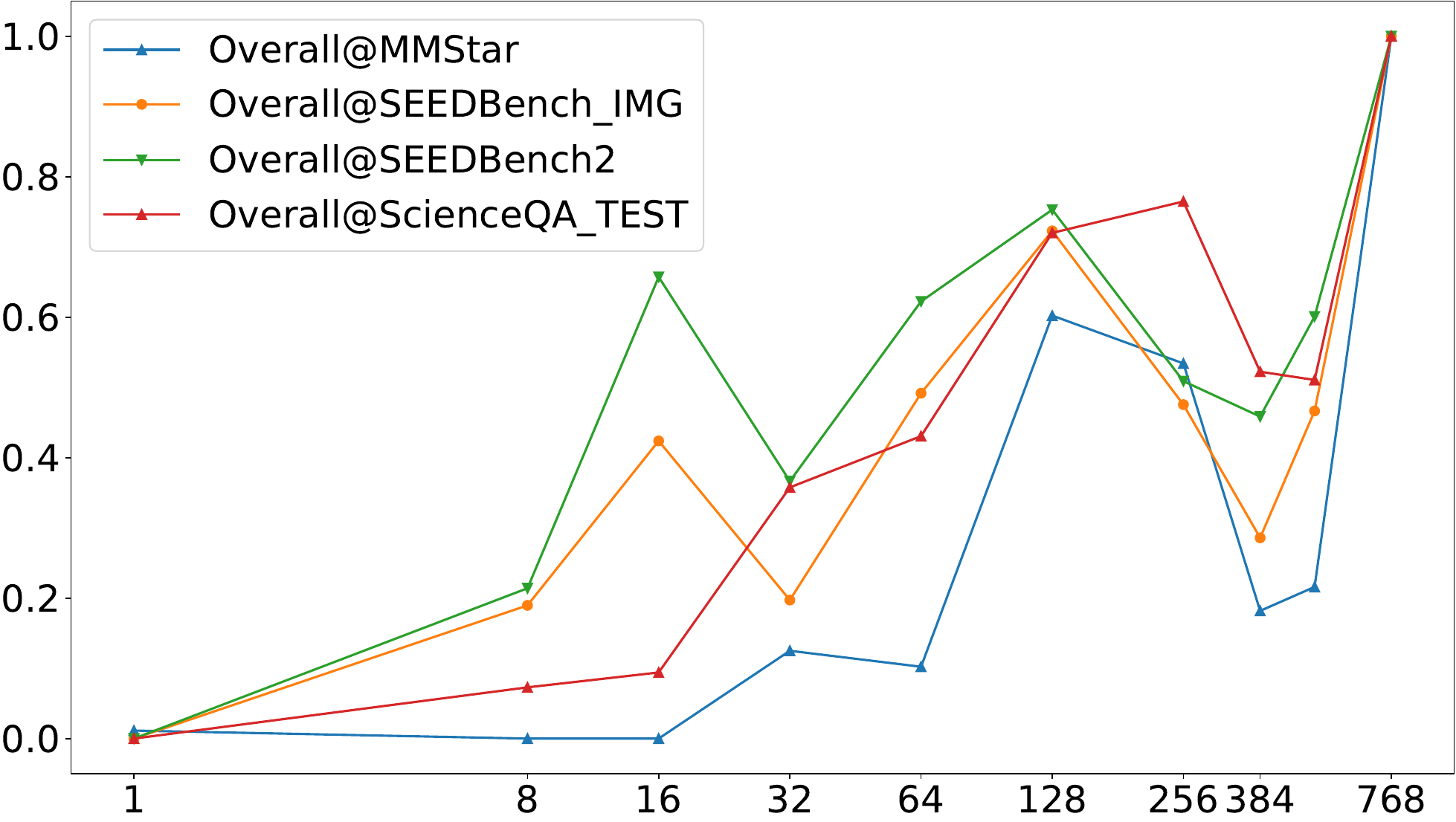}
    \caption{}
    \label{fig:tab_merged_vq_ft_c1}
  \end{subfigure}%
  \hspace*{1em}
  \begin{subfigure}{0.45\textwidth}
    \centering
    \includegraphics[width=\linewidth]{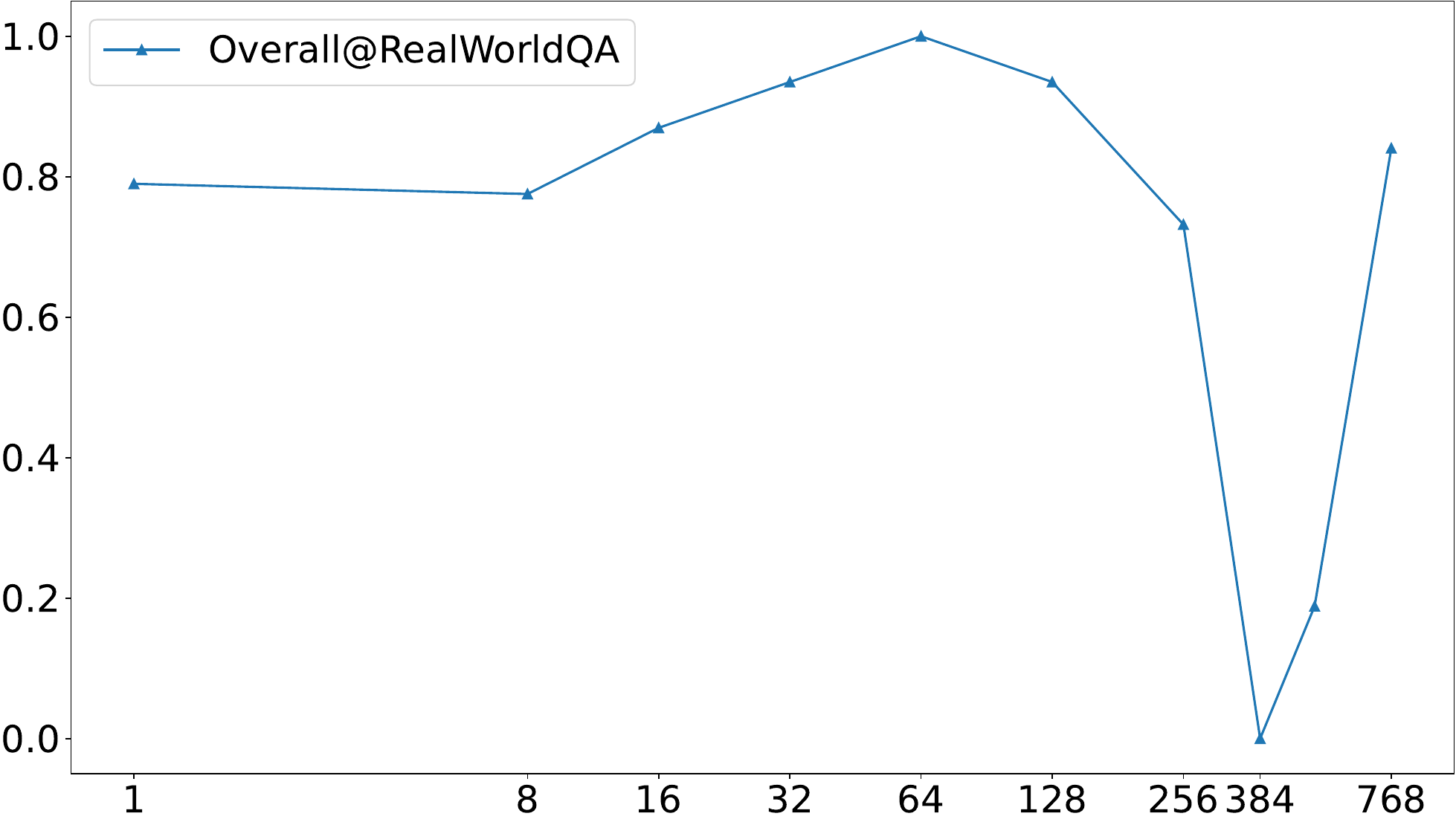}
    \caption{}
    \label{fig:tab_merged_vq_ft_c2}
  \end{subfigure}
  \caption{
    Clustered view of performances.
    Inputs: vision tokens, learnable queries (excluding user questions).
    Values are normalized using min-max scaling.
    X-axis is \(\log_2(n_l)\).
  }
  \label{fig:tab_merged_vq_ft}
\end{figure}

\textbf{Analysis without user questions.}
To examine the impact of incorporating user text questions on model performance, an alternative configuration is evaluated where the model excludes user questions as input.
In this setup, the model processes only vision tokens and learnable queries.
The model described in Appendix \ref{appendix:implementation_details} is fine-tuned under this modified configuration.
The resulting performance trends are presented in Figure \ref{fig:tab_merged_vq_ft}, which reveals four distinct clusters with varying patterns relative to \(n_l\).

Figures \ref{fig:tab_merged_vq_ft_c3} and \ref{fig:tab_merged_vq_ft_c0} demonstrate behavior approximately proportional to \(\log_2(n_l)\), indicating consistent logarithmic scaling trends within these clusters.
Figure \ref{fig:tab_merged_vq_ft_c1} exhibit less consistent linear relationships, however the performance still improves with increasing \(n_l\).
Figure \ref{fig:tab_merged_vq_ft_c2} is an exception, showing a sharp drop in performance at \(n_l=128\). However, if this point is excluded, the rest of the data can show a logarithmic trend.

\section{Example of Inference}
\label{appendix:example-inference}

The following illustrate two demonstrations of the generation results produced by the models.

\begin{small}
  
\subsection{Example 1}

\begin{figure}[htb]
  \centering
  \includegraphics[width=0.45\linewidth]{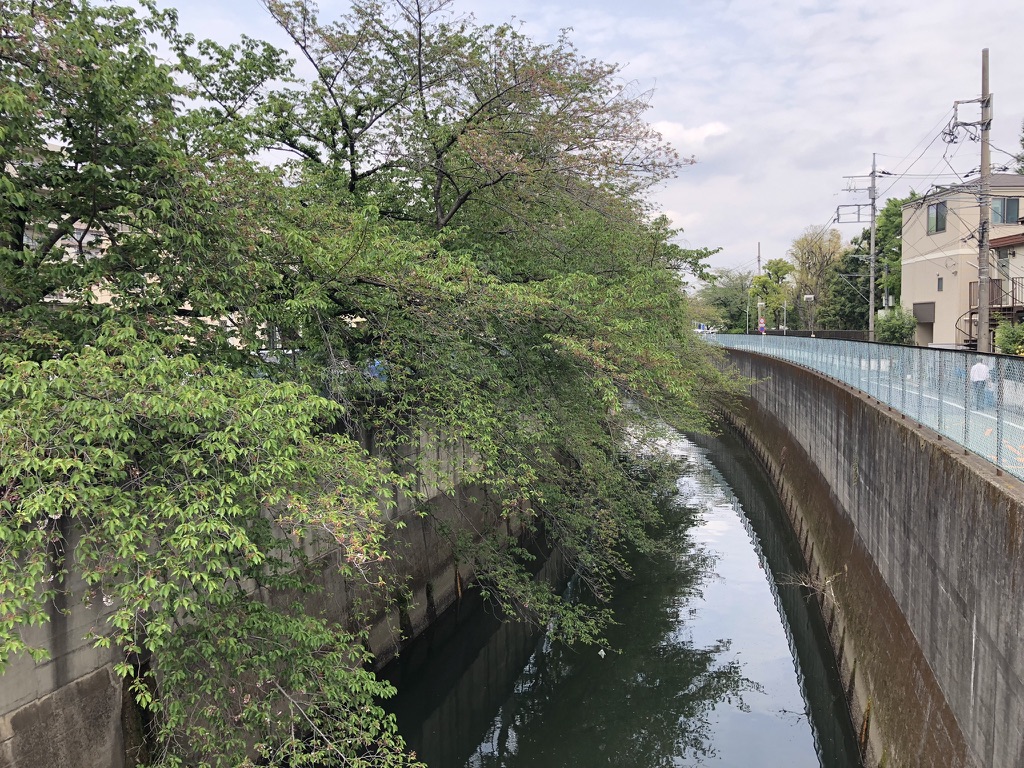}
  \caption{Image of example 1.}
  \label{fig:appendix_example_1}
\end{figure}

User's question: ``Please describe the image in full detail.''

Configuration \(n_l = 1\):
\begin{tcolorbox}[demobox]
The image is of a small river flowing through a green landscape. The river is surrounded by trees and {\color{red} rocks}, and there are several {\color{red} bridges} crossing it. The river is also lined with rocks and greenery.
\end{tcolorbox}

Configuration \(n_l = 8\):
\begin{tcolorbox}[demobox]
The image is of a river flowing through a city. There are trees and buildings along the riverbank. The river is surrounded by a concrete wall. There are also {\color{red}some rocks} in the river.
\end{tcolorbox}

Configuration \(n_l = 16\):
\begin{tcolorbox}[demobox]
The image is of a river flowing through a concrete channel. The river is surrounded by trees and buildings. There are also some houses and a bridge in the background.
\end{tcolorbox}

Configuration \(n_l = 32\):
\begin{tcolorbox}[demobox]
The image shows a long, narrow, concrete drainage ditch running along a wall of a building. The ditch is surrounded by trees and greenery, and there are people walking along the wall. The wall is made of brick and has a large pipe running along it.
\end{tcolorbox}

Configuration \(n_l = 64\):
\begin{tcolorbox}[demobox]
The image is of a canal or river with a wall on one side and a building on the other. There are trees and greenery along the wall.
\end{tcolorbox}

Configuration \(n_l = 128\):
\begin{tcolorbox}[demobox]
A canal is surrounded by a wall and has a green tree growing on it.
\end{tcolorbox}

Configuration \(n_l = 256\):
\begin{tcolorbox}[demobox]
The image is a scenic view of a canal, with a wall on the left and a fence on the right. The canal is surrounded by trees, with leaves and branches visible. There are also houses and buildings visible in the background. The image is quite detailed, with a lot of elements visible.
\end{tcolorbox}

Configuration \(n_l = 384\):
\begin{tcolorbox}[demobox]
A canal with a concrete wall and a tree growing on the side.
\end{tcolorbox}

Configuration \(n_l = 512\):
\begin{tcolorbox}[demobox]
The image shows a scene of a canal with a wall on the side, surrounded by trees and houses. The canal is filled with water, and there are reflections of the trees and houses on the water. The wall is made of concrete, and there are houses on the other side of the canal.
\end{tcolorbox}

Configuration \(n_l = 768\):
\begin{tcolorbox}[demobox]
The image shows a scene of a canal, with a wall of trees on one side and a fence on the other. The canal is surrounded by a concrete wall, and there are trees growing on the wall. The trees are lush and green, with leaves and branches visible. There are also people walking along the fence, adding to the scene's vibrancy.
\end{tcolorbox}

Baseline model LLaVA-v1.5-7B-hf:
\begin{tcolorbox}[demobox]
The image features a small stream of water flowing through a city, surrounded by trees and buildings. The stream is lined with trees, creating a serene atmosphere. There are several people scattered throughout the scene, some closer to the water and others further away.

In the background, a {\color{red}car} can be seen parked near the stream, indicating that the area is accessible to pedestrians and {\color{red}vehicles}. The combination of the waterway, trees, and people creates a picturesque urban scene.
\end{tcolorbox}

Baseline model InstructBLIP-7B-hf:
\begin{tcolorbox}[demobox]
The image features a narrow canal or river running alongside a building, with trees lining the water's edge. There are several trees visible along the canal, with some closer to the water and others further away. The trees are spread out along the canal, creating a serene and peaceful atmosphere. A {\color{red} few cars can be seen parked} along the side of the canal, adding to the urban setting. 
\end{tcolorbox}

\subsection{Example 2}

\begin{figure}[htb]
  \centering
  \includegraphics[width=0.7\linewidth]{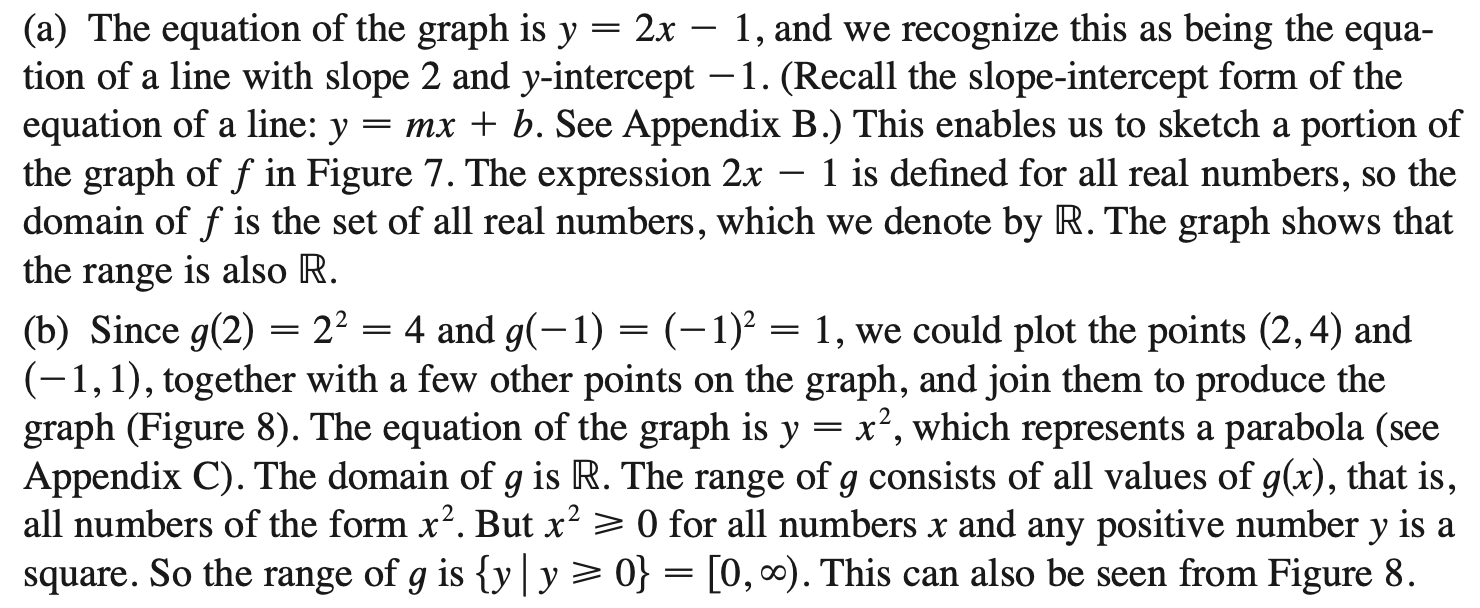}
  \caption{Image of example 2.}
  \label{fig:appendix_example_2}
\end{figure}

User's question: ``Please read all text from the image.''

Configuration \(n_l = 1\):
\begin{tcolorbox}[demobox]
\begin{tabularlstlisting}
_ _ _ _ _ _ _ _ _ _ _ _ _ _ _ _ _ _ _ _ _ 
\end{tabularlstlisting}
\end{tcolorbox}

Configuration \(n_l = 8\):
\begin{tcolorbox}[demobox]
\begin{tabularlstlisting}
x 2 = 0.
The equation is a linear equation in two variables, x and y.
The graph of the equation is a straight line.
The equation is a function of x and y.
The equation is a function of
\end{tabularlstlisting}
\end{tcolorbox}

Configuration \(n_l = 16\):

\begin{tcolorbox}[demobox]
\begin{tabularlstlisting}
The equation of the line is:
\(y=m\*x+b\),
where m is the slope of the line and b is the y-intercept.
The slope of the line is the ratio of the change in the y-coordinate to the change in the x-coordinate. In other words, it is the rate at which the line rises or falls. The y-intercept is the point at which the line crosses the y-axis.
The slope of a line can be calculated by dividing the change in the y-coordinate by the change in the x-coordinate. For example, if the line passes through the point (2, 3), the slope is 3/2.
The y-intercept can be calculated by adding the slope to the point at which the line crosses the y-axis. For example, if the line passes through the point (2, 3), the y-intercept is 3.
The equation of a line can be written in the form:
\(y=m\*x+b\),
where m is the slope and b is the y-intercept.
The equation of a line can also be written in the form:
\(y=m\*x+c\),
where m
\end{tabularlstlisting}
\end{tcolorbox}

Configuration \(n_l = 32\):
\begin{tcolorbox}[demobox]
\begin{tabularlstlisting}
## 3.1.1 The equation is:
\begin{aligned}
\text{the equation is} &= \text{as we are expecting} \\
\text{and} &= \text{the slope of the line is} \\
\text{the slope of the line is} &= \text{the slope of the line is} \\
\text{the slope of the line is} &= \text{the slope of the line is
\end{tabularlstlisting}
\end{tcolorbox}

Configuration \(n_l = 64\):
\begin{tcolorbox}[demobox]
\begin{tabularlstlisting}
## The equation of a line with slope and y-intercept
The equation of a line: y = mx + b. + See Appendix B. + See Appendix C. + See Appendix D.
The graph of the equation in Figure 7.2 is defined for all values of x. The expression x = 0 is defined for all values of y. The range of the graph is also R.
Since g = 2 since g = 1 and g = -1, we can write the equation of the line as:
y = mx + b. + See Appendix B. + See Appendix C. + See Appendix D.
The equation of a line with slope and y-intercept is a linear equation. The slope of the line is m, and the y-intercept is b. The equation of a line with slope and y-intercept is a linear equation. The slope of the line is m, and the y-intercept is b.
The equation of a line with slope and y-intercept is a linear
\end{tabularlstlisting} 
\end{tcolorbox}

Configuration \(n_l = 128\):
\begin{tcolorbox}[demobox]
\begin{tabularlstlisting}
The equation of a line with slope s and a point (x,y) is 2x - y = s. We recognize this as the equation of a line in slope-intercept form of a line: mx + b = x. See Appendix B. This enables us to sketch a portion of the graph of f in Figure 7. The expression 2 is defined for all real numbers, so the domain of f is the set of all real numbers, which we denote by R. The graph shows that so is also R. (2) Since g(2) = s since g(2) = s, we could plot the points (2,4) and (1,1) on the graph. (1,-1), 1, 1, 
\end{tabularlstlisting}
\end{tcolorbox}

Configuration \(n_l = 256\):
\begin{tcolorbox}[demobox]
\begin{tabularlstlisting}
The equation of the graph is y = 2x - 1, and we recognize this as being the equation of a line with slope 2 and y-intercept -1. (Recall the slope-intercept form of the equation of a line: y = mx + b. See Appendix B.) This enables us to sketch a portion of the graph of f in Figure 7. The expression 2x - 1 is defined for all real numbers, so the domain of f is the set of all real numbers, which we denote by R. The graph shows that the range is also R.
\(\begin{array}{c} \text{Since} \ g(2)=2=1\text{, we could plot the points (2,4) and} \\ \text{together, with a few other points on the graph, and join them to produce the graph (Figure 8). The equation of the graph is y = x2, which represents a parabola (see Appendix C). The domain of g is R. The range of g consists of all values of g(x), that is, all numbers of the form x2. But x2 > 0 for any positive number y is a square. So the range of g is [y | y | y] = 0. This can also be seen from Figure 8.
\end{tabularlstlisting} 
\end{tcolorbox}

Configuration \(n_l = 384\):
\begin{tcolorbox}[demobox]
\begin{tabularlstlisting}
(a) The equation of the graph is y = 2x - 1, and we recognize this as being the equation of a line with slope 2 and y-intercept (1. Recall the slope-intercept form of the equation of a line: y = mx + b. See Appendix B.) This enables us to sketch a portion of the graph of f in Figure 7. The expression 2x - 1 is defined for all real numbers, so the domain of f is the set of all real numbers, which we denote by R. The graph shows that the range is also R. (b) Since g(2) = 2a = -1, we could plot the points (2, 4) and (1, 1), together with a few other points on the graph, and join them to produce the graph (Figure 8). The equation of the graph is y = x2, which represents a parabola (see Appendix C). The domain of g is R. The range of g consists of all values of g(x), that is, all numbers of the form x2. But x2 > 0 for all numbers and any positive number y is a. So the range of g is [0, y] = [0, y] = [0, y] = [0, y] = 
\end{tabularlstlisting}
\end{tcolorbox}

Configuration \(n_l = 512\):
\begin{tcolorbox}[demobox]
\begin{tabularlstlisting}
* The equation of the graph is y = 2x - 1, and we recognize this as being the equation of a line with slope 2 and y-intercept -1. (Recall the slope-intercept form of the equation of a line: y = mx + b. See Appendix B.) This enables us to sketch a portion of the graph of f in Figure 7. The expression 2x - 1 is defined for all real numbers, so the domain of f is the set of all real numbers, which we denote by R. The graph shows that the range is also R. (b) Since g(2) = 2 = 1, we could plot the points (2, 4) and (1, -1), together with a few other points on the graph, and join them to produce the graph (Figure 8). The equation of the graph is y = x2, which represents a parabola (see Appendix C). The domain of g is R. The range of g consists of all numbers of the form x2. But x2 > 0 for any positive number y is a square. So the range of g is y = 0 = 0, or y = 0 = 0. This can also be seen from Figure 8.
\end{tabularlstlisting}
\end{tcolorbox}

Configuration \(n_l = 768\):
\begin{tcolorbox}[demobox]
\begin{tabularlstlisting}
<table> <thead>
<tr>
<td>The equation of the graph is y = 2x - 1, and we recognize this as being the equation of a line with slope 2 and y-intercept -1. (Recall the slope-intercept form of the equation of a line: y = mx + b. See Appendix B.) This enables us to sketch a portion of the graph of f in Figure 7. The expression 2x - 1 is defined for all real numbers, so the domain of f is the set of all real numbers, which we denote by R. The graph shows that the range is also R. (b) Since g(2) = 2 = (1 - 1) = (1 - 1), together with a few other points on the graph, and join them to produce the graph (Figure 8). The equation of the graph is y = x2, which represents a parabola (see Appendix C). The domain of g is R. The range of g consists of all values of g, that is, all numbers of the form x2. But x2 > 0 for any positive number y is a square. So the range of g is [y | y | y] = [0 | 0 | 0]. This can also be seen from Figure 8. </td> <td>The equation of the graph is y = 2x - 1, and we recognize this as being the equation of a line with slope 2 and y-intercept -1. (Recall the slope-intercept form of the equation of a line: y = mx + b. See Appendix B.) This enables us to sketch a portion of the graph of f in Figure 7. The expression 2x - 1 is defined for all real numbers, so the domain of f is the set of all real numbers, which we denote by R. The graph shows that the range is also R. (b) Since g(2) = 2 = (1 - 1) = (1 - 1), together with a few other points on the graph, and join them to produce the graph (Figure 8). The equation of the graph is y = x2, which represents a parabola (see Appendix C). The domain of g is R. The range of g consists of all values of g, that
\end{tabularlstlisting}
\end{tcolorbox}

Baseline model LLaVA-v1.5-7B-hf:
\begin{tcolorbox}[demobox]
\begin{tabularlstlisting}
The image displays a graph with a slope, and the equation is written below it. The graph is described as a "real number" graph, and the slope is defined as a "real number." The equation is written in terms of the slope, and the graph is described as a "real number" graph. The slope is also described as a "real number." The equation is written in terms of the slope, and the graph is described as a "real number" graph. The slope is also described as a
\end{tabularlstlisting}
\end{tcolorbox}

Baseline model InstructBLIP-7B-hf:
\begin{tcolorbox}[demobox]
\begin{tabularlstlisting}
the example of graph is shown in the following
\end{tabularlstlisting}
\end{tcolorbox}

\end{small}

\end{document}